\documentclass{article}

 \usepackage[main, final]{neurips_2026}
\usepackage[utf8]{inputenc} % allow utf-8 input
\usepackage[T1]{fontenc}    % use 8-bit T1 fonts
\usepackage{hyperref}       % hyperlinks
\usepackage{url}            % simple URL typesetting
\usepackage{booktabs}       % professional-quality tables
\usepackage{amsfonts}       % blackboard math symbols
\usepackage{nicefrac}       % compact symbols for 1/2, etc.
\usepackage{microtype}      % microtypography
\usepackage{xcolor}         % colors
\usepackage{cite}
\usepackage{graphicx}
\usepackage{subcaption}
\usepackage{float}
\usepackage{amsmath}
\usepackage{array}
\usepackage{multirow}
\usepackage{caption} 
\usepackage{booktabs}  
\usepackage{tabularx}  
\usepackage{amsmath}
\usepackage{makecell}
\usepackage{algorithm}
\usepackage[noend]{algpseudocode}
\usepackage{setspace}
\usepackage{comment}

\title{CARE: A Cascaded Framework for Efficient and Reliable Time Series Anomaly Detection}

\author{%
  Zemin Chao, Qianhui Xu, Jianhe Cen, Guangzhi Ge, Xiao Chen, Hongzhi Wang\thanks{Corresponding author.}\\
  Massive Data Computing Lab, Harbin Institute of Technology\\
  \mbox{\texttt{chaozm@hit.edu.cn}
  \qquad
  \texttt{wangzh@hit.edu.cn}}
}

\begin{document}

\maketitle

\begin{abstract}

While deep learning models have achieved state-of-the-art performance in time series anomaly detection, their complex architectures incur substantial inference overhead. Existing methods typically apply a uniform inference strategy across all data points, which is inefficient given that anomalies are inherently scarce and the vast majority of temporal data consists of predictable normal patterns. To mitigate this bottleneck, we propose CARE, a model-agnostic cascaded inference framework that integrates a Lightweight Pre-filter Model (LPM) with an existing high-capacity Complex Detection Model (CDM). The LPM rapidly filters high-confidence normal samples using a Residual MLP AutoEncoder and a Normality-Conditioned Gating mechanism. Crucially, we introduce a Structure Attention module to explicitly capture channel-wise anomaly contributions, and optimize the gating network via a confidence-guided selective routing objective that learns reliable routing decisions to reduce unnecessary CDM invocations. Extensive experiments across eight real-world benchmarks demonstrate that CARE effectively isolates high-confidence normal samples. By routing only uncertain samples to the CDM, our framework achieves $2.7\times$ to $4.8\times$ inference speedup compared to the most accurate SOTA approaches, while still maintaining competitive detection quality.

\end{abstract}

\section{Introduction}

Time series anomalies serve as critical early indicators of underlying faults and emerging risks in modern cyber-physical systems~\citep{DarbanWPAS25, Hawkins80, Blazquez-Garcia21}. Consequently, Time Series Anomaly Detection (TSAD) has emerged as a crucial research frontier in data mining, driving essential real-world applications such as fraud detection~\citep{YuXCZJZ24}, healthcare monitoring~\citep{RoyMHB23}, and urban management~\citep{DouYJQR25}. 
Although early traditional methods~\citep{BreunigKNS00, icdm/LiuTZ08, nips/ScholkopfWSSP99} laid the groundwork for identifying basic outliers,  they generally struggle to capture the highly complex, non-linear temporal dependencies pervasive in real-world data streams~\citep{00030CB0W25}.

Deep learning has since transformed TSAD by enabling automated feature extraction and the modeling of intricate temporal dynamics. To capture the complex dynamics of real-world data, recent methodologies have increasingly relied on sophisticated, high-capacity architectures, such as Transformer-based models for long-term dependencies~\citep{nips/DaiHYL24,nips/Beibu25}, contrastive learning for latent normal patterns~\citep{TangDWZ24,pvldb/ZhuangZZGYWF24}, and frequency-domain analysis for multi-view relationships~\citep{icde/FangXZ0G024, WuQL0HGXY25}. However, this continuous pursuit of superior detection quality via architectural complexity comes at a steep computational price. Consequently, deploying these deep TSAD models in practical, efficiency-sensitive scenarios exposes critical limitations.

\begin{figure}[t]
  \centering
  
  \begin{subfigure}[t]{0.51\textwidth}
    \centering
    \includegraphics[width=\textwidth]{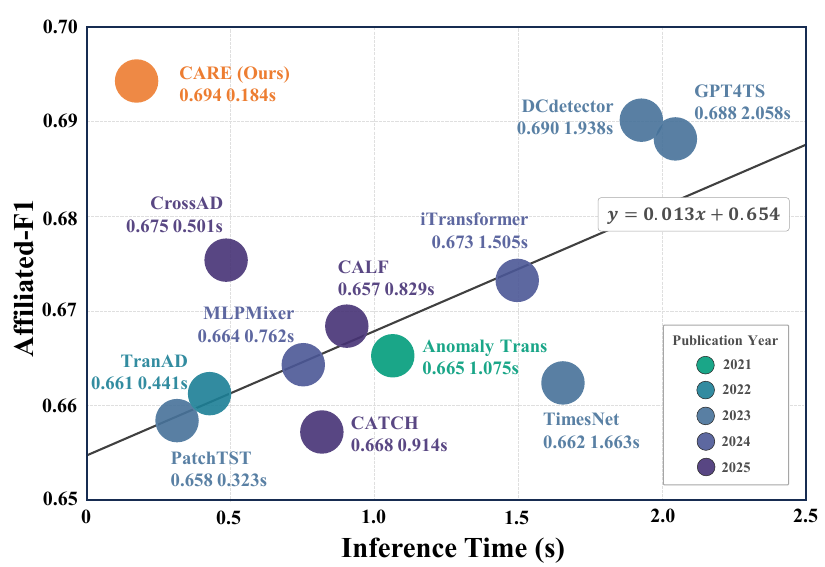}
    \caption{Model performance comparison on NYC.}
    \label{fig:fig1_a}
  \end{subfigure}
  \hfill
  \begin{subfigure}[t]{0.465\textwidth}
    \centering
    \includegraphics[width=\textwidth]{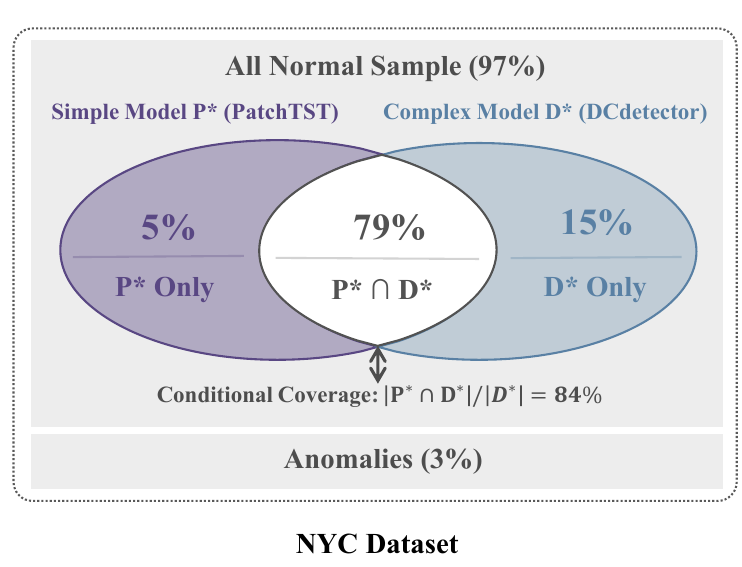}
    \caption{Normal coverage overlap between models.}
    \label{fig:fig1_b}
  \end{subfigure}

\caption{
 Motivation of the proposed CARE framework. (a) The trade-off between detection performance and inference cost among existing methods. CARE explicitly bridges this gap, delivering state-of-the-art accuracy with exceptionally low latency. (b) Over 84\% of the normal samples identified by the high-capacity model (DCdetector) are also successfully captured by the lightweight model (PatchTST),  indicating most of the candidates can be detected with simple models.
}

  \label{fig:Figure1}
\end{figure}

\textbf{Limitation 1: Severe inference bottlenecks in quality-leading detection models.} 
While increasing model complexity generally yields superior detection quality, it inherently introduces substantial inference overhead. As depicted in Figure~\ref{fig:fig1_a}, sota methods (e.g., DCdetector and GPT4TS) exhibit prohibitively high inference costs~\citep{QiuLQHZWLGZSHJY25}. Conversely, lightweight models (e.g., TranAD and PatchTST) offer desirable low-latency inference but at the expense of competitive detection accuracy. Unfortunately, TSAD methodologies that explicitly co-optimize for inference acceleration and detection quality remain insufficiently explored, leaving a critical gap in deploying these models at scale.

\textbf{Limitation 2: Uniform complex inference on predominantly normal patterns.} 
Due to the inherent scarcity of anomalous events~\citep{aggarwal2017, SchmidlWP22}, real-world temporal data is dominated by normal samples.
A substantial portion of these normal instances exhibits stable, predictable patterns that can be effectively captured by relatively simple architectures.
As illustrated in Figure~\ref{fig:fig1_b}, simple architectures (e.g., PatchTST) can accurately identify the vast majority of normal temporal patterns alongside high-capacity models, indicating that most data is trivial enough to be processed without complex inference.
Despite this, existing TSAD frameworks typically adopt a rigid, uniform inference strategy, indiscriminately applying complex models to all data points. This forces trivial normal samples to undergo costly and redundant inference procedures, leading to unnecessary overhead.

To address these aforementioned limitations, we propose a \textbf{CA}scaded framework for \textbf{R}eliable and \textbf{E}fficient time series anomaly detection (CARE), which rethinks the inference process by seamlessly integrating a Lightweight Pre-filter Model (LPM) with a high-capacity Complex Detection Model (CDM).
Operating as an efficient front-end, the LPM employs a Residual MLP-based AutoEncoder coupled with a Normality-Conditioned Gating Module. This gating mechanism rapidly assesses the normality confidence of incoming temporal windows: high-confidence normal instances are safely intercepted and evaluated locally with the fast LPM model, while uncertain, high-risk samples are routed to the CDM for refined but more expensive analysis.
To ensure reliable routing without compromising detection quality, 
the gating network utilizes a structure attention mechanism that captures channel-wise anomaly contributions. It is optimized via a confidence-guided objective that learns reliable routing to reduce unnecessary CDM invocations while maintaining detection quality.
Therefore, the proposed approach breaks the traditional performance-efficiency trade-off, achieving efficient and reliable anomaly judgment. Our main contributions are:

\begin{itemize}

\item We propose CARE, the first cascaded inference framework tailored for efficient time series anomaly detection. By fundamentally rethinking the uniform complex inference paradigm, CARE serves as a versatile, model-agnostic architecture that seamlessly integrates with, and substantially accelerates, existing high-capacity anomaly detectors.

\item We design a Lightweight Pre-filter Model (LPM) equipped with a novel structure attention mechanism to explicitly capture channel-wise anomaly contributions. Furthermore, we introduce a confidence-guided selective routing strategy to optimize the gating network, enabling reliable routing decisions that reduce unnecessary CDM invocations while preserving detection quality.

\item Extensive experiments across eight diverse real-world benchmarks demonstrate that CARE effectively isolates trivial normal samples, breaking the traditional performance-efficiency bottleneck. Notably, CARE achieves a speedup of inference by $2.7\times$ to $4.8\times$ while maintaining, and in several cases even surpassing, state-of-the-art detection quality.

\end{itemize}

\section{Related Work}
\label{related work}

Recent advancements in deep learning have significantly improved Time Series Anomaly Detection (TSAD). State-of-the-art methodologies, including forecasting-based, reconstruction-based, and representation-based models, as well as recent adaptations of Large Language Models (LLMs), have achieved remarkable detection accuracy~\citep{TangDWZ24, icde/FangXZ0G024, LiuHZWWML24, nips/ZhouNW0023}. However, this continuous pursuit of accuracy frequently relies on highly complex architectures (e.g., Transformers or multi-view fusions). While effective, these methods incur prohibitive inference latency, rendering them computationally expensive and impractical for latency-sensitive practical deployments.

To balance performance and efficiency, cascaded inference models have been extensively explored in domains such as computer vision and natural language processing~\citep{pami/HanHSYWW22, tmlr/ChenZ024}. The core principle is to process straightforward samples with lightweight models while strategically reserving high-capacity architectures for complex cases. In the context of TSAD, anomalous events are inherently scarce, meaning the vast majority of normal samples can be accurately identified without invoking complex neural architectures~\citep{aggarwal2017, SchmidlWP22}. Despite this, existing TSAD frameworks typically adopt a rigid, uniform inference strategy. Our CARE framework bridges this gap by introducing a cascaded paradigm tailored for time series, effectively circumventing redundant computations.

\textit{Due to space constraints, an extended and comprehensive review of the related literature is provided in Appendix~\ref{appendix_extended_rw}.}

\section{Methodology}
\label{method}

Given an unlabeled multivariate time series $X = \{x_1, x_2, \dots, x_T\}\in \mathbb{R}^{T \times C}$, where $T$ denotes the sequence length and $C$ represents the number of feature channels, the objective of TSAD is to assign an anomaly score $\textit{AS}(x_t) \in [0, 1]$ to each observation, with higher scores reflecting greater anomalous probabilities.

\paragraph{Cascaded Time Series Anomaly Detection.} 
Given a pre-trained, high-capacity and complex anomaly detection model CDM along with its inferred anomaly scores $\textit{AS}^\textit{~train}_\textit{c}$ on the training set $X^\textit{train}$,
we introduce a lightweight pre-filter model LPM, which is trained using $X^\textit{train}$ and guided by the prior knowledge $\textit{AS}^\textit{~train}_\textit{c}$.
 The objective of LPM is to generate a coarse-grained anomaly score $\textit{AS}_l(x_t)\in [0, 1]$ and estimate a normality confidence score $P(x_t) \in [0, 1]$ for any observation $x_t \in X^\textit{test}$.
 
During inference, the evaluation path of $x_t$ is dynamically routed based on $P(x_t)$. A high $P(x_t)$ indicates a high-confidence normal sample, allowing the final anomaly score $\textit{AS}(x_t)$ to be efficiently derived directly from $\textit{AS}_l$. Conversely, uncertain samples are forwarded to CDM for a refined, albeit computationally heavier, evaluation.

\subsection{Overall Architecture of CARE}

\begin{figure}
  \centering
  \includegraphics[width=1\textwidth]{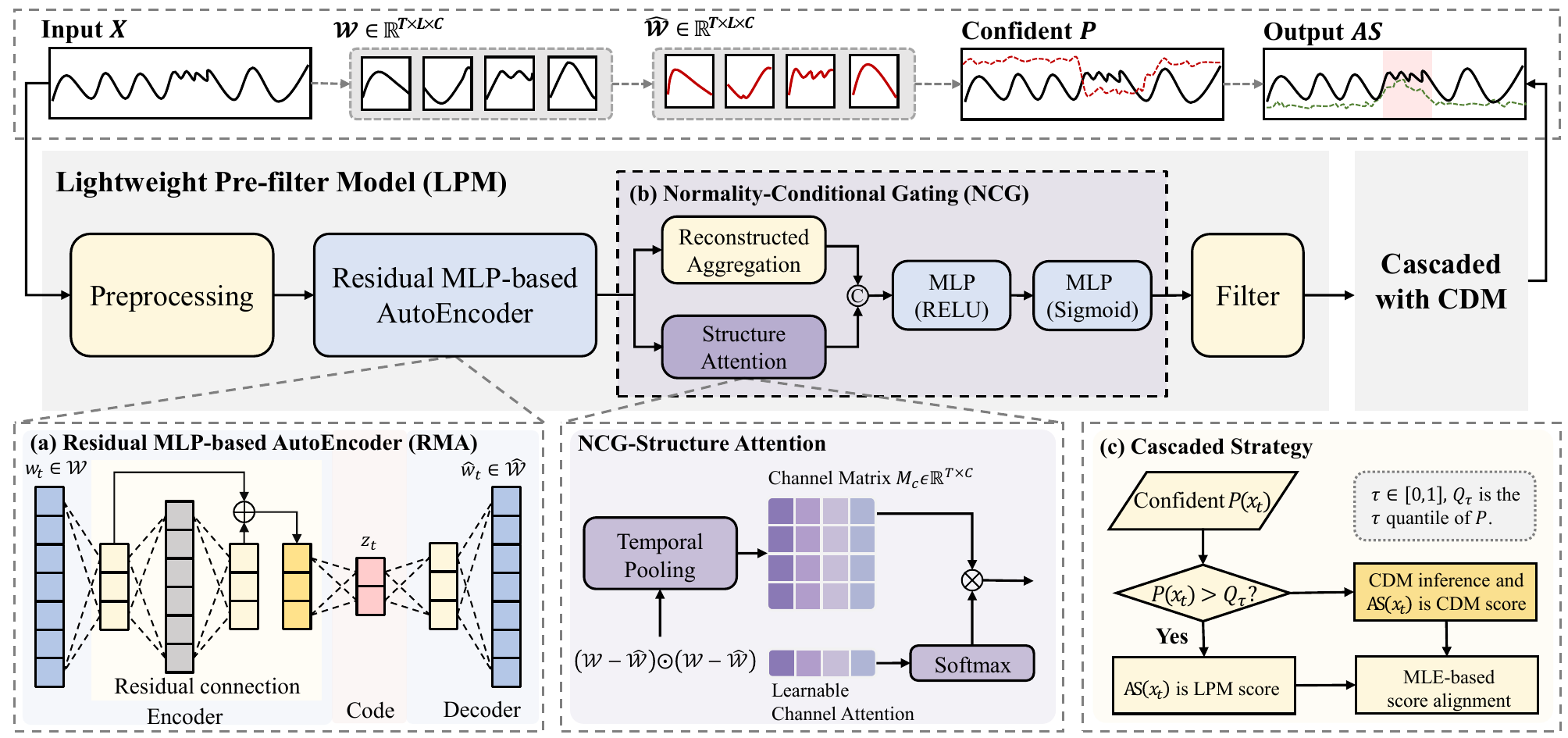}
  \caption{The overall architecture of the proposed CARE framework. (a) The Residual AutoEncoder performs lightweight and rapid input reconstruction. (b) The Normality-Conditioned Gating network fuses aggregated reconstruction deviations with structural features (channel-wise anomaly) to dynamically estimate a normality confidence score for each window.  
   (c)The Cascaded Inference Strategy leverages the confidence score to safely bypass trivial normal samples, selectively invoking the CDM only for hard anomalies to guarantee both inference efficiency and detection reliability.}
  \label{Figure2}
\end{figure}

The proposed CARE framework cascades a Lightweight Pre-filter Model (LPM) with a Complex Detection Model (CDM). The LPM rapidly filters predictable normal samples and routes challenging instances to the CDM, significantly enhancing efficiency without compromising detection quality. Since CARE aims to accelerate existing high-capacity detectors, it only assumes the backend CDM provides deterministic continuous anomaly scores, a standard property of most TSAD models. Consequently, we focus on the architectural design of the LPM, which consists of three principal components as illustrated in Figure~\ref{Figure2}:

\paragraph{Time Series Preprocessing.} We apply standard Z-score normalization to mitigate distribution shifts \citep{00030CB0W25}. The normalized sequence is then partitioned into fixed-length sliding windows (using prefix padding for initial steps) to serve as model inputs. Specifically, the input at time step $t$ is formulated as $w_t \in \mathbb{R}^{L \times C}$, encompassing the most recent $L$ observations of the time series. This yields the complete window set $\mathcal{W} = \{w_1, \dots, w_T\} \in \mathbb{R}^{T \times L \times C}$ for all $T$ time steps.

\paragraph{Residual MLP-based AutoEncoder (RMA).} To enable efficient anomaly inference, LPM adopts an MLP-based structure for time series reconstruction. The preprocessed windows $\mathcal{W} = \{w_1, \dots, w_T\}$ are fed into the encoder to capture underlying normal patterns, where residual connections are introduced to enhance feature representation. The resulting latent embeddings are subsequently passed to a decoder to reconstruct the original sequence.

\paragraph{Normality-Conditioned Gating (NCG).}
NCG is designed to rapidly identify and filter out trivial, unequivocally normal samples, thereby preventing them from undergoing complex inference.
To achieve this, we incorporate a confidence gating network based on a structure attention mechanism.
This module extracts multi-granularity  reconstruction errors to capture the deviations of the current observation from established normal patterns.
Subsequently, these multi-granularity deviation features are concatenated with the structural anomaly features to form a comprehensive joint representation. 
Ultimately, this fused representation is projected into a normalized confidence score through the gating network, which directly dictates the downstream routing decision in the cascaded framework.

\subsection{Residual MLP-based AutoEncoder (RMA)}
\label{AutoEncoder}

To enable rapid and computationally efficient anomaly inference, the RMA adopts a pure Multilayer Perceptron (MLP) autoencoder architecture, deliberately avoiding computationally expensive self-attention mechanisms. 
Specifically, the encoder maps flattened sliding windows through an initial MLP layer, followed by a residual block to effectively capture complex temporal dynamics without heavy parameterization. The decoder then reconstructs the sequence to obtain the reconstructed window set $\hat{\mathcal{W}} \in \mathbb{R}^{T \times L \times C}$ via two consecutive MLP layers and a reshaping operation. The formal expression of RMA is detailed in Appendix~\ref{appendix_A}.

\subsection{Normality-Conditioned Gating (NCG)}
As the core decision component of the LPM, the Normality-Conditioned Gating (NCG)  identifies high-confidence normal samples using the reconstruction outputs from Section~\ref{AutoEncoder}. 
To guarantee reliable routing, NCG introduces a structure attention mechanism as its key module that explicitly models channel-wise anomaly contributions,  thereby enhancing decision reliability.

\paragraph{Structure Attention.} 
In practice, sensor channels exhibit varying sensitivities for anomaly detection, with certain channels being inherently more fault-prone~\citep{TuliCJ22}. However, 
standard MLP-based feature fusion typically performs implicit and uniform aggregation, failing to explicitly capture this channel-specific importance.
To address this, CARE introduces a structure attention mechanism that explicitly quantifies channel-wise anomaly contributions.

Specifically, to extract the anomaly severity per channel, we compute the squared reconstruction deviation for each window and apply temporal average pooling as a channel matrix, which quantifies the anomaly contribution of each individual channel for a given sample. 
 To adaptively scale these contributions, we introduce a learnable \textbf{channel attention vector} $\mathbf{v} \in \mathbb{R}^{C}$, normalized via a softmax function.
 Finally, the channel-wise deviations are then weighted and 
 aggregated using this attention vector to yield the \textbf{structural anomaly feature} $\mathbf{f}_\text{st}$, which serves as a vital auxiliary signal for the subsequent gating network. Formally, this process is defined as:
\begin{equation}
\begin{aligned}
\mathbf{f}_\text{st} = \operatorname{TemporalPool}( (\mathcal{W}-\hat{\mathcal{W}}) \odot (\mathcal{W}-\hat{\mathcal{W}})) \cdot \operatorname{softmax}(\mathbf{v}),
\end{aligned}
\end{equation}
where $\mathbf{f}_\text{st} \in \mathbb{R}^{T \times 1}$ is the structural anomaly feature,  $\odot$ denotes the Hadamard (element-wise) product, and $\operatorname{TemporalPool}(\cdot)$ denotes the average pooling operation along the window length dimension $L$ (the temporal axis).  

\paragraph{Confidence Gating Network} 
In the final stage of the LPM, we design an MLP-based gating network to estimate the normality confidence of each sample.
To capture reconstruction quality at multiple granularities, we extract both window-level ($\mathbf{f}_\text{win}$) and point-level ($\mathbf{f}_\text{pot}$) features by applying global and channel-wise average pooling over the squared reconstruction errors, respectively. 
These features are then concatenated with the structural anomaly feature $\mathbf{f}_\text{st}$ to form a comprehensive joint representation. Finally, a two-layer MLP projects this fused representation into a normalized confidence score $\mathbf{P}$, which directly dictates whether the sample requires further refined evaluation by the CDM. The process can be formally  defined as follows:
\begin{equation}
\begin{aligned}
 & \mathbf{f}_\text{win}=\operatorname{WindowPool}((\mathcal{W}-\hat{\mathcal{W}}) \odot (\mathcal{W}-\hat{\mathcal{W}})) \\
 & \mathbf{f}_\text{pot}=\operatorname{ChannelPool}((\mathcal{W}_{:,L,:}-\hat{\mathcal{W}}_{:,L,:}) \odot (\mathcal{W}_{:,L,:}-\hat{\mathcal{W}}_{:,L,:})) \\
 & \mathbf{F} = \operatorname{Concat}(\mathbf{f}_\text{win}, \mathbf{f}_\text{pot},  \mathbf{f}_\text{st}) \\
 &\mathbf{P} =1-\mathrm{Sigmoid}((\mathrm{ReLU}(\mathbf{F}\mathbf{W}_1^{(g)}+\mathbf{b}_1^{(g)})\mathbf{W}_2^{(g)})+\mathbf{b}_2^{(g)})\\
\end{aligned}
\end{equation}
where $\mathbf{W}_1^{(g)}\in \mathbb{R}^{3 \times D_g}$, $\mathbf{W}_2^{(g)}\in \mathbb{R}^{D_g \times 1}$ are learnable weight matrices, and $\mathbf{b}_1^{(g)}\in \mathbb{R}^{D_g}$ and $\mathbf{b}_2^{(g)}\in \mathbb{R}$ are their corresponding bias terms. $D_g$ denotes the hidden dimension size. 
$\mathbf{f}_\text{win}, \mathbf{f}_\text{pot} \in \mathbb{R}^{T \times 1}$ represent the window-level and point-level reconstruction features, $\mathbf{F} \in \mathbb{R}^{T \times 3}$ is the joint representation, and $\mathbf{P} \in \mathbb{R}^{T \times 1}$ is the ultimate confidence score. 
$\operatorname{WindowPool}(\cdot)$ denotes the global average pooling across both the temporal and channel dimensions, while $\operatorname{ChannelPool}(\cdot)$ computes the average over channels at any time step.

\subsection{Cascaded Model Training and Inference}
This section details how the aforementioned Lightweight Pre-filter Model (LPM) is trained and the efficient cascaded inference process of  CARE framework. The corresponding pseudocode of CARE is summarized in Appendix~\ref{algorithm}.

\paragraph{Two-stage LPM Training.} 

To prevent the gating objective from distorting the underlying temporal representations (i.e., semantic drift), we decouple the optimization process into a two-stage training strategy. 

In Stage 1, the RMA is optimized only with the  sequence reconstruction loss. This guarantees that the RMA focuses exclusively on capturing intrinsic normal temporal patterns, independent of any downstream routing decisions. The loss function is formulated as:
\begin{equation}
\mathcal{L}_\textit{recon} = \lVert\mathcal{W} - \hat{\mathcal{W}}\rVert_{2}^{2} 
\label{equation:4}
\end{equation}

In Stage 2, both the RMA (trained in Stage 1) and the pre-trained CDM are frozen to provide stable prior knowledge. The NCG is then optimized to partition the input space into high-confidence and uncertain subsets.

To train the gating network, we utilize the anomaly scores inferred by the CDM on the training set ($\textit{AS}_c$) as supervisory signals. A fundamental challenge in unsupervised TSAD, however, is that raw anomaly scores lack inherent physical interpretability and exhibit severe scale variations across datasets. 
 Consequently, directly optimizing the network NCG against absolute $\textit{AS}_c$ values is prone to yield unreliable routing decisions.
To circumvent this, we exploit the robust relative ranking preserved within $\textit{AS}_c$. Guided by the intrinsic prior that normal patterns dominate the data distribution, we introduce a quantile-based risk threshold to robustly partition the training instances. Specifically, samples with an $\textit{AS}_c$ falling below this quantile are designated as high-confidence safe instances, whereas those exceeding it are strictly identified as high-risk.
The gating optimization is further formalized as a Quantile-Anchored Margin Loss, composed of the following two principal components:

\paragraph{Bipartite Loss.} This component explicitly enforces a margin-based separation, ensuring that designated safe samples consistently yield higher normality confidences than their high-risk counterparts. By penalizing ranking violations, it robustly preserves the global relative ordering between the two disjoint subsets. Formally, it is defined as:
\begin{equation}
\mathcal{L}_\textit{bipartite} = \mathbb{E}_{x_i, x_j \sim X^{\textit{train}}} \Big[ \mathbb{I}_\text{risk}(x_i) (1 - \mathbb{I}_\text{risk}(x_j)) \cdot \max \big(0, \xi_{1} - (P(x_j) - P(x_i)) \big) \Big]
\label{eq:loss_bipartite}
\end{equation}
where  $\mathbb{I}_\text{risk}(x_t) = \mathbb{I}(\textit{AS}_{c}(x_t) > Q_\text{risk})$ denotes a binary indicator function, and $Q_{\text{risk}}$ represents the predefined risk quantile threshold established over the $\textit{AS}_{c}$ distribution, and $\xi_1$ is a predefined margin hyperparameter enforcing the separation between the two subsets.

\paragraph{Boundary Loss.} To resolve ambiguity near the decision threshold, we enforce localized hinge constraints on samples immediately adjacent to the risk boundary. This explicitly penalizes incorrect local orderings within the most uncertain data region. Formally, it is expressed as:
\begin{equation}
\mathcal{L}_\textit{boundary} = \mathbb{E}_{x_i, x_j \sim X^\textit{train}} \Big[ \mathbb{I}_\text{risk}^\text{high}(x_i)\mathbb{I}_\text{risk}^\text{low}(x_j)  \cdot \max \big(0, \xi_{2} - (P(x_j) - P(x_i)) \big) \Big]
\label{eq:loss_boundary}
\end{equation}
where $\mathbb{I}_{\text{risk}}^{\text{high}}(x_t) = \mathbb{I}\big(Q_{\text{risk}} < \textit{AS}_c(x_t) \le Q_{\text{risk}}+w\big)$ and $\mathbb{I}_{\text{risk}}^{\text{low}}(x_t) = \mathbb{I}\big(Q_{\text{risk}}-w \le \textit{AS}_c(x_t) \le Q_{\text{risk}}\big)$ act as binary indicators capturing instances strictly above and below the boundary, respectively, with $w$ denoting a predefined margin width, and $\xi_2$ serving as the predefined margin hyperparameter for this localized constraint.

To further stabilize the optimization and encourage the network to output high confidences unless restricted by the ranking margins, we incorporate a penalty regularization $(1 - P(x_{t}))^2$. Combining these components, the overall training objective for the gating network is formalized as:
\begin{equation}
\mathcal{L}_\textit{gate} = \mathcal{L}_\textit{bipartite} + \lambda \mathcal{L}_\textit{boundary} + \alpha \mathbb{E}_{x_{t} \sim X^\textit{train}} \big[ (1 - P(x_{t}))^2 \big]
\label{eq:loss_gate_final}
\end{equation}

\paragraph{Cascaded CARE Inference.}
As illustrated in Figure~\ref{Figure2}, the raw time series $X$ is preprocessed into a set of sliding windows $\mathcal{W}$. For any observation $x_t$, its corresponding window $w_t$ is first evaluated by the LPM to rapidly generate a normality confidence score $P(x_t)$. Governed by a predefined filter threshold $\tau$, the framework dynamically partitions the input space into two mutually exclusive subsets: a high-confidence normal set $\mathcal{N}$ and an uncertain set $\mathcal{U}$. Specifically, if $P(x_t) \ge Q_{\tau}$, $x_t \in \mathcal{N}$ and safely bypasses the CDM; conversely, if $P(x_t) < Q_{\tau}$, $x_t \in \mathcal{U}$ and is routed to the CDM for refined evaluation, where $Q_{\tau}$ is the $\tau$ quantile of $P$.

\paragraph{MLE-based Score Alignment.} 
Since the raw anomaly scores from LPM ($\textit{AS}_l$) and CDM ($\textit{AS}_c$) reside in heterogeneous scale spaces, they cannot be directly unified. Motivated by the empirically observed right-skewed distributions of anomaly scores, we employ Maximum Likelihood Estimation (MLE) on the training set to fit log-normal distributions for both models: $\textit{AS}^\textit{~train}_{l} \sim \mathrm{LogNormal}(\mu_{l}, \sigma_{l}^2)$ and $\textit{AS}^\textit{~train}_{c} \sim \mathrm{LogNormal}(\mu_{c}, \sigma_{c}^2)$. During inference, we project the raw scores into a unified probabilistic domain using their respective cumulative distribution functions (CDFs, denoted as $F_l$ and $F_c$). The analysis of this score alignment strategy are provided in Appendix \ref{justification of mle}. The statistically aligned final score is:
\begin{equation}
\textit{AS}(x_t) =
\begin{cases}
-\log\big(1 - F_{l}(\textit{AS}_{l}(x_t))\big), & x_t \in \mathcal{N}, \\
-\log\big(1 - F_{c}(\textit{AS}_{c}(x_t))\big), & x_t \in \mathcal{U}.
\end{cases}
\end{equation}

\section{Experiments}
\label{experiments}

\paragraph{Datasets}
We evaluate the performance of the CARE framework on 8 real-world datasets. The datasets span  diverse application scenarios, such as spacecraft, server machines, and water treatment systems and include CalIt2~\citep{uci/cailt2}, NYC~\citep{ijcnn/nyc}, GECCO~\citep{moritz2018gecco}, MSL~\citep{HundmanCLCS18}, PSM~\citep{kdd/AbdulaalLL21/psm}, ECG~\citep{kdd/Yoon0L20/ecg}, SMAP~\citep{HundmanCLCS18} and KDD21~\citep{QiuLQHZWLGZSHJY25}. More details about the datasets  are provided in Appendix~\ref{datasets}.

\paragraph{Baselines} CARE is systematically compared with 14 baselines, covering representative methods and recent advances in TSAD. Specifically, these baselines include the classical methods: 
% LOF~\citep{BreunigKNS00}, PCA~\citep{shyu2003novel}, IForest (IF)~\citep{icdm/LiuTZ08},
OCSVM~\citep{nips/ScholkopfWSSP99}, Autoencoder (AE)~\citep{corr/abs-2109-09265}, LSTMED~\citep{SakuradaY14}; the recent representative deep learning methods: Anomaly Transformer (ATrans)~\citep{iclr/XuWWL22}, TranAD~\citep{TuliCJ22}, DCdetector (DC)~\citep{YangZZW023}, MLP-Mixer (Mixer)~\citep{TangDWZ24}, CATCH~\citep{00030CB0W25}, CrossAD (Cross)~\citep{nips/Beibu25}; the approaches adapted from general time series analysis: TimesNet (TsNet)~\citep{WuHLZ0L23}, iTransformer (iTrans)~\citep{LiuHZWWML24}, PatchTST (Patch)~\citep{iclr/NieNSK23}; and the large language model-based methods: GPT4TS (GS)~\citep{nips/ZhouNW0023}, CALF~\citep{aaai/LiuG0LBR0X25}. Among them, CrossAD, MLP-Mixer, and TranAD are regarded as representative state-of-the-art methods in TSAD.  

\paragraph{Setup} To demonstrate the effectiveness of the proposed CARE, we evaluate it from two perspectives: detection quality and inference efficiency. For detection quality, we adopt the label-based metric Affiliated-F1 (Aff-F) and the ranking-based metric Area Under the Precision-Recall Curve (A-P) as the primary evaluation metrics. For inference efficiency, we use average inference time (Time) as the core metric. In addition, we report comprehensive results for all metrics in Appendix~\ref{full results}. To ensure a strong and up-to-date backbone, we employ the SOTA CrossAD as the CDM. Notably, CARE is model-agnostic, and the CDM can be replaced with any existing method. Experimental results in Section~\ref{care generality describion} demonstrate this point. More implementation details are presented in the Appendix~\ref{implement}.

\subsection{Main Results} 
\label{Main Results}

We conducted a comprehensive evaluation of the proposed CARE on 8 datasets, comparing it against 14 baselines as shown in Table~\ref{tab:dataset_performance}. For detection quality, CARE achieves SOTA performance in terms of Aff-F1 and AUC-PR on 6 and 5 datasets, respectively. On the remaining datasets, it ranked second, with most deviating from the best by no more than 1\%. This result is mainly attributed to the structure attention-enhanced NCG, which identifies high-confidence normal samples early and avoids CDM-induced false positives. In terms of inference efficiency, CARE reached SOTA performance across all datasets. The substantial inference speedup is strongly correlated with the LPM's filtering ratio. As demonstrated by the latency reductions, the gating mechanism effectively reduces CDM invocations and lowers the overall inference cost. Overall, CARE achieves significant gains in inference efficiency while maintaining high detection quality.

\begin{table}[t]
\centering
\footnotesize
\caption{Average Aff-F (Affiliated-F1), A-P (AUC-PR) and Inference time  across 8 real-world datasets. The best results are in bold, while the second-best results are underlined.}
\renewcommand{\arraystretch}{1.3}
\resizebox{\textwidth}{!}{%
\begin{tabular}{c | c | c c c c c c c c c c c c c c c}

\noalign{\vskip +3ex}
\toprule
\textbf{Dataset} & \textbf{Metric} & \textbf{CARE} & \textbf{Cross} & \textbf{CATCH} & \textbf{Mixer} & \textbf{DC} & \textbf{TranAD} & \textbf{ATrans} & \textbf{CALF} & \textbf{GS} & \textbf{Patch} & \textbf{iTrans} & \textbf{TsNet} & \textbf{LSTM} & \textbf{AE} & \textbf{OCSVM} \\
\midrule
\multirow{3}{*}{\textbf{CalIt2}} & \textbf{Aff-F} & \underline{0.787} & \textbf{0.788} & 0.780 & 0.779 & 0.673 & 0.780 & 0.673 & 0.755 & 0.783 & 0.782 & 0.761 & 0.747 & 0.782 & 0.786 & \textbf{0.788} \\
                         & \textbf{A-P}   & \textbf{0.100} & \underline{0.099} & 0.094 & 0.097 & 0.029 & 0.052 & 0.030 & 0.048 & 0.087 & 0.093 & 0.094 & 0.079 & 0.078 & 0.081 & 0.095 \\
                         & \textbf{Time }  & \textbf{0.091} & 0.312 & 0.583 & 0.252 & 0.877 & 0.265 & 0.713 & 0.522 & 1.016 & \underline{0.213} & 0.222 & 0.572 & 0.474 & 0.472 & 0.259 \\
\multirow{3}{*}{\textbf{NYC}} & \textbf{Aff-F} & \textbf{0.694} & 0.675 & 0.668 & 0.664 & \underline{0.690} & 0.661 & 0.665 &                            0.657 & 0.689 & 0.658 & 0.673 & 0.662 & 0.670 & 0.688 & 0.667 \\
                         & \textbf{A-P}   & \textbf{0.090} & \underline{0.086} & 0.048 & 0.024 & 0.041 & 0.027 & 0.026 & 0.051 & 0.034 & 0.032 & 0.040 & 0.033 & 0.047 & 0.043 & 0.020 \\
                         & \textbf{Time}  & \textbf{0.184} & 0.501 & 0.914 & 0.762 & 1.938 & 0.441 & 1.074 & 0.829 & 2.058 & \underline{0.323} & 0.751 & 1.663 & 0.834 & 0.557 & 3.622 \\
\multirow{3}{*}{\textbf{GECCO}} & \textbf{Aff-F} & \textbf{0.938} & 0.884 & 0.863 & 0.379 & 0.669 & 0.166 & 0.669 &                            0.783 & 0.907 & 0.844 & 0.820 & \underline{0.908} & 0.351 & 0.392 & 0.468 \\
                         & \textbf{A-P}   & \textbf{0.624} & \underline{0.596} & 0.331 & 0.310 & 0.011 & 0.191 & 0.011 & 0.321 & 0.295 & 0.347 & 0.154 & 0.479 & 0.284 & 0.277 & 0.039 \\
                         & \textbf{Time}   & \textbf{3.889} & 11.98 & 23.64 & 15.20 & 22.79 & 9.520 & 17.03 & 15.22 & 26.01 & 6.121 & \underline{4.654} & 17.44 & 17.77 & 9.712 & 291.6 \\
\multirow{3}{*}{\textbf{MSL}} & \textbf{Aff-F} & \textbf{0.770} & 0.761 & \underline{0.769} & 0.768 & 0.679 & 0.763 & 0.679 &                            0.699 & 0.768 & 0.759 & 0.762 & 0.760 & 0.470 & 0.689 & 0.688 \\
                         & \textbf{A-P}   & \underline{0.266} & \textbf{0.272} & 0.228 & 0.247 & 0.108 & 0.188 & 0.098 & 0.178 & 0.225 & 0.201 & 0.213 & 0.207 & 0.193 & 0.144 & 0.153 \\
                         & \textbf{Time}   & \textbf{3.087} & 6.862 &  11.56 & 7.111  & 24.80  & 7.276 & 17.71 & 14.75 &  27.51  & 6.009 & \underline{4.292} & 18.31 & 17.56 & 10.73 & 114.7\\
\multirow{3}{*}{\textbf{ECG}} & \textbf{Aff-F} & \textbf{0.813} & \underline{0.800} & 0.773 & 0.786 & 0.728 & 0.663 & 0.728 &                            0.754 & 0.700 & 0.789 & 0.741 & 0.709 & 0.790 & 0.793 & 0.790 \\
                         & \textbf{A-P}   & \textbf{0.529} & 0.503 & \underline{0.505} & 0.495 & 0.147 & 0.272 & 0.157 & 0.486 & 0.506 & 0.490 & 0.483 & 0.412 & 0.500 & 0.478 & 0.480 \\
                         & \textbf{Time}   & \textbf{6.750} & 32.59 & 71.86 & 38.31 & 29.72 & 7.050 & 15.04 & 24.55 & 21.19 & 7.693 & \underline{6.823} & 14.12 & 14.18 & 8.786 & 145.5 \\
\multirow{3}{*}{\textbf{PSM}} & \textbf{Aff-F} & \underline{0.744} & \textbf{0.764} & 0.699 & 0.729 & 0.694 & 0.715 & 0.694 &                            0.694 & 0.694 & 0.696 & 0.699 & 0.705 & 0.693 & 0.694 & 0.693 \\
                         & \textbf{A-P}   & \textbf{0.569} & 0.465 & 0.381 & 0.438 & 0.278 & 0.380 & 0.290 & 0.344 & 0.379 & 0.370 & 0.377 & 0.394 & 0.433 & \underline{0.476} & 0.418 \\
                         & \textbf{Time}   & \textbf{9.057} & 30.56 & 72.60 & 35.70 & 36.81 & 13.50 & 24.06 & 32.78 & 33.33 & 10.86 & 
                         \underline{9.365} & 23.45 & 23.81 & 14.86 & 537.0 \\
\multirow{3}{*}{\textbf{SMAP}} & \textbf{Aff-F} & \textbf{0.721} & \underline{0.712} & 0.704 & 0.680 & 0.680 & 0.682 & 0.679 & 0.680 & 0.692 & 0.706 & 0.679 & 0.703 & 0.681 & 0.683 & 0.679 \\
                         & \textbf{A-P}  & 0.133 & 0.132 & 0.124 & 0.133 & \underline{0.136} & 0.110 & 0.124 & 0.109 & 0.124 & \textbf{0.140} & 0.114 & 0.134 & 0.103 & 0.109 & 0.101 \\
                         & \textbf{Time}   & \textbf{16.71}& 38.46 & 67.94 & 40.98 & 152.4 & 42.33 & 101.7 & 80.23 & 164.9 & 35.88 & \underline{22.67} & 108.6 & 187.6 & 155.7 & 1423 \\

\multirow{3}{*}{\textbf{KDD21}} & \textbf{Aff-F} & \textbf{0.716} & 0.712 & 0.693 & 0.714 & 0.669 & 0.714 & 0.695 & \underline{0.715} & 0.709 & 0.696 & 0.707 & 0.683 & 0.698 & 0.708 & 0.704 \\
                                & \textbf{A-P}  & \underline{0.047} & 0.044 & 0.042 & 0.046 & 0.009 & 0.033 & 0.040 & \textbf{0.058} & 0.023 & 0.025 & 0.042 & 0.021 & 0.023 & 0.030 & 0.028 \\
                                & \textbf{Time}   & \textbf{2.354} & 7.101 & 13.055 & 4.162 & 11.265 & 6.108 & 10.729 & 8.330 & 23.534 & 3.396 & \underline{2.957} & 18.519 & 9.861 & 8.409 & 38.392 \\
\midrule   

\multirow{3}{*}{\textbf{Avg}} & \textbf{Aff-F} & \textbf{0.781} & \underline{0.769} & 0.751 & 0.684 & 0.688 & 0.633 & 0.684 & 0.718 & 0.748 & 0.748 & 0.734 & 0.742 & 0.634 & 0.675 & 0.682 \\
                                & \textbf{A-P}  & \textbf{0.330} & \underline{0.308} & 0.245 & 0.249 & 0.107 & 0.174 & 0.105 & 0.219 & 0.236 & 0.239 & 0.211 & 0.248 & 0.234 & 0.230 & 0.187 \\
                                & \textbf{Time}  & \textbf{5.681} & 17.32 & 35.58 & 19.76 & 38.48 & 11.48 & 25.34 & 24.13 & 39.44 & 9.585 & \underline{6.967} & 26.31 & 37.46 & 28.68 & 359.4 \\
\bottomrule
\end{tabular}%
}
\label{tab:dataset_performance}
\end{table}

\subsection{Ablation Study} 
\label{Ablation Study}

\begin{table}[t]
\centering
\footnotesize
\caption{Ablation studies for CARE. Aff-F and A-P are the Affiliated-F1 and AUC-PR, higher is better. All variants are evaluated under the same CDM invocation rate (approximately 20\%) for fair comparison. The best values are highlighted in bold.}
\renewcommand{\arraystretch}{1.2}
\resizebox{\textwidth}{!}{
\begin{tabular}{
>{\centering\arraybackslash}p{2.3cm} 
>{\raggedright\arraybackslash}p{3.2cm}| 
>{\centering\arraybackslash}p{1cm} 
>{\centering\arraybackslash}p{1cm}| 
>{\centering\arraybackslash}p{1cm}  
>{\centering\arraybackslash}p{1cm}| 
>{\centering\arraybackslash}p{1cm} 
>{\centering\arraybackslash}p{1cm}| 
>{\centering\arraybackslash}p{1cm}  
>{\centering\arraybackslash}p{1cm}| 
>{\centering\arraybackslash}p{1cm} 
>{\centering\arraybackslash}p{1cm} 
}
\noalign{\vskip +3ex}
\toprule
\multicolumn{2}{c|}{\multirow{2}{*}{\textbf{Variations}}} & \multicolumn{2}{c|}{\textbf{NYC}} & \multicolumn{2}{c|}{\textbf{GECCO}} & \multicolumn{2}{c|}{\textbf{ECG}} & \multicolumn{2}{c|}{\textbf{PSM}} & \multicolumn{2}{c}{\textbf{Avg}} \\
\cmidrule{3-12}
 & & \textbf{Aff-F} & \textbf{A-P} & \textbf{Aff-F} & \textbf{A-P} & \textbf{Aff-F} & \textbf{A-P} & \textbf{Aff-F} & \textbf{A-P} & \textbf{Aff-F} & \textbf{A-P}\\
\midrule

\multirow{3}{*}{\shortstack[c]{\textbf{Model} \\ \textbf{Components}}} 
& w/o NCG (Random routing) & 0.658 & 0.056 & 0.873 & 0.329 & 0.804 & 0.509 & 0.713 & 0.514 & 0.762 & 0.352 \\
& w/o NCG (RMA routing) & 0.678 & 0.082 & 0.514 & 0.353 & 0.810 & 0.521 & 0.709 & 0.391 & 0.678 & 0.337 \\
& w/o structure attention & 0.662 & 0.055 & 0.937 & 0.614 & 0.803 & 0.497 & 0.708 & 0.390 & 0.778 & 0.389\\
\midrule

\multirow{3}{*}{\shortstack[c]{\textbf{Optimization} \\ \textbf{Objectives}}} 

& w/o $\mathcal{L}_{gate}$ Bipartite loss & 0.692 & 0.089 & 0.574 & 0.297 & 0.813 & 0.528 & 0.744 & 0.543 & 0.706 & 0.365\\
& w/o $\mathcal{L}_{gate}$ Boundary loss & 0.693 & 0.087 & 0.938 & 0.623 &0.812 & 0.523 & 0.709 & 0.380 & 0.788 & 0.404 \\
& w/o $\mathcal{L}_{gate}$ Regular Loss & 0.677 & 0.071 & 0.872 & 0.480 &0.809 & 0.521 & 0.740 & 0.517 & 0.775 & 0.398 \\
\midrule
\multicolumn{2}{c|}{\textbf{Training w/o Two-stage Strategy}} & 0.694 & 0.083 & 0.934 &0.621 & 0.799 & 0.517 & 0.715 & 0.411 & 0.786 & 0.408\\
\midrule
\multicolumn{2}{c|}{\textbf{CARE (ours)}} & \textbf{0.694} & \textbf{0.090} & \textbf{0.938} & \textbf{0.624} & \textbf{0.813} & \textbf{0.529} & \textbf{0.744} & \textbf{0.569} & \textbf{0.798} & \textbf{0.453} \\
\bottomrule
\end{tabular}
}
\label{tab:ablation_care}
\end{table}

Given that NCG is the core component of CARE, we replaced it with two alternative routing strategies, including random routing and RMA-based routing, under the same CDM invocation rate for a fair comparison. We further evaluate the contributions of structure attention, individual loss terms in $\mathcal{L}_{gate}$, and the two-stage training strategy. Based on the ablation results presented in Table~\ref{tab:ablation_care}, we derive the following observations: \textbf{First}, replacing NCG with random or RMA-based routing leads to consistent performance degradation, demonstrating the superiority of confidence-guided routing under the same inference budget. \textbf{Second}, removing structure attention significantly degrades detection quality, indicating its importance in capturing channel-wise anomaly contributions for gating.  \textbf{Third}, all three loss terms complement each other, enabling NCG to achieve optimal decision-making. Notably, the bipartite loss being the most critical for separating normal and uncertain samples. \textbf{Finally}, end-to-end training reduces detection quality, as RMA tends to learn features that favor gating decisions rather than capturing the underlying normal temporal patterns.

\begin{figure}[t]
    \centering
    \begin{subfigure}[b]{0.248\textwidth}
        \centering
        \includegraphics[width=\textwidth]{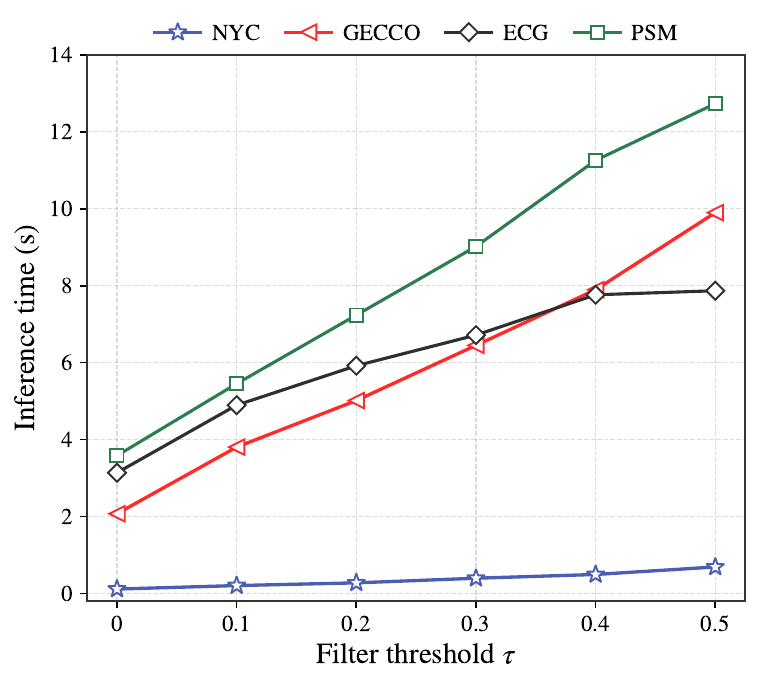}
        \caption{Inference time with $\tau$}
        \label{fig:tau_time}
    \end{subfigure}
    \hfill
    \begin{subfigure}[b]{0.24\textwidth}
        \centering
        \includegraphics[width=\textwidth]{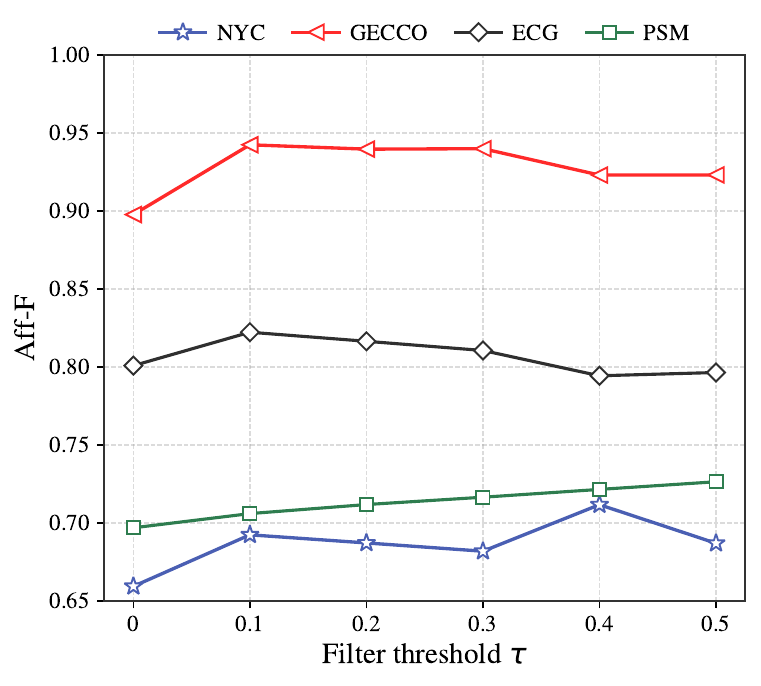}
        \caption{Affiliated-F1 with $\tau$}
        \label{fig:tau_f1}
    \end{subfigure}
    \hfill
    \begin{subfigure}[b]{0.24\textwidth}
        \centering
        \includegraphics[width=\textwidth]{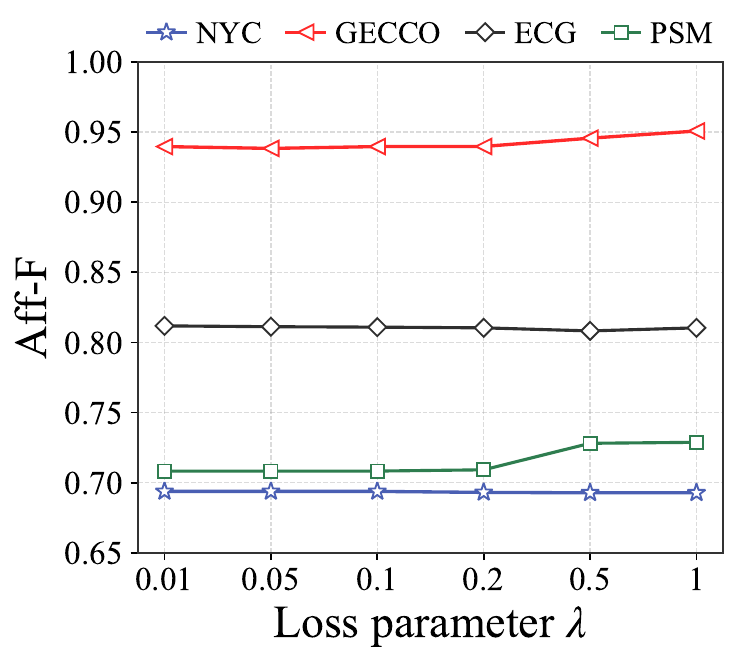}
        \caption{Loss parameter $\lambda$}
        \label{fig:lambda_f1}
    \end{subfigure}
    \hfill
    \begin{subfigure}[b]{0.24\textwidth}
        \centering
        \includegraphics[width=\textwidth]{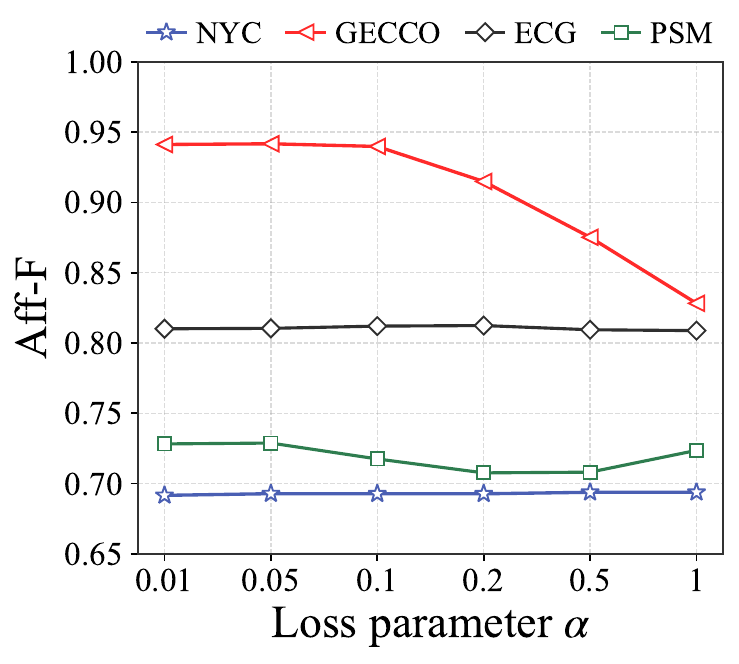}
        \caption{Loss parameter $\alpha$}
        \label{fig:alpha_f1}
    \end{subfigure}
        
    \caption{Parameter sensitivity studies of main hyper-parameters in CARE.}
    \label{fig:param_sensitivity}
\end{figure}

\begin{figure}[t]
    \centering
    \begin{subfigure}[b]{0.24\textwidth}
        \centering
        \includegraphics[width=\textwidth]{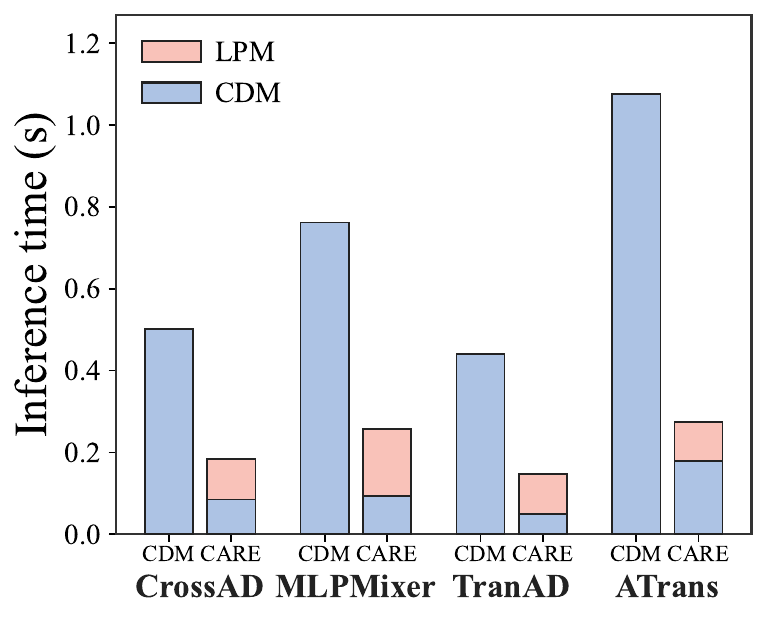}
        \caption{NYC}
        \label{fig:nyc_time}
    \end{subfigure}
    \hfill
    \begin{subfigure}[b]{0.24\textwidth}
        \centering
        \includegraphics[width=\textwidth]{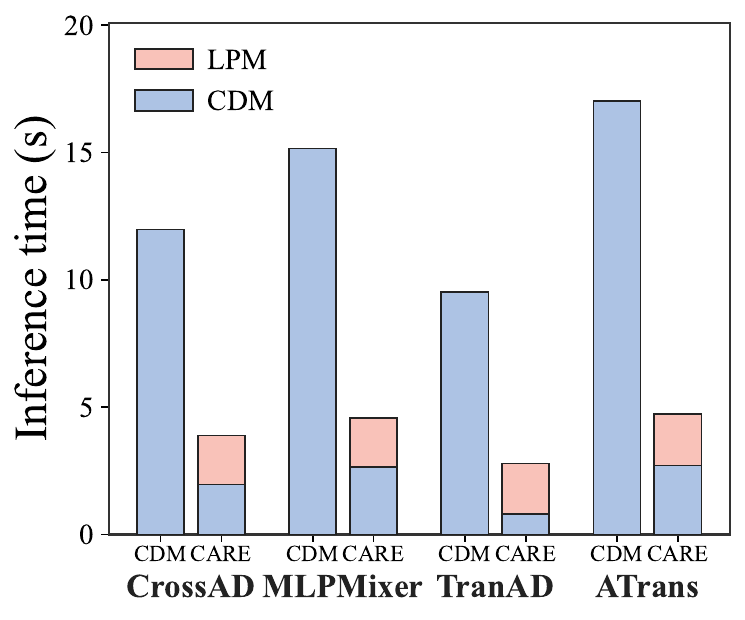}
        \caption{GECCO}
        \label{fig:gecco_time}
    \end{subfigure}
    \hfill
    \begin{subfigure}[b]{0.24\textwidth}
        \centering
        \includegraphics[width=\textwidth]{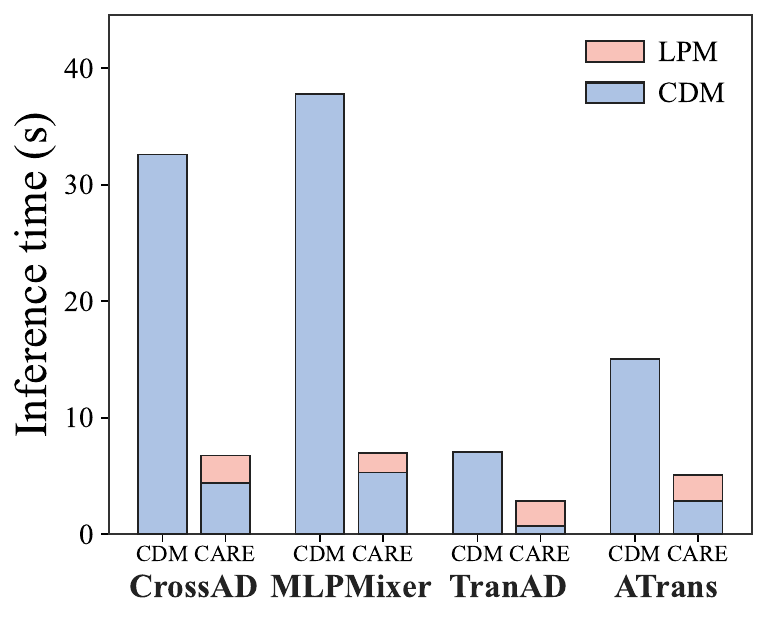}
        \caption{ECG}
        \label{fig:ecg_time}
    \end{subfigure}
    \hfill
    \begin{subfigure}[b]{0.24\textwidth}
        \centering
        \includegraphics[width=\textwidth]{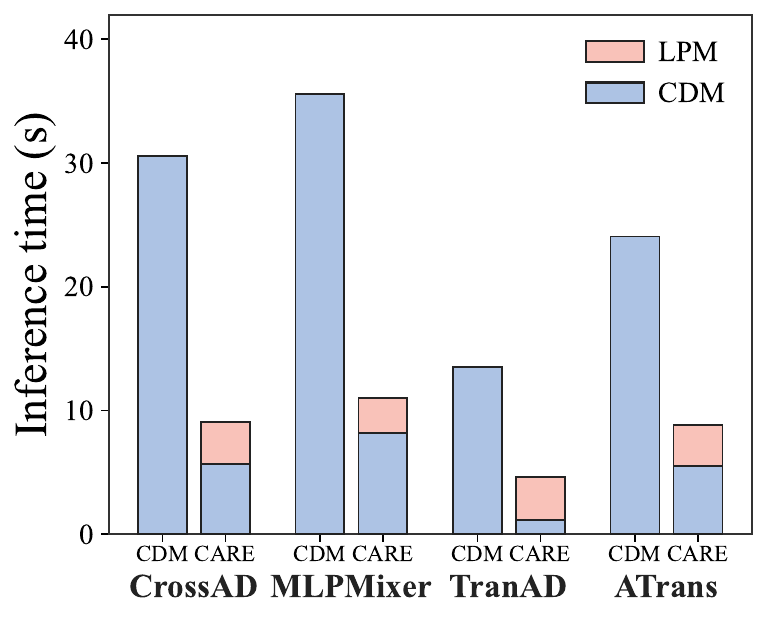}
        \caption{PSM}
        \label{fig:psm_time}
    \end{subfigure}
    \caption{Inference time comparison of CARE with various CDM backbones.}
    \label{fig:different CDM}
\end{figure}

\subsection{Parameter Sensitivity} 
\label{Parameter Sensitivity}
The filtering threshold $\tau$ is the most critical parameter, significantly affecting both inference efficiency and detection quality. As shown in Figure~\ref{fig:tau_time}, decreasing $\tau$ substantially improves inference efficiency by filtering out the majority of samples during the LPM stage, thereby reducing CDM invocations. However, both excessively high and low values of $\tau$ degrade detection quality. When $\tau$ is too large, excessive samples are routed to the CDM, weakening the cascaded mechanism and increasing false positives. A low $\tau$ causes CARE to effectively rely solely on LPM, making it difficult to identify hard samples, as illustrated in Figure~\ref{fig:tau_f1}. Overall, a $\tau$ value in the range of 0.1-0.3 strikes a balance between stability and efficiency. We also analyze $\lambda$ and $\alpha$ in the confidence-guided optimization function (Eq.(\ref{eq:loss_gate_final})) of CARE, which respectively control the influence of the boundary constraints and the regularization term on NCG decisions. As depicted in Figures~\ref{fig:lambda_f1} and~\ref{fig:alpha_f1}, $\lambda$ is stable and easily tunable in the range of 0.5-1, while $\alpha$ performs best within the range of 0.05-0.1. Additional hyperparameter analyses are provided in Appendix~\ref{additional sensitivity}.

\subsection{Adaptability of CARE to Diverse Detection Models} 
\label{care generality describion}

In this subsection, we demonstrate the versatility of the cascaded framework CARE by evaluating different existing models as the CDM. As shown in Table~\ref{tab:care generality}, we use CrossAD, MLPMixer, TranAD, and ATrans as backend models for CARE and compare their detection results with the original models. On three of four datasets, CARE outperforms the original models, achieving up to 20.9\% relative improvement in detection performance. 
This improvement suggests that the lightweight pre-filtering mechanism  effectively filters normal temporal patterns. By intercepting instances that complex models might otherwise over-interpret, CARE reduces false positives and potential over-fitting, thereby enhancing overall detection reliability.
Simultaneously, CARE significantly improves detection efficiency on all datasets, with a maximum speedup of 5.47x. Figure~\ref{fig:different CDM} illustrates this improvement by decomposing  the computational cost of the LPM and CDM stages: under the current settings, the minimal decision overhead of LPM substantially reduces the number of CDM invocations. These results validate the generality of CARE: it can be combined with diverse TSAD models to achieve efficient and reliable anomaly detection across different datasets and scenarios. Additional Adaptability analyses are provided in Appendix~\ref{additional adaptability}.

\begin{table}[htbp]
\centering
\footnotesize
\caption{The generality of CARE with different CDM models. Aff-F, Time, and $S\times$ are Affiliated F1-score, Inference time, and Speedup, respectively. The best ones are in bold.}
\renewcommand{\arraystretch}{1.2}
\resizebox{\textwidth}{!}{
\begin{tabular}{
>{\centering\arraybackslash}p{1cm} 
>{\raggedright\arraybackslash}p{1cm}| 
>{\centering\arraybackslash}p{0.8cm}  
>{\centering\arraybackslash}p{0.7cm}  
>{\centering\arraybackslash}p{0.7cm}|
>{\centering\arraybackslash}p{0.8cm}  
>{\centering\arraybackslash}p{0.7cm}  
>{\centering\arraybackslash}p{0.7cm}| 
>{\centering\arraybackslash}p{0.8cm}  
>{\centering\arraybackslash}p{0.7cm}  
>{\centering\arraybackslash}p{0.7cm}| 
>{\centering\arraybackslash}p{0.8cm}  
>{\centering\arraybackslash}p{0.7cm}  
>{\centering\arraybackslash}p{0.7cm}| 
>{\centering\arraybackslash}p{0.8cm}  
>{\centering\arraybackslash}p{0.7cm}  
>{\centering\arraybackslash}p{0.7cm} 
}
\noalign{\vskip +3ex}
\toprule
\multicolumn{2}{c|}{\multirow{2}{*}{\textbf{Method}}} & \multicolumn{3}{c|}{\textbf{NYC}} & \multicolumn{3}{c|}{\textbf{GECCO}} & \multicolumn{3}{c|}{\textbf{ECG}} & \multicolumn{3}{c|}{\textbf{PSM}} & \multicolumn{3}{c}{\textbf{Avg}} \\
\cmidrule{3-17}
 & & \textbf{Aff-F} & \textbf{Time} & \textbf{S$\times$} & \textbf{Aff-F} & \textbf{Time} & \textbf{S$\times$} & \textbf{Aff-F} & \textbf{Time} & \textbf{S$\times$} & \textbf{Aff-F} & \textbf{Time} & \textbf{S$\times$} & \textbf{Aff-F} & \textbf{Time} & \textbf{S$\times$}\\
\midrule
\multicolumn{2}{c|}{\textbf{CrossAD}}  & 0.675 & 0.501 & 1.000 & 0.884 & 11.98 & 1.000 & 0.800 & 32.59 & 1.000 & \textbf{0.764} & 30.56 & 1.000 & 0.781 & 18.91 & 1.000 \\
\multicolumn{2}{c|}{\textbf{CARE} (CrossAD as CDM)}  & \textbf{0.694} & \textbf{0.184} & \textbf{2.723} & \textbf{0.938} & \textbf{3.889} & \textbf{3.081} & \textbf{0.813} & \textbf{6.750} & \textbf{4.828} & 0.744 & \textbf{9.057} & \textbf{3.375} & \textbf{0.797} & \textbf{4.970} & \textbf{3.502}\\
\midrule
\multicolumn{2}{c|}{\textbf{MLPMixer}}  & 0.664 & 0.762 & 1.000 & 0.379 & 15.20 & 1.000 & 0.786 & 38.31 & 1.000 & \textbf{0.729} & 35.70 & 1.000 & 0.640 & 22.49 & 1.000\\
\multicolumn{2}{c|}{\textbf{CARE} (MLPMixer as CDM)} & \textbf{0.665} & \textbf{0.257} & \textbf{2.970} & \textbf{0.437} & \textbf{4.575} & \textbf{3.323} & \textbf{0.792} & \textbf{6.996} & \textbf{5.476} & 0.720 & \textbf{11.02} & \textbf{3.241} & \textbf{0.654} & \textbf{5.71}1 & \textbf{3.752} \\
\midrule
\multicolumn{2}{c|}{\textbf{TranAD}} & 0.661 & 0.441 & 1.000 & 0.166 & 9.520 & 1.000 & 0.663 & 7.050 & 1.000 & 0.715 & 13.50 & 1.000 & 0.551 & 7.627 & 1.000\\
\multicolumn{2}{c|}{\textbf{CARE} (TranAD as CDM)} & \textbf{0.663} & \textbf{0.147} & \textbf{3.001} & \textbf{0.166} & \textbf{2.789} & \textbf{3.414} & \textbf{0.801} & \textbf{2.840} & \textbf{2.483} & \textbf{0.729} & \textbf{4.626} & \textbf{2.918} & \textbf{0.590} & \textbf{2.600} & \textbf{2.954}\\
\midrule
\multicolumn{2}{c|}{\textbf{ATrans}} & 0.665 & 1.074 & 1.000 & 0.669 & 17.03 & 1.000 & 0.728 & 15.04 & 1.000 & 0.694 & 24.06 & 1.000 & 0.689 & 14.30 & 1.000\\

\multicolumn{2}{c|}{\textbf{CARE} (Atrans as CDM)} & \textbf{0.676} & \textbf{0.275} & \textbf{3.910} & \textbf{0.669} & \textbf{4.726} & \textbf{3.603} & \textbf{0.835} & \textbf{5.114} & \textbf{2.941} & \textbf{0.697} & \textbf{8.808} & \textbf{2.732} & \textbf{0.719} & \textbf{4.731} & \textbf{3.296} \\
\bottomrule
\end{tabular}
}
\label{tab:care generality}
\end{table}

\section{Conclusion}
\label{conclusion}

In this paper, we propose a novel cascaded inference framework, CARE, which accelerates inference efficiency while maintaining reliable time series anomaly detection. To this end, we propose a lightweight pre-filter model for efficiently learning normal patterns and gating decisions. A structure attention mechanism is introduced to enhance anomaly representation capabilities. A confidence-guided selective routing strategy is designed to separate high-confidence normal samples from uncertain samples, reducing complex model calls and overall computational cost. Extensive experiments on real-world datasets demonstrate the superiority and generality of CARE. As future work, we will investigate adaptive routing strategies for evolving time series environments, where continuously changing data distributions may require dynamic updates of routing decisions.

\bibliographystyle{unsrt}
\bibliography{references}

@article{DarbanWPAS25,
  title={Deep learning for time series anomaly detection: A survey},
  author={Zamanzadeh Darban, Zahra and Webb, Geoffrey I and Pan, Shirui and Aggarwal, Charu and Salehi, Mahsa},
  journal={ACM Computing Surveys},
  volume={57},
  number={1},
  pages={1--42},
  year={2024},
  publisher={ACM New York, NY}
}

@book{Hawkins80,
  author       = {D. M. Hawkins},
  title        = {Identification of Outliers},
  series       = {Monographs on Applied Probability and Statistics},
  publisher    = {Springer},
  year         = {1980},
  url          = {https://doi.org/10.1007/978-94-015-3994-4},
  doi          = {10.1007/978-94-015-3994-4},
  isbn         = {978-94-015-3996-8},
  bibsource    = {dblp computer science bibliography, https://dblp.org}
}

@article{Blazquez-Garcia21,
  title={A review on outlier/anomaly detection in time series data},
  author={Bl{\'a}zquez-Garc{\'\i}a, Ane and Conde, Angel and Mori, Usue and Lozano, Jose A},
  journal={ACM computing surveys (CSUR)},
  volume={54},
  number={3},
  pages={1--33},
  year={2021},
  publisher={ACM New York, NY, USA}
}

@inproceedings{YuXCZJZ24,
  title={Credit card fraud detection using advanced transformer model},
  author={Yu, Chang and Xu, Yongshun and Cao, Jin and Zhang, Ye and Jin, Yixin and Zhu, Mengran},
  booktitle={2024 IEEE international conference on metaverse computing, networking, and applications (MetaCom)},
  pages={343--350},
  year={2024},
  organization={IEEE}
}

@article{RoyMHB23,
  title={ECG-NET: A deep LSTM autoencoder for detecting anomalous ECG},
  author={Roy, Moumita and Majumder, Sukanta and Halder, Anindya and Biswas, Utpal},
  journal={Engineering Applications of Artificial Intelligence},
  volume={124},
  pages={106484},
  year={2023},
  publisher={Elsevier}
}

@article{DouYJQR25,
  title={Anomaly detection in event-triggered traffic time series via similarity learning},
  author={Dou, Shaoyu and Yang, Kai and Jiao, Yang and Qiu, Chengbo and Ren, Kui},
  journal={IEEE Transactions on Dependable and Secure Computing},
  volume={22},
  number={2},
  pages={888--902},
  year={2024},
  publisher={IEEE}
}

@inproceedings{BreunigKNS00,
  title={LOF: identifying density-based local outliers},
  author={Breunig, Markus M and Kriegel, Hans-Peter and Ng, Raymond T and Sander, J{\"o}rg},
  booktitle={Proceedings of the 2000 ACM SIGMOD international conference on Management of data},
  pages={93--104},
  year={2000}
}

@inproceedings{icdm/LiuTZ08,
  title={Isolation forest},
  author={Liu, Fei Tony and Ting, Kai Ming and Zhou, Zhi-Hua},
  booktitle={2008 eighth ieee international conference on data mining},
  pages={413--422},
  year={2008},
  organization={IEEE}
}

@article{nips/ScholkopfWSSP99,
  title={Support vector method for novelty detection},
  author={Sch{\"o}lkopf, Bernhard and Williamson, Robert C and Smola, Alex and Shawe-Taylor, John and Platt, John},
  journal={Advances in neural information processing systems},
  volume={12},
  year={1999}
}

@inproceedings{00030CB0W25,
  title={TSINR: capturing temporal continuity via implicit neural representations for time series anomaly detection},
  author={Li, Mengxuan and Liu, Ke and Chen, Hongyang and Bu, Jiajun and Wang, Hongwei and Wang, Haishuai},
  booktitle={Proceedings of the 31st ACM SIGKDD Conference on Knowledge Discovery and Data Mining V. 1},
  pages={671--682},
  year={2025}
}

@article{nips/DaiHYL24,
  title={Sarad: Spatial association-aware anomaly detection and diagnosis for multivariate time series},
  author={Dai, Zhihao and He, Ligang and Yang, Shuang-Hua and Leeke, Matthew},
  journal={Advances in Neural Information Processing Systems},
  volume={37},
  pages={48371--48410},
  year={2024}
}

@article{nips/Beibu25,
  title={CrossAD: Time series anomaly detection with cross-scale associations and cross-window modeling},
  author={Li, Beibu and Shentu, Qichao and Shu, Yang and Zhang, Hui and Li, Ming and Jin, Ning and Yang, Bin and Guo, Chenjuan},
  journal={arXiv preprint arXiv:2510.12489},
  year={2025}
}

@article{TangDWZ24,
  title={Mlp-mixer based masked autoencoders are effective, explainable and robust for time series anomaly detection},
  author={Tang, Qideng and Dai, Chaofan and Wu, Yahui and Zhou, Haohao},
  journal={Proceedings of the VLDB Endowment},
  volume={18},
  number={3},
  pages={798--811},
  year={2024},
  publisher={VLDB Endowment}
}

@article{pvldb/ZhuangZZGYWF24,
  title={Noise matters: Cross contrastive learning for flink anomaly detection},
  author={Zhuang, Zhihao and Zhang, Yingying and Zhao, Kai and Guo, Chenjuan and Yang, Bin and Wen, Qingsong and Fan, Lunting},
  journal={Proceedings of the VLDB Endowment},
  volume={18},
  number={4},
  pages={1159--1168},
  year={2024},
  publisher={VLDB Endowment}
}

@inproceedings{icde/FangXZ0G024,
  title={Temporal-frequency masked autoencoders for time series anomaly detection},
  author={Fang, Yuchen and Xie, Jiandong and Zhao, Yan and Chen, Lu and Gao, Yunjun and Zheng, Kai},
  booktitle={2024 IEEE 40th international conference on data engineering (ICDE)},
  pages={1228--1241},
  year={2024},
  organization={IEEE}
}

@article{WuQL0HGXY25,
  title={Catch: Channel-aware multivariate time series anomaly detection via frequency patching},
  author={Wu, Xingjian and Qiu, Xiangfei and Li, Zhengyu and Wang, Yihang and Hu, Jilin and Guo, Chenjuan and Xiong, Hui and Yang, Bin},
  journal={arXiv preprint arXiv:2410.12261},
  year={2024}
}

@article{QiuLQHZWLGZSHJY25,
  title        = {Tab: Unified benchmarking of time series anomaly detection methods},
  author={Qiu, Xiangfei and Li, Zhe and Qiu, Wanghui and Hu, Shiyan and Zhou, Lekui and Wu, Xingjian and Li, Zhengyu and Guo, Chenjuan and Zhou, Aoying and Sheng, Zhenli and others},
  journal={Proceedings of the VLDB Endowment},
  volume       = {18},
  number       = {9},
  pages        = {2775--2789},
  year         = {2025},
  publisher={VLDB Endowment}
}

@incollection{Aggarwal2017,
  title={An introduction to outlier analysis},
  author={Aggarwal, Charu C},
  booktitle={Outlier analysis},
  pages={1--34},
  year={2016},
  publisher={Springer}
}

@article{SchmidlWP22,
  title={Anomaly detection in time series: a comprehensive evaluation},
  author={Schmidl, Sebastian and Wenig, Phillip and Papenbrock, Thorsten},
  journal={Proceedings of the VLDB Endowment},
  volume       = {15},
  number       = {9},
  pages        = {1779--1797},
  year         = {2022},
  publisher={VLDB Endowment}
}

@article{LiuHZWWML24,
  title={itransformer: Inverted transformers are effective for time series forecasting},
  author={Liu, Yong and Hu, Tengge and Zhang, Haoran and Wu, Haixu and Wang, Shiyu and Ma, Lintao and Long, Mingsheng},
  journal={The Twelfth International Conference on Learning Representations (ICLR)},
  year={2024}
}

@article{nips/ZhouNW0023,
  title={One fits all: Power general time series analysis by pretrained lm},
  author={Zhou, Tian and Niu, Peisong and Sun, Liang and Jin, Rong and others},
  journal={Advances in neural information processing systems},
  volume={36},
  pages={43322--43355},
  year={2023}
}

@article{pami/HanHSYWW22,
  title={Dynamic neural networks: A survey},
  author={Han, Yizeng and Huang, Gao and Song, Shiji and Yang, Le and Wang, Honghui and Wang, Yulin},
  journal={IEEE transactions on pattern analysis and machine intelligence},
  volume={44},
  number={11},
  pages={7436--7456},
  year={2021},
  publisher={IEEE}
}

@article{tmlr/ChenZ024,
  title={Frugalgpt: How to use large language models while reducing cost and improving performance},
  author={Chen, Lingjiao and Zaharia, Matei and Zou, James},
  journal      = {Trans. Mach. Learn. Res.},
  year         = {2024}
}

@article{TuliCJ22,
  title={Tranad: Deep transformer networks for anomaly detection in multivariate time series data},
  author={Tuli, Shreshth and Casale, Giuliano and Jennings, Nicholas R},
  journal      = {Proc. {VLDB} Endow.},
  volume       = {15},
  number       = {6},
  pages        = {1201--1214},
  year         = {2022},
  publisher={VLDB Endowment}
}

@misc{uci/cailt2,
  author = {Arthur Asuncion and David Newman},
  title = {UCI Machine Learning Repository},
  booktitle = {University of California, Irvine, School of Information and Computer Sciences},
  year = {2007}
}

@inproceedings{ijcnn/nyc,
  title={A comparative study of HTM and other neural network models for online sequence learning with streaming data},
  author={Cui, Yuwei and Surpur, Chetan and Ahmad, Subutai and Hawkins, Jeff},
  booktitle={2016 International joint conference on neural networks (IJCNN)},
  pages={1530--1538},
  year={2016},
  organization={IEEE}
}

@article{moritz2018gecco,
  title={GECCO industrial challenge 2018 dataset},
  author={Moritz, S and Rehbach, F and Chandrasekaran, S and Rebolledo, M and Bartz-Beielstein, T},
  journal={Tech. Rep.},
  year={2018}
}

@inproceedings{HundmanCLCS18,
  title={Detecting spacecraft anomalies using lstms and nonparametric dynamic thresholding},
  author={Hundman, Kyle and Constantinou, Valentino and Laporte, Christopher and Colwell, Ian and Soderstrom, Tom},
  booktitle={Proceedings of the 24th ACM SIGKDD international conference on knowledge discovery \& data mining},
  pages={387--395},
  year={2018}
}

@inproceedings{kdd/AbdulaalLL21/psm,
  title={Practical approach to asynchronous multivariate time series anomaly detection and localization},
  author={Abdulaal, Ahmed and Liu, Zhuanghua and Lancewicki, Tomer},
  booktitle={Proceedings of the 27th ACM SIGKDD conference on knowledge discovery \& data mining},
  pages={2485--2494},
  year={2021}
}

@inproceedings{kdd/Yoon0L20/ecg,
  title={Ultrafast local outlier detection from a data stream with stationary region skipping},
  author={Yoon, Susik and Lee, Jae-Gil and Lee, Byung Suk},
  booktitle={Proceedings of the 26th ACM SIGKDD international conference on knowledge discovery \& data mining},
  pages={1181--1191},
  year={2020}
}

@article{corr/abs-2109-09265,
  title={Merlion: A machine learning library for time series},
  author={Bhatnagar, Aadyot and Kassianik, Paul and Liu, Chenghao and Lan, Tian and Yang, Wenzhuo and Cassius, Rowan and Sahoo, Doyen and Arpit, Devansh and Subramanian, Sri and Woo, Gerald and others},
  journal={arXiv preprint arXiv:2109.09265},
  year={2021}
}

@inproceedings{SakuradaY14,
  title={Anomaly detection using autoencoders with nonlinear dimensionality reduction},
  author={Sakurada, Mayu and Yairi, Takehisa},
  booktitle={Proceedings of the MLSDA 2014 2nd workshop on machine learning for sensory data analysis},
  pages={4--11},
  year={2014}
}

@inproceedings{iclr/XuWWL22,
  title={Anomaly transformer: Time series anomaly detection with association discrepancy},
  author={Xu, Jiehui and Wu, Haixu and Wang, Jianmin and Long, Mingsheng},
  booktitle    = {The Tenth International Conference on Learning Representations (ICLR)},
  publisher    = {OpenReview.net},
  year         = {2022},
}

@inproceedings{YangZZW023,
  title={Dcdetector: Dual attention contrastive representation learning for time series anomaly detection},
  author={Yang, Yiyuan and Zhang, Chaoli and Zhou, Tian and Wen, Qingsong and Sun, Liang},
  booktitle={Proceedings of the 29th ACM SIGKDD conference on knowledge discovery and data mining},
  pages={3033--3045},
  year={2023}
}

@inproceedings{WuHLZ0L23,
  title={Timesnet: Temporal 2d-variation modeling for general time series analysis},
  author={Wu, Haixu and Hu, Tengge and Liu, Yong and Zhou, Hang and Wang, Jianmin and Long, Mingsheng},
  booktitle    = {The Eleventh International Conference on Learning Representations (ICLR)},
  year         = {2023},
  publisher    = {OpenReview.net}
}

@inproceedings{iclr/NieNSK23,
  title={A time series is worth 64 words: Long-term forecasting with transformers},
  author={Nie, Yuqi and Nguyen, Nam H and Sinthong, Phanwadee and Kalagnanam, Jayant},
  booktitle    = {The Eleventh International Conference on Learning Representations (ICLR)},
  publisher    = {OpenReview.net},
  year         = {2023},
}

@inproceedings{aaai/LiuG0LBR0X25,
  title={Calf: Aligning llms for time series forecasting via cross-modal fine-tuning},
  author={Liu, Peiyuan and Guo, Hang and Dai, Tao and Li, Naiqi and Bao, Jigang and Ren, Xudong and Jiang, Yong and Xia, Shu-Tao},
  booktitle={Proceedings of the AAAI Conference on Artificial Intelligence},
  volume={39},
  number={18},
  pages={18915--18923},
  year={2025}
}

@inproceedings{DengH21,
  title={Graph neural network-based anomaly detection in multivariate time series},
  author={Deng, Ailin and Hooi, Bryan},
  booktitle={Proceedings of the AAAI conference on artificial intelligence},
  volume={35},
  number={5},
  pages={4027--4035},
  year={2021}
}

@inproceedings{SiPLZLZDLXP23,
  title={Beyond sharing: Conflict-aware multivariate time series anomaly detection},
  author={Si, Haotian and Pei, Changhua and Li, Zhihan and Zhao, Yadong and Li, Jingjing and Zhang, Haiming and Diao, Zulong and Li, Jianhui and Xie, Gaogang and Pei, Dan},
  booktitle={Proceedings of the 31st ACM Joint European Software Engineering Conference and Symposium on the Foundations of Software Engineering},
  pages={1635--1645},
  year={2023}
}

@inproceedings{SuZNLSP19,
  title={Robust anomaly detection for multivariate time series through stochastic recurrent neural network},
  author={Su, Ya and Zhao, Youjian and Niu, Chenhao and Liu, Rong and Sun, Wei and Pei, Dan},
  booktitle={Proceedings of the 25th ACM SIGKDD international conference on knowledge discovery \& data mining},
  pages={2828--2837},
  year={2019}
}

@inproceedings{AudibertMGMZ20,
  title={Usad: Unsupervised anomaly detection on multivariate time series},
  author={Audibert, Julien and Michiardi, Pietro and Guyard, Fr{\'e}d{\'e}ric and Marti, S{\'e}bastien and Zuluaga, Maria A},
  booktitle={Proceedings of the 26th ACM SIGKDD international conference on knowledge discovery \& data mining},
  pages={3395--3404},
  year={2020}
}

@inproceedings{aaai/ZengCZ023,
  title={Are transformers effective for time series forecasting?},
  author={Zeng, Ailing and Chen, Muxi and Zhang, Lei and Xu, Qiang},
  booktitle={Proceedings of the AAAI conference on artificial intelligence},
  volume={37},
  number={9},
  pages={11121--11128},
  year={2023}
}

@article{nips/WangWDQZLQWL24,
  title={Timexer: Empowering transformers for time series forecasting with exogenous variables},
  author={Wang, Yuxuan and Wu, Haixu and Dong, Jiaxiang and Qin, Guo and Zhang, Haoran and Liu, Yong and Qiu, Yunzhong and Wang, Jianmin and Long, Mingsheng},
  journal={Advances in Neural Information Processing Systems},
  volume={37},
  pages={469--498},
  year={2024}
}

@article{iccv/ViolaJ01,
  title={Robust real-time face detection},
  author={Viola, Paul and Jones, Michael J},
  journal={International journal of computer vision},
  volume={57},
  number={2},
  pages={137--154},
  year={2004},
  publisher={Springer}
}

@inproceedings{cvpr/CaiV18,
  title={Cascade r-cnn: Delving into high quality object detection},
  author={Cai, Zhaowei and Vasconcelos, Nuno},
  booktitle={Proceedings of the IEEE conference on computer vision and pattern recognition},
  pages={6154--6162},
  year={2018}
}

@inproceedings{eccv/ZhangLTYZWB24,
  title={Make your vit-based multi-view 3d detectors faster via token compression},
  author={Zhang, Dingyuan and Liang, Dingkang and Tan, Zichang and Ye, Xiaoqing and Zhang, Cheng and Wang, Jingdong and Bai, Xiang},
  booktitle={European Conference on Computer Vision},
  pages={56--72},
  year={2024},
  organization={Springer}
}

@article{nips/ChenJLK024,
  title={Routerdc: Query-based router by dual contrastive learning for assembling large language models},
  author={Chen, Shuhao and Jiang, Weisen and Lin, Baijiong and Kwok, James and Zhang, Yu},
  journal={Advances in Neural Information Processing Systems},
  volume={37},
  pages={66305--66328},
  year={2024}
}

@article{corr/abs-2506-04203,
  title={Cascadia: A Cascade Serving System for Large Language Models},
  author={Jiang, Youhe and Fu, Fangcheng and Zhao, Wanru and Rabanser, Stephan and Lane, Nicholas D and Yuan, Binhang},
  journal={arXiv e-prints},
  pages={arXiv--2506},
  year={2025}
}

@article{corr/abs-2404-14294,
  title={A survey on efficient inference for large language models},
  author={Zhou, Zixuan and Ning, Xuefei and Hong, Ke and Fu, Tianyu and Xu, Jiaming and Li, Shiyao and Lou, Yuming and Wang, Luning and Yuan, Zhihang and Li, Xiuhong and others},
  journal={arXiv preprint arXiv:2404.14294},
  year={2024}
}

@inproceedings{aaai/ChenZZ0DT24,
  title={Data shunt: Collaboration of small and large models for lower costs and better performance},
  author={Chen, Dong and Zhuang, Yueting and Zhang, Shuo and Liu, Jinfeng and Dong, Su and Tang, Siliang},
  booktitle={Proceedings of the AAAI Conference on Artificial Intelligence},
  volume={38},
  number={10},
  pages={11249--11257},
  year={2024}
}

@article{shyu2003novel,
  title={A novel anomaly detection scheme based on principal component classifier},
  author={Shyu, Mei-Ling and Chen, Shu-Ching and Sarinnapakorn, Kanoksri and Chang, LiWu},
  year={2003}
}

%%%%%%%%%%%%%%%%%%%%%%%%%%%%%%%%%%%%%%%%%%%%%%%%%%%%%%%%%%%%

\newpage
\appendix

\section{Extended Related Work}\label{appendix_extended_rw}

\subsection{Deep Time Series Anomaly Detection}

Deep learning-based TSAD utilizes neural networks to learn feature representations or anomaly scores, outperforming traditional methods in many complex scenarios~\citep{DarbanWPAS25}. Existing approaches can be broadly categorized into three paradigms. Forecasting-based methods~\citep{corr/abs-2109-09265, DengH21, SiPLZLZDLXP23} detect anomalies by measuring deviations between predicted and observed future steps. Reconstruction-based methods, predominantly relying on AutoEncoders and Transformers~\citep{TangDWZ24, icde/FangXZ0G024, SakuradaY14, SuZNLSP19, AudibertMGMZ20, iclr/XuWWL22}, identify anomalies through reconstruction errors, with recent advancements incorporating adversarial training~\citep{TuliCJ22} or channel fusion~\citep{00030CB0W25}. Alternatively, representation-based approaches~\citep{YangZZW023, WuHLZ0L23} focus on learning discriminative temporal representations in the latent space. Beyond dedicated TSAD models, general time series forecasting architectures~\citep{LiuHZWWML24, aaai/ZengCZ023, iclr/NieNSK23, nips/WangWDQZLQWL24} and recent pre-trained large language models~\citep{nips/ZhouNW0023, aaai/LiuG0LBR0X25} have also been adapted for anomaly detection.

Despite their architectural diversity, a common limitation unites these methodologies: a predominant focus on maximizing detection quality (e.g. accuracy or precision) through increasingly complex networks. This continuous pursuit of accuracy rarely incorporates explicit designs for inference efficiency, rendering them computationally expensive for latency-sensitive practical deployments.

\subsection{Cascaded Inference Models}

Cascaded inference models optimize the adaptive allocation of computational resources by integrating models of varying complexity \citep{pami/HanHSYWW22}. The fundamental principle is to process straightforward samples with lightweight, low-cost models, strategically reserving high-capacity architectures for complex cases. These models have been extensively explored across domains such as computer vision \citep{iccv/ViolaJ01,cvpr/CaiV18,eccv/ZhangLTYZWB24} and natural language processing \citep{tmlr/ChenZ024,nips/ChenJLK024,corr/abs-2506-04203,corr/abs-2404-14294}. The core mechanism typically involves leveraging the prediction confidence of a smaller model to dynamically route uncertain samples to a larger model \citep{aaai/ChenZZ0DT24}. This strategy significantly reduces inference latency without sacrificing overall performance.

In the domain of time series anomaly detection, anomalous events are inherently scarce, resulting in highly skewed data distributions dominated by normal patterns \citep{aggarwal2017, SchmidlWP22}. Consequently, the vast majority of normal samples can be accurately identified without invoking complex neural architectures. This intrinsic data characteristic naturally motivates our adoption of a cascaded inference framework to circumvent redundant computations, enabling highly efficient yet reliable anomaly detection.

\section{Formal Definition of RMA}\label{appendix_A}

This section presents a formal representation of the proposed Residual MLP Autoencoder (RMA), providing a structured complement to the preceding model description. Following its architectural design, we describe the model separately from the encoder and decoder perspectives. 

 Specifically, the encoding process can be formally stated as:
\begin{equation}\label{equation_rma}
\begin{split}
 \mathbf h & = \operatorname{GELU} (
(\operatorname{Flatten}(\mathcal{W}))\mathbf{W}_1^{(e)} + \mathbf{b}_1^{(e)})\\
 \mathbf r & =\operatorname{GELU}((\operatorname{GELU}(\mathbf h\mathbf{W}_2^{(e)}+\mathbf{b}_2^{(e)})\mathbf{W}_3^{(e)})+\mathbf{b}_3^{(e)})\\
 \mathbf{z} & =(\mathbf h+ \mathbf{r})\mathbf{W}_4^{(e)}+\mathbf{b}_4^{(e)}
\end{split}
\end{equation}
where $\mathbf{W}_1^{(e)} \in \mathbb{R}^{(LC) \times D_h}$, $\mathbf{W}_2^{(e)}\in \mathbb{R}^{D_h \times 2D_h}$, $\mathbf{W}_3^{(e)}\in \mathbb{R}^{2D_h \times D_h}$, and $\mathbf{W}_4^{(e)}\in \mathbb{R}^{D_h \times D_z}$. $\mathbf h, \mathbf r\in\mathbb{R}^{T \times D_h}$ and $\mathbf z\in\mathbb{R}^{T\times D_z}$ are the hidden representations. $D_h$ and $D_z$ denote the hidden dimensions of the encoder.

The decoding process can be formally expressed as:  
\begin{equation}\label{equation2}
\begin{split}
 \hat{\mathbf h} &= \operatorname{GELU}(\mathbf{z}\mathbf{W}_1^{(d)}+\mathbf{b}_1^{(d)})\\
 \hat{\mathcal{W}} &= \operatorname{Reshape}(\hat{\mathbf h}\mathbf{W}_2^{(d)} + \mathbf{b}_2^{(d)})\\
\end{split}
\end{equation}
where $\mathbf{W}_1^{(d)}\in \mathbb{R}^{D_z \times D_h}$, $\mathbf{W}_2^{(d)}\in \mathbb{R}^{D_h \times (LC)}$ are learnable weight matrices, and $\mathbf{b}_1^{(d)} \in \mathbb{R}^{D_h}$ and $\mathbf{b}_2^{(d)} \in \mathbb{R}^{LC}$ are their corresponding bias terms. $\hat{\mathbf h}\in\mathbb{R}^{T \times D_h}$ and $\hat{\mathcal{W}} \in\mathbb{R}^{T\times L \times C}$ denote the hidden representations of the decoder and the set of reconstruction windows, respectively.

\section{Detailed Algorithm of CARE}\label{algorithm}

\begin{algorithm}[h]
\caption{CARE - Overall Architecture.}
\begin{spacing}{1.2}
\begin{algorithmic}[1]
\Require Input time series $X = \{x_1, x_2, \dots, x_T\} \in \mathbb{R}^{T \times C}$; input length $T$; window size $L$; filter threshold $\tau$.
\Ensure Anomaly Score $\textit{AS}$.
\State $\triangleright$ Input time series is processed into a sliding window set.
\State $\mathcal{W} \leftarrow  \operatorname{Windowing}(\operatorname{ZScore}(X))$ \hfill $\triangleright~\mathcal{W} \in \mathbb{R}^{T \times L\times C}$

\State $\triangleright$ Residual MLP-based AutoEncoder is applied to reconstruct the window.

\State $\mathbf{z} \leftarrow  \operatorname{Encoder}(\mathcal{W})$ \hfill $\triangleright~\mathbf{z} \in \mathbb{R}^{T \times D_z}$

\State $\mathcal{\hat{W}} \leftarrow  \operatorname{Decoder
}(\mathbf{z})$ \hfill 
$\triangleright~\mathcal{\hat{W}} \in \mathbb{R}^{T \times L \times C}$

\State $\triangleright$ Normality-Conditioned Gating is utilized to generate confidence score.

\State $\mathbf{P},\textit{AS}_l\leftarrow \textit{Normality-Conditioned Gating}(\mathcal{W},\hat{\mathcal{W}})$

\State $\triangleright$ Filter samples based on threshold $\tau$.

\State $\mathcal{U}\leftarrow \textit{Filtering}(\mathbf{P}, X,\tau)$

\If{$\mathcal{U}$ is empty}
    \State $\textit{AS}\leftarrow \textit{AS}_l$
\Else
    \State $\triangleright$ Uncertain sample anomaly scores are calculated by CDM.
    \State $\textit{AS}_{c}\leftarrow\operatorname{CDM}(\mathcal{U})$ \hfill 
$\triangleright~\textit{$AS_c$} \in \mathbb{R}^{|\mathcal{U}| \times 1}$
\EndIf
\State $\triangleright$ Anomaly scores are aligned based on MLE.
\State $\textit{AS}\leftarrow \operatorname{Aligner}(\textit{AS}_{l},\textit{AS}_{c})$ \hfill 
$\triangleright~\textit{AS} \in \mathbb{R}^{T \times 1}$

\State \textbf{return }$\textit{AS}$ \hfill $\triangleright$ Return the anomaly score.

\noindent\rule[0.25\baselineskip]{0.96\textwidth}{0.4pt}
\Procedure{Normality-Conditioned Gating}{$\mathcal{W},\hat{\mathcal{W}}$} 

\State $\mathbf{e} \leftarrow  (\mathcal{W}-\hat{\mathcal{W}}) \odot (\mathcal{W}-\hat{\mathcal{W}}) $ \hfill 
$\triangleright~\mathbf{e} \in \mathbb{R}^{T \times L \times C}$

\State $\mathbf{f}_\text{st} \leftarrow \operatorname{TemporalPool}(\mathbf{e}) \cdot \operatorname{softmax}(\mathbf{v}^{\top})$ \hfill 
$\triangleright~\mathbf{f}_\text{st} \in \mathbb{R}^{T \times 1}$

\State $\mathbf{f}_\text{win} \leftarrow \operatorname{WindowPool}(\mathbf{e})$ \hfill 
$\triangleright~\mathbf{f}_\text{win} \in \mathbb{R}^{T \times 1}$

\State $\mathbf{f}_\text{pot} \leftarrow \operatorname{ChannelPool}(\mathbf{e}_{:,L,:})$ \hfill 
$\triangleright~\mathbf{f}_\text{pot} \in \mathbb{R}^{T \times 1}$

\State $\mathbf{F} \leftarrow \operatorname{Concat}( \mathbf{f}_\text{win}, \mathbf{f}_\text{pot},\mathbf{f}_\text{st})$ \hfill 
$\triangleright~\mathbf{F} \in \mathbb{R}^{T \times 3}$

\State $\mathbf{P} \leftarrow 1-\operatorname{MLP}_{\mathrm{Sigmoid}}(\operatorname{MLP}_{\mathrm{ReLU}}(\mathbf{F}))$ \hfill 
$\triangleright~\mathbf{P} \in \mathbb{R}^{T \times 1}$

\State $\textit{AS}_l \leftarrow \mathbf{f}_\text{pot}$

\State \textbf{return} $\mathbf{P},\textit{AS}_l$ \hfill Return the confidence score.
\EndProcedure

\noindent\rule[0.25\baselineskip]{0.96\textwidth}{0.4pt}
\Procedure{Filtering}{${\mathbf{P}}, X,\tau$} 

\For{$t = 1, \dots, T$}
    \If{$\mathbf{P}_t \geq \tau$ quantile of $\mathbf{P}$}
        \State Add $x_t$ to $\mathcal{N}$ \hfill $\triangleright~\mathcal{N}$ is the set of high-confidence normal samples.
    \Else
        \State Add $x_t$ to $\mathcal{U}$ \hfill $\triangleright~\mathcal{U}$ is the set of uncertain samples.
    \EndIf
\EndFor

\State \textbf{return} $\mathcal{U}$ \hfill $\triangleright~\mathcal{N}$ is filtered and only return $\mathcal{U}$.
\EndProcedure

\end{algorithmic}
\end{spacing}
\end{algorithm}

\section{Experimental Details}\label{experimental}

\subsection{Datasets}\label{datasets}

To verify the effectiveness of the proposed framework, we conduct extended experiments on several publicly available real-world datasets, which are described in detail below:

\begin{itemize}

\item \textbf{CalIt2}~\citep{uci/cailt2} is a visitor flow rate dataset that records building entry and exit counts over 15 weeks to detect event-related anomalies.

\item \textbf{NYC}~\citep{ijcnn/nyc} is a transportation dataset that records passenger counts from New York City taxi rides at 30-minute intervals.

\item \textbf{GECCO}~\citep{moritz2018gecco} is a water quality dataset published in the GECCO Industrial Challenge.

\item \textbf{MSL}~\citep{HundmanCLCS18} is a spacecraft telemetry dataset that contains incident and anomaly data from the Mars Science Laboratory rover, Curiosity.

\item \textbf{PSM}~\citep{kdd/AbdulaalLL21/psm} is a server metrics dataset that records key performance indicators from servers on an online shopping platform.

\item \textbf{ECG}~\citep{kdd/Yoon0L20/ecg} is a medical dataset containing features extracted from electrocardiogram signals, where abnormal heartbeats are labeled as anomalies.

\item \textbf{SMAP}~\citep{HundmanCLCS18} is a spacecraft telemetry dataset that contains soil moisture measurements and telemetry information from the Soil Moisture Active Passive satellite.

\item \textbf{KDD21}~\citep{QiuLQHZWLGZSHJY25} is a multi-domain dataset comprising time-series entities from diverse areas, including healthcare, sports, industry, and robotics.

\end{itemize}
These datasets span multiple application domains and are highly diverse and representative. Specifically, the feature dimensionality ranges from 1 to 32, the average length of the test sequences ranges from 2,520 to 427,617, and the anomaly ratio ranges from 0.57\% to 16.27\%. Such diversity in data characteristics provides a comprehensive and reliable basis for evaluating the performance of CARE. More detailed statistics are reported in Table~\ref{tab:dataset_description}. 

\begin{table}[h]
\centering
\footnotesize
\caption{Statistics of the datasets. (AR: anomaly ratio)}
\renewcommand{\arraystretch}{1.3}
\resizebox{\textwidth}{!}{%
\begin{tabular}{c c c c c c c}

\noalign{\vskip +3ex}
\toprule
\textbf{Dataset} & \textbf{Domain} & \textbf{Dimension} & \textbf{Avg Train Length} & \textbf{Avg Val Length} & \textbf{Avg Test Length} & \textbf{Avg AR (\%)} \\
\midrule
CalIt2 & Visitors flowrate & 2 & 2016 & 504 & 2520 & 4.09 \\
NYC & Transport & 3 & 14016 & 3504 & 4416 & 0.57 \\
GECCO & Water treatment & 9 & 55408 & 13852 & 69261 & 1.25 \\
MSL & Spacecraft & 1 & 46653 & 11663 & 73730 & 5.88\\
ECG & Health & 32 & 44883 & 11221 & 56105 & 16.27\\
PSM & Server Machine & 25 & 105984 & 26497 & 87841 & 11.07\\
SMAP & Spacecraft & 1 & 108146 & 27037 & 427617 & 9.72\\
KDD21 & Multiple & 1 & 14886 & 3721 & 47294 & 0.58\\
\bottomrule
\end{tabular}
}
\label{tab:dataset_description}
\end{table}

\subsection{Metrics}\label{metrics}

Since CARE aims to improve inference efficiency while maintaining detection quality, we categorize the evaluation metrics into two groups: detection-quality metrics and inference-efficiency metrics. 

The detection-quality metrics include Precision (P), Recall (R), F1-score (F1), Affiliated Precision (Aff-P), Affiliated Recall (Aff-R), Affiliated F1-score (Aff-F1), the Area Under the Precision-Recall Curve (A-P), the Area Under the Receiver Operating Characteristic Curve (A-R), the Range-based Area Under the Precision-Recall Curve (R-A-P), the Range-based Area Under the Receiver Operating Characteristic Curve (R-A-R), the Volume Under the Precision-Recall Surface (V-PR), and the Volume Under the Receiver Operating Characteristic Surface (V-ROC).

For inference efficiency, we report the average inference time (Time), per-sample latency (PSL), and throughput (Thr.). Specifically, Time is computed as the average elapsed time over three complete inference runs on each dataset, PSL denotes the average inference time required to process a single sample, and Thr. measures the number of windows processed per unit time under the same experimental settings.

\subsection{Implementation Details}\label{implement}

In our experiments, the dimension of hidden states ($D_h$) and the dimension of latent states ($D_z$) of RMA are set to 16 and 5, respectively. For NCG, the dimension of hidden states is set to 5. RMA  and NCG are trained for 10 epochs and 20 epochs, respectively. All models are trained using the Adam optimizer with an initial learning rate of $10^{-4}$, and a fixed batch size of 64. The sliding-window size is set to be consistent with that of the cascaded complex model by default to ensure semantic consistency. During the optimization of NCG with the confidence-guided selective inference loss,  $\xi_1$ and $\xi_2$ are both set to 0.01, the regularization coefficient $\alpha$ is set to 0.05, the boundary constraint coefficient $\lambda$ is set to 1.0, the quantile threshold risk is set to 0.7, and the fixed width $w$ is set to 0.1. The filtering threshold $\tau$ used in the cascaded inference process is set to 0.2. Before formal timing, we perform three batches of warm-up forward passes to reduce cold-start effects and improve the stability of time measurements. All experiments are implemented in PyTorch with Python 3.8 and conducted on an NVIDIA RTX A6000 GPU with 48 GB of memory.

\begin{figure}[htbp]
    \centering
     \includegraphics[width=1.0\textwidth]{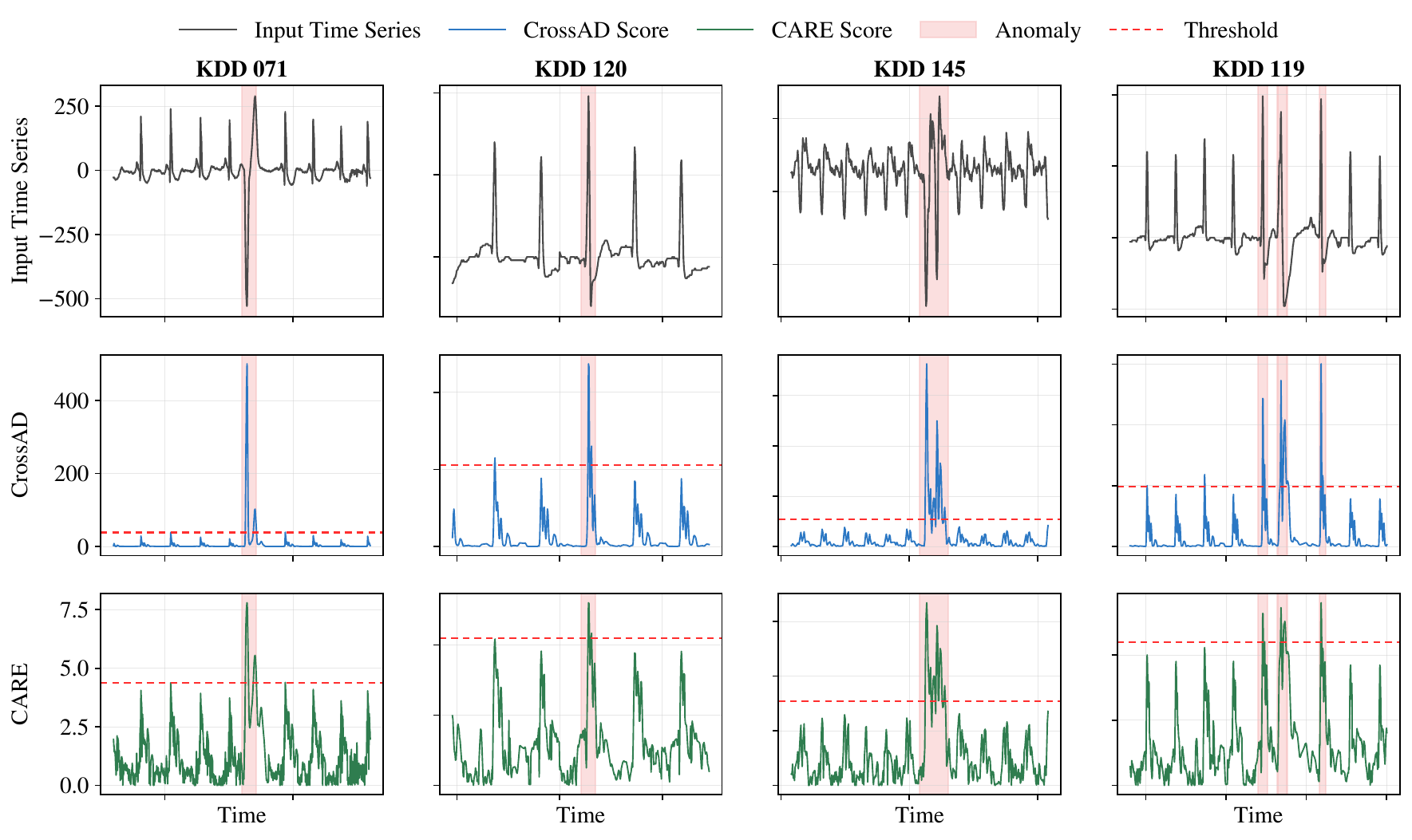}
    \hfill

    \caption{Visualization of anomalies detected by CARE and CrossAD on the KDD21 dataset.}
    \label{fig:visualization}
\end{figure}

\section{Visualization of Anomaly Scores} \label{Anomaly criterion visualization}
We conducted a comparative visualization analysis of the proposed CARE against its corresponding CDM model, CrossAD, to demonstrate the effectiveness of the cascaded framework. As illustrated in Figure~\ref{fig:visualization}, four anomalous subsequences are selected from the KDD21 dataset, where the raw time series and the anomaly scores produced by two methods are reported. The results show that CARE and CrossAD exhibit similar sensitivity to anomalous regions. This indicates that CARE largely preserves the anomaly detection capability of CDM while introducing cascaded inference.  This is because CARE filters out most high-confidence normal samples through LPM and forwards high-risk samples to CrossAD for further evaluation.  For normal regions, the anomaly score distributions of the two methods differ to some extent. This is mainly because the scores of normal samples in CARE are partially produced by LPM, and the score alignment between LPM and CDM further affects the final score distribution. As a result, CARE and CrossAD maintain similar anomaly sensitivity and ranking behavior, while their absolute score values may differ. Moreover, the pre-filtering mechanism of CARE can also help reduce false positives. As shown in the visualization on KDD120 and KDD119, CrossAD incorrectly identifies a normal segment as anomalous, whereas CARE alleviates this issue. This suggests that CARE may achieve more robust detection performance in certain scenarios, in addition to improving inference efficiency.  

\begin{figure*}[t]
    \centering

    % ==================== one line ====================
    \begin{subfigure}[b]{0.24\textwidth}
        \centering
        \includegraphics[width=\textwidth]{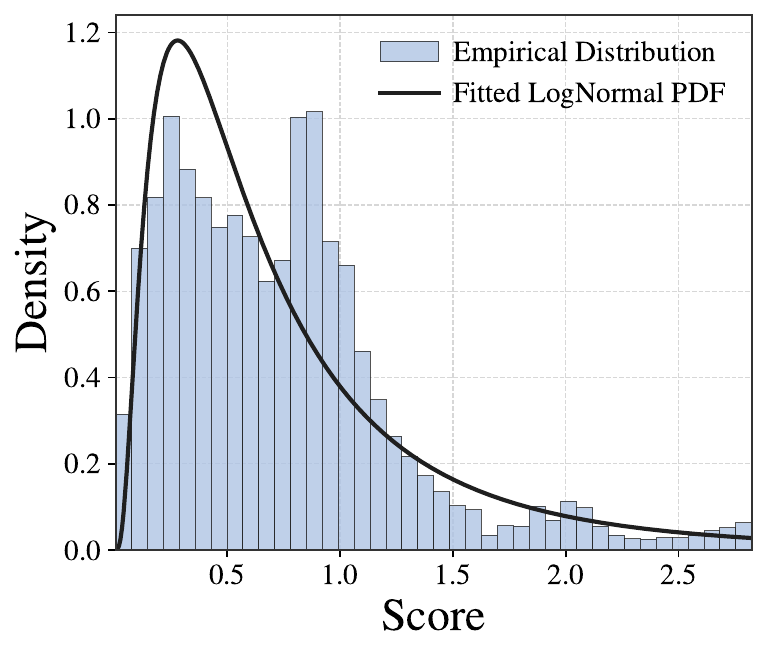}
        \caption{NYC~(LPM)}
        \label{fig:nyc_lpm}
    \end{subfigure}
    \hfill
    \begin{subfigure}[b]{0.24\textwidth}
        \centering
        \includegraphics[width=\textwidth]{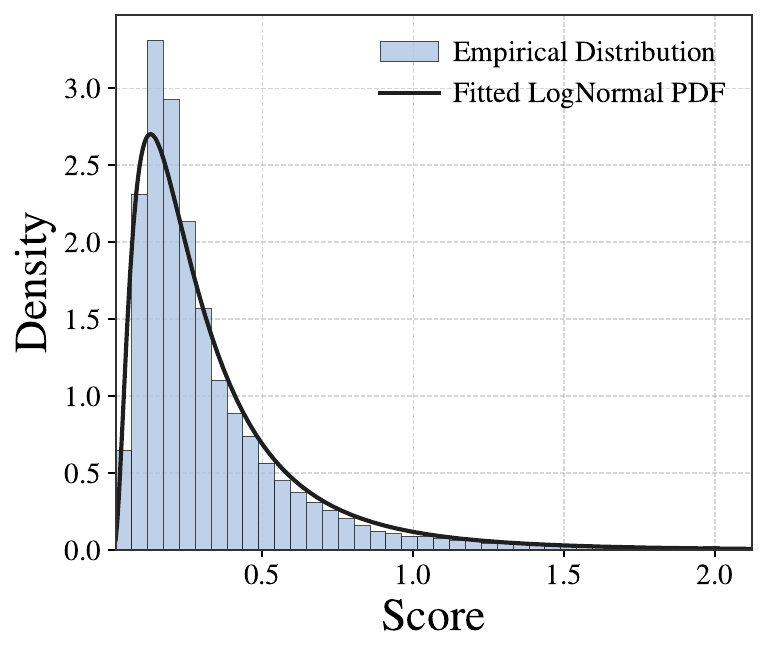}
        \caption{GECCO (LPM)}
        \label{fig:gecco_lpm}
    \end{subfigure}
    \hfill
    \begin{subfigure}[b]{0.24\textwidth}
        \centering
        \includegraphics[width=\textwidth]{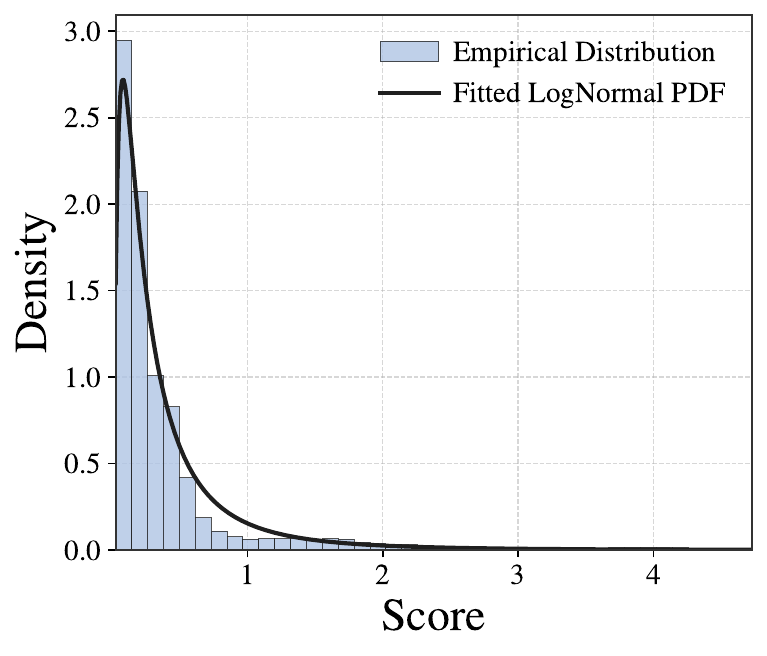}
        \caption{ECG (LPM)}
        \label{fig:ecg_lpm}
    \end{subfigure}
    \hfill
    \begin{subfigure}[b]{0.24\textwidth}
        \centering
        \includegraphics[width=\textwidth]{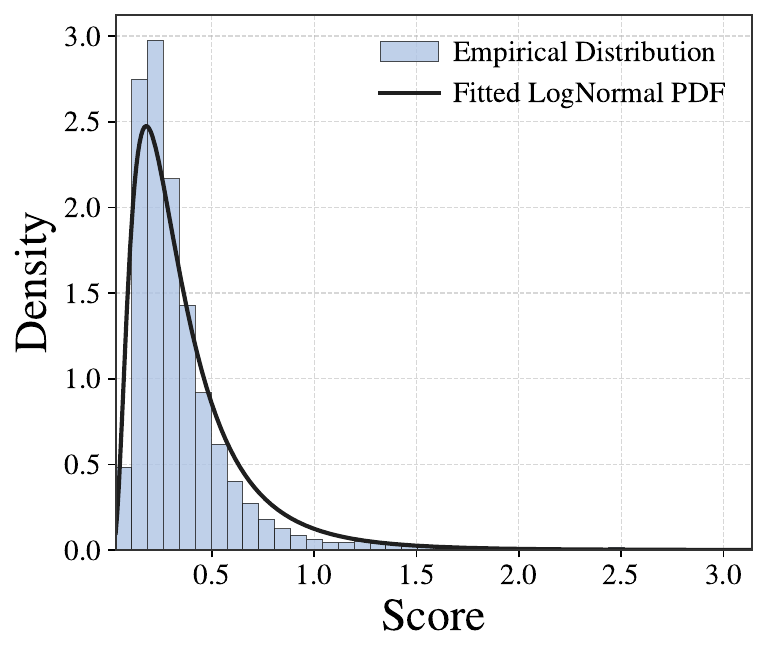}
        \caption{PSM (LPM)}
        \label{fig:psm_lpm}
    \end{subfigure}

    \par\vspace{2mm}

    % ==================== two line ====================
    \begin{subfigure}[b]{0.24\textwidth}
        \centering
        \includegraphics[width=\textwidth]{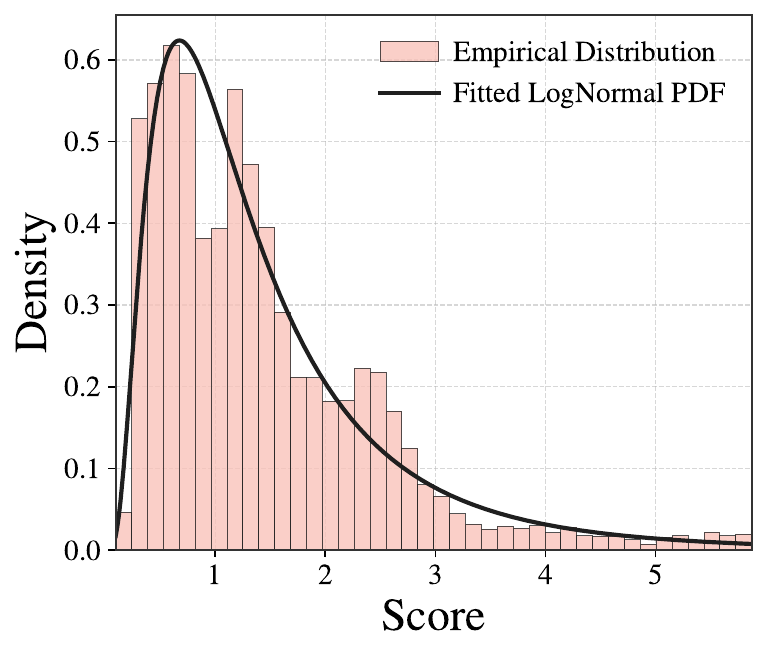}
        \caption{NYC (CDM)}
        \label{fig:NYC_CDM}
    \end{subfigure}
    \hfill
    \begin{subfigure}[b]{0.24\textwidth}
        \centering
        \includegraphics[width=\textwidth]{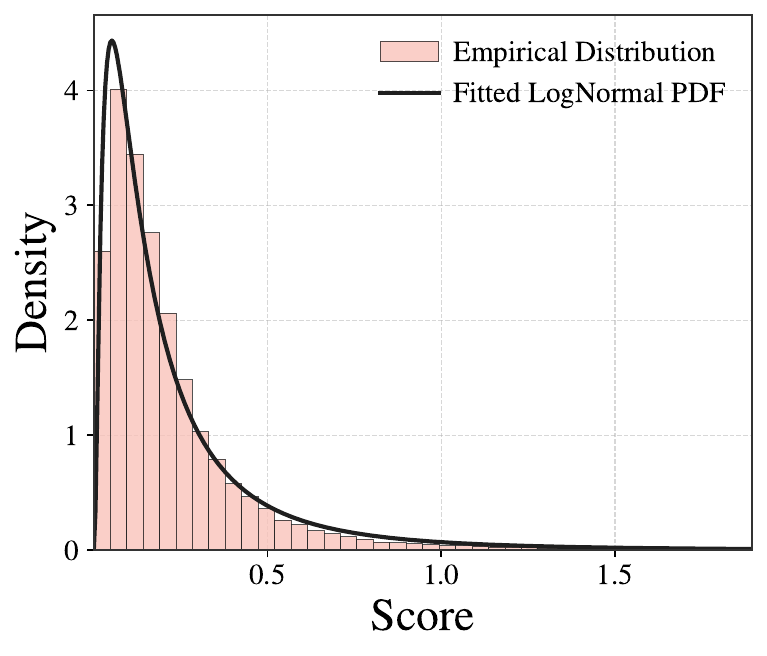}
        \caption{GECCO (CDM)}
        \label{fig:msl_runtime}
    \end{subfigure}
    \hfill
    \begin{subfigure}[b]{0.24\textwidth}
        \centering
        \includegraphics[width=\textwidth]{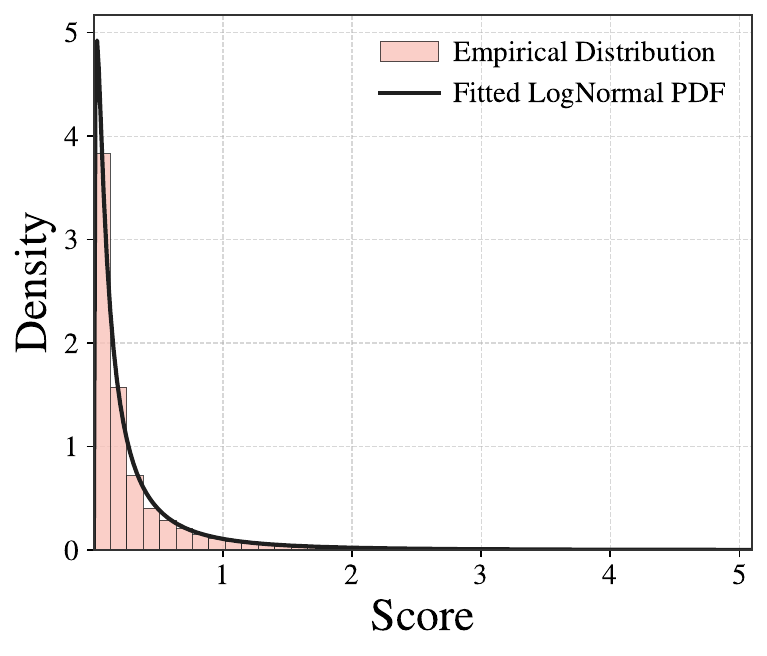}
        \caption{ECG (CDM)}
        \label{fig:smap_runtime}
    \end{subfigure}
    \hfill
    \begin{subfigure}[b]{0.24\textwidth}
        \centering
        \includegraphics[width=\textwidth]{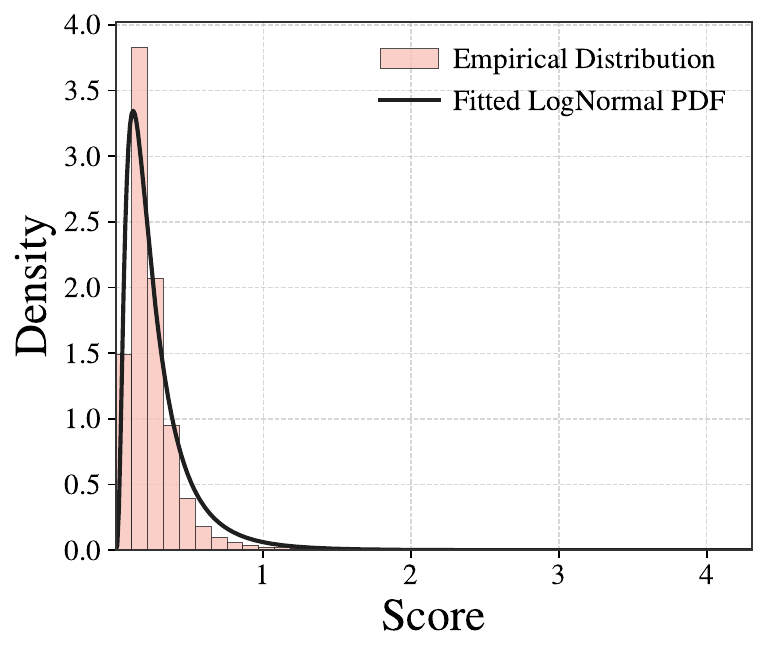}
        \caption{PSM (CDM)}
        \label{fig:kdd21_runtime}
    \end{subfigure}

    \caption{
        Empirical distributions of training anomaly scores and their fitted log-normal PDFs on representative datasets. For readability, each subplot displays scores up to the 99th percentile of the corresponding training distribution.
    }
    \label{fig6:additional_experiments}
\end{figure*}

\begin{table}[htbp]
\centering
\footnotesize
\caption{Comparison of different score alignment strategies in CARE.
All variants use the same trained models and evaluation protocol, with
only the score alignment module replaced. All alignment mappings are
estimated exclusively from training anomaly scores and remain fixed
during testing. The best result in each column is highlighted in bold.}
\renewcommand{\arraystretch}{1.2}
\resizebox{\textwidth}{!}{
\begin{tabular}{
>{\centering\arraybackslash}p{1.3cm} 
>{\raggedright\arraybackslash}p{2.2cm}| 
>{\centering\arraybackslash}p{1cm} 
>{\centering\arraybackslash}p{1cm}| 
>{\centering\arraybackslash}p{1cm}  
>{\centering\arraybackslash}p{1cm}| 
>{\centering\arraybackslash}p{1cm} 
>{\centering\arraybackslash}p{1cm}| 
>{\centering\arraybackslash}p{1cm}  
>{\centering\arraybackslash}p{1cm}| 
>{\centering\arraybackslash}p{1cm} 
>{\centering\arraybackslash}p{1cm} 
}
\noalign{\vskip +3ex}
\toprule
\multicolumn{2}{c|}{\multirow{2}{*}{\textbf{Alignment method}}} & \multicolumn{2}{c|}{\textbf{NYC}} & \multicolumn{2}{c|}{\textbf{GECCO}} & \multicolumn{2}{c|}{\textbf{ECG}} & \multicolumn{2}{c|}{\textbf{PSM}} & \multicolumn{2}{c}{\textbf{Avg}} \\
\cmidrule{3-12}
 & & \textbf{Aff-F} & \textbf{A-P} & \textbf{Aff-F} & \textbf{A-P} & \textbf{Aff-F} & \textbf{A-P} & \textbf{Aff-F} & \textbf{A-P} & \textbf{Aff-F} & \textbf{A-P}\\
\midrule

\multicolumn{2}{c|}{\textbf{No Alignment}} & 0.690 & 0.085 & 0.914 & 0.604 & 0.805 & 0.522 & 0.743 & 0.566 & 0.788 & 0.444\\
\multicolumn{2}{c|}{\textbf{Gaussian MLE}} & 0.697 & 0.088 & 0.927 & 0.621 & 0.810 & 0.526 & 0.744 & 0.568 & 0.795 & 0.451\\
\multicolumn{2}{c|}{\textbf{Empirical CDF}} & \textbf{0.699} & 0.086 & 0.934 & 0.622 & 0.808 & 0.524 & 0.744 & 0.569 & 0.796 & 0.450 \\
\multicolumn{2}{c|}{\textbf{Log-normal MLE (Ours)}} & 0.694 & \textbf{0.090} & \textbf{0.938} & \textbf{0.623} & \textbf{0.813} & \textbf{0.529} &\textbf{ 0.744} & \textbf{0.569} & \textbf{0.797} & \textbf{0.453} \\
\midrule
\end{tabular}
}
\label{tab5:aligment method}
\end{table}

\section{Analysis of MLE-based Score Alignment}  \label{justification of mle}

To assess the score alignment strategy, we visualize the training anomaly
scores generated by LPM and CDM in Figure~\ref{fig6:additional_experiments}. Across all four datasets, the scores exhibit pronounced right-skewness and
long tails. Despite local irregularities, the fitted log-normal PDFs capture
their overall trends, providing a reasonable parametric approximation for score alignment. 
We further compare the proposed log-normal MLE Alignment strategy with no alignment, Gaussian MLE, and empirical CDF. As shown in Table~\ref{tab5:aligment method}, all three aligned variants achieve better average performance than the unaligned variant, demonstrating the benefit of mapping the heterogeneous LPM and CDM scores into a comparable
probability space. Log-normal MLE achieves the best average Aff-F and
AUC-PR and obtains the strongest results on GECCO and ECG. Meanwhile,
Gaussian MLE and empirical CDF remain competitive, indicating that CARE
does not critically depend on a specific distributional assumption.
Together with the observed right-skewed score distributions, these results
support log-normal MLE as a reasonable and stable choice for score
alignment.

\section{Additional Hyperparameter Sensitivity Analysis}\label{additional sensitivity}

\begin{figure}[htbp]
    \centering
    \begin{subfigure}[b]{0.24\textwidth}
        \centering
        \includegraphics[width=\textwidth]{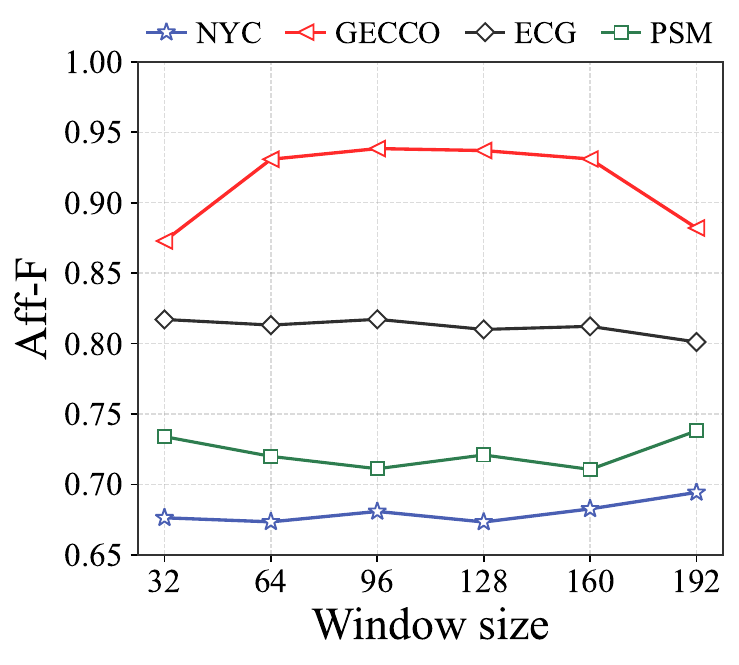}
        \caption{Window size}
        \label{fig:windowsize}
    \end{subfigure}
    \hfill
    \begin{subfigure}[b]{0.24\textwidth}
        \centering
        \includegraphics[width=\textwidth]{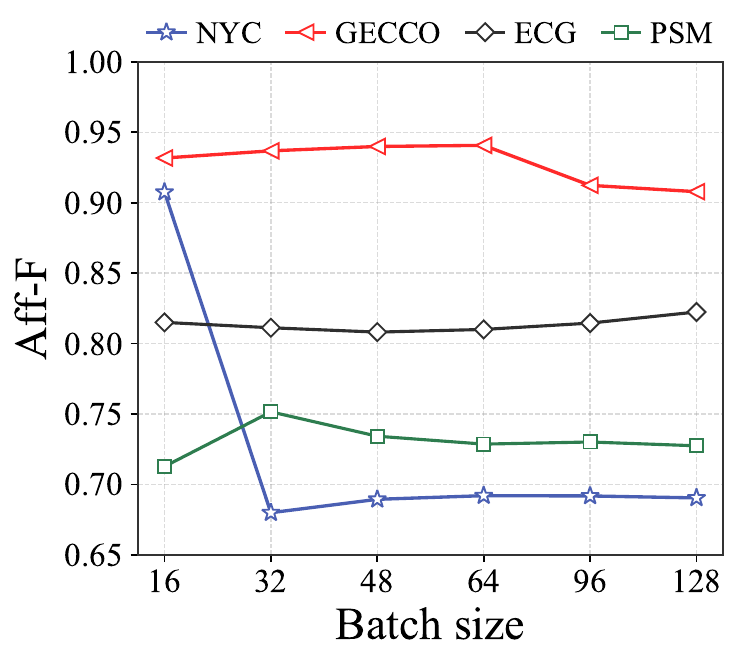}
        \caption{Batch size}
        \label{fig:batchsize}
    \end{subfigure}
    \hfill
    \begin{subfigure}[b]{0.24\textwidth}
        \centering
        \includegraphics[width=\textwidth]{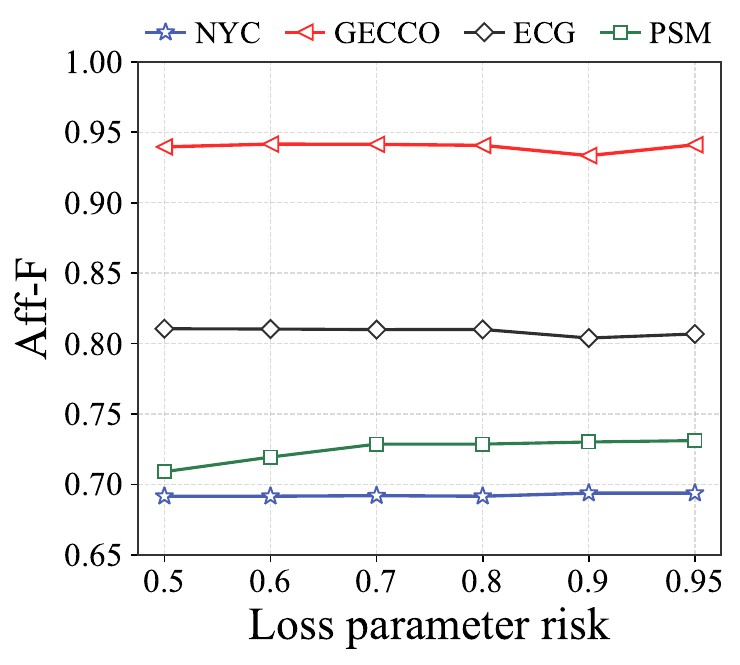}
        \caption{Loss parameter $risk$}
        \label{fig:loss_risk}
    \end{subfigure}
    \hfill
    \begin{subfigure}[b]{0.24\textwidth}
        \centering
        \includegraphics[width=\textwidth]{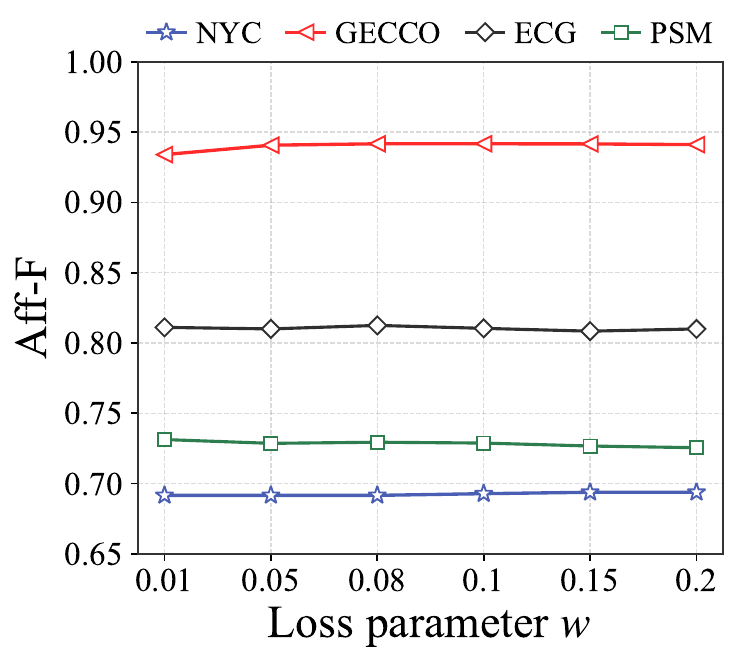}
        \caption{Loss parameter $\omega$}
        \label{fig:loss_width}
    \end{subfigure}
    \caption{Additional parameter sensitivity studies of main hyperparameters in CARE.}
    \label{fig:add_sensitivity}
\end{figure}

Key hyperparameters affecting anomaly detection performance include the ratio $\text{risk}$ and boundary width $w$ in the confidence-guided selective inference loss function. During NCG optimization, $\text{risk}$ controls the proportion of training samples, determining the boundary between high-confidence normal samples and high-risk samples. Both excessively large and small $\text{risk}$ values negatively impact boundary learning, thus reducing detection quality. When the $\text{risk}$ is too large, more anomaly samples may be misclassified as high-confidence normal samples, increasing the risk of missed detections; when $\text{risk}$ is too small, many normal samples may be misclassified as high-risk samples, making it difficult for the model to obtain reliable training targets. As shown in Figure~\ref{fig:loss_risk}, $\text{risk}$ remains stable and easily adjustable within the range of 0.7-0.8. The boundary width $w$ determines the range of samples involved in fine-grained ranking near the risk threshold. As shown in Figure~\ref{fig:loss_width}, model performance gradually stabilizes when $w$ exceeds 0.08.  A smaller $w$ covers only a limited number of boundary samples within each batch, making it difficult to effectively supplement the main loss, while appropriately increasing $w$ introduces more threshold-adjacent samples, thereby improving the stability of boundary learning.  

Furthermore, we analyzed the impact of window size and batch size on model performance. Window size is a fundamental parameter in time series analysis. This paper primarily focuses on the window size of LPM, while the window size of CDM is roughly referenced from the original model. In our experiments, CDM adopts the SOTA model CrossAD, with window sizes of 192, 96, 48, and 192 for the NYC, GECCO, ECG, and PSM datasets, respectively, to suit the current datasets. As shown in Figure~\ref{fig:windowsize}, when the window size of LPM matches that of CDM, the model typically achieves stable or relatively superior detection performance, suggesting that consistent window sizes help enhance semantic alignment between LPM and CDM. Regarding batch size, experimental results indicate that a range of 32-64 is relatively stable, as shown in Figure~\ref{fig:batchsize}.  

\begin{figure}[htbp]
    \centering
    \begin{subfigure}[b]{0.24\textwidth}
        \centering
        \includegraphics[width=\textwidth]{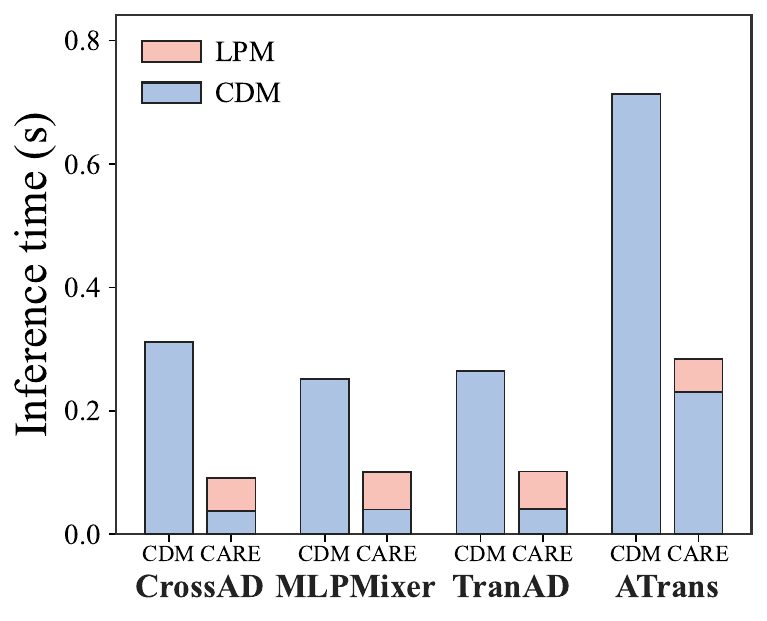}
        \caption{CalIt2}
        \label{fig:nyc_time}
    \end{subfigure}
    \hfill
    \begin{subfigure}[b]{0.24\textwidth}
        \centering
        \includegraphics[width=\textwidth]{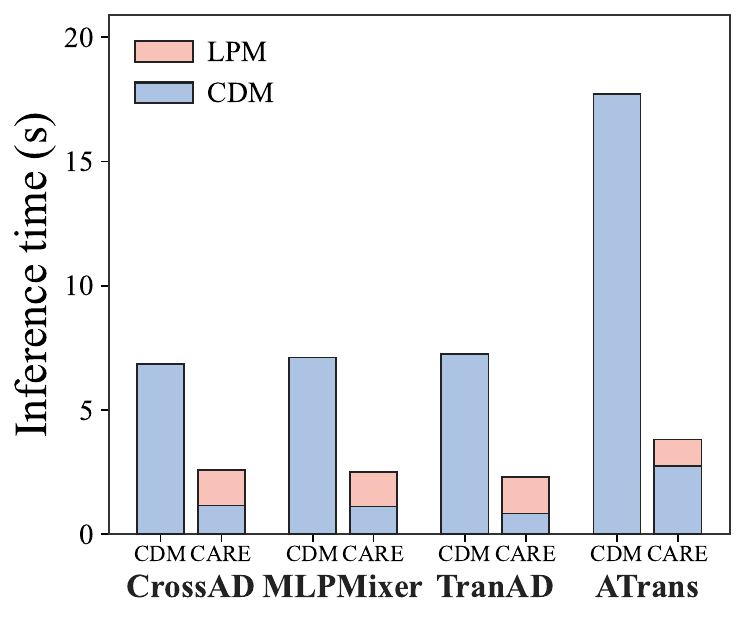}
        \caption{MSL}
        \label{fig:gecco_time}
    \end{subfigure}
    \hfill
    \begin{subfigure}[b]{0.24\textwidth}
        \centering
        \includegraphics[width=\textwidth]{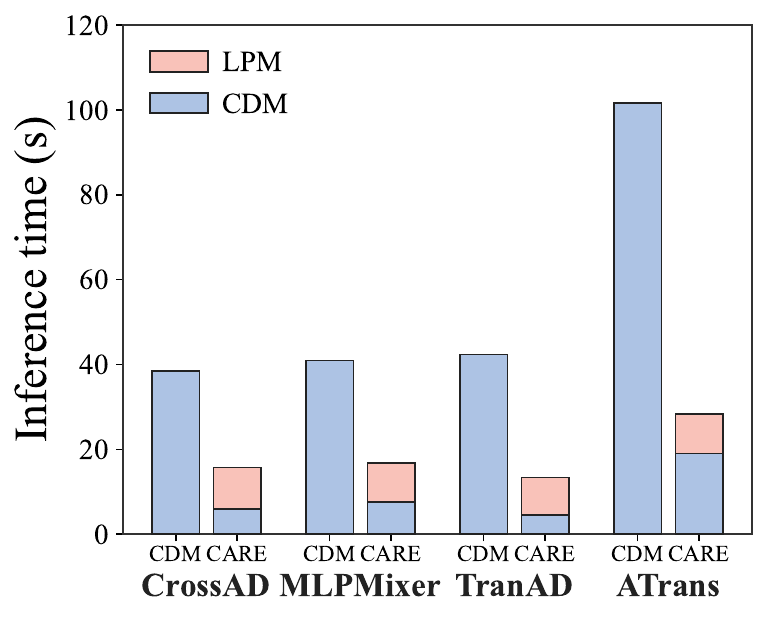}
        \caption{SMAP}
        \label{fig:ecg_time}
    \end{subfigure}
    \hfill
    \begin{subfigure}[b]{0.24\textwidth}
        \centering
        \includegraphics[width=\textwidth]{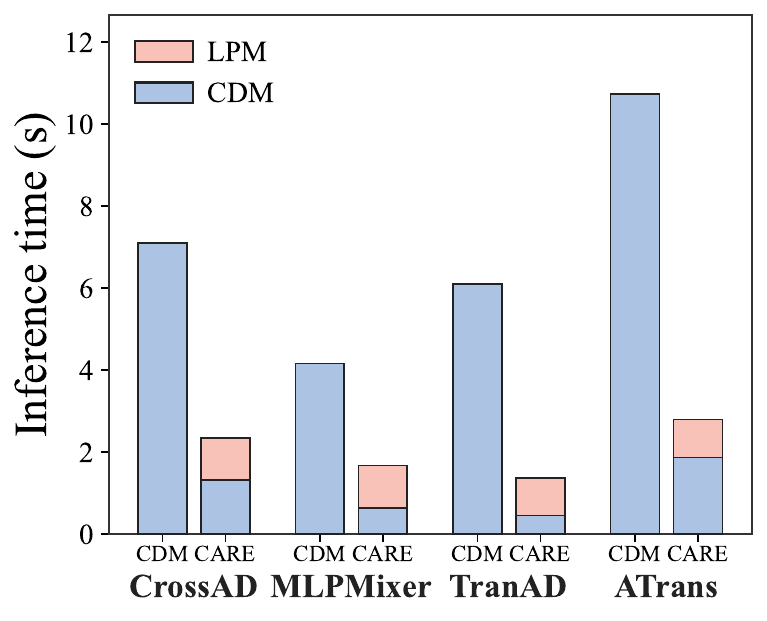}
        \caption{KDD21}
        \label{fig:psm_time}
    \end{subfigure}
    \caption{Additional inference time comparison of CARE with various CDM backbones.}
    \label{fig7:different CDM}
\end{figure}

\section{Additional Adaptability Analysis}\label{additional adaptability}
Due to space limitations, we provide an additional generality analysis of CARE on the remaining four datasets in this section. As shown in Table~\ref{tab5:care generality}, CARE consistently reduces inference time while maintaining competitive detection performance across different CDM models. Although the detection performance is not always superior to the original CDM, the degradation remains within 1\% in terms of Aff-F1, demonstrating that CARE can effectively preserve detection quality under substantially reduced inference costs. The corresponding inference time comparisons are presented in Figure~\ref{fig7:different CDM}. These results further demonstrate the generality of CARE as a cascaded inference framework for diverse TSAD models.

\begin{table}[htbp]
\centering
\footnotesize
\caption{Additional analysis of CARE generality with different CDM models. Aff-F, Time, and $S\times$ are Affiliated F1-score, Inference time, and Speedup, respectively. The best ones are in bold.}
\renewcommand{\arraystretch}{1.2}
\resizebox{\textwidth}{!}{
\begin{tabular}{
>{\centering\arraybackslash}p{1cm} 
>{\raggedright\arraybackslash}p{1cm}| 
>{\centering\arraybackslash}p{0.8cm}  
>{\centering\arraybackslash}p{0.7cm}  
>{\centering\arraybackslash}p{0.7cm}|
>{\centering\arraybackslash}p{0.8cm}  
>{\centering\arraybackslash}p{0.7cm}  
>{\centering\arraybackslash}p{0.7cm}| 
>{\centering\arraybackslash}p{0.8cm}  
>{\centering\arraybackslash}p{0.7cm}  
>{\centering\arraybackslash}p{0.7cm}| 
>{\centering\arraybackslash}p{0.8cm}  
>{\centering\arraybackslash}p{0.7cm}  
>{\centering\arraybackslash}p{0.7cm}| 
>{\centering\arraybackslash}p{0.8cm}  
>{\centering\arraybackslash}p{0.7cm}  
>{\centering\arraybackslash}p{0.7cm} 
}
\noalign{\vskip +3ex}
\toprule
\multicolumn{2}{c|}{\multirow{2}{*}{\textbf{Method}}} & \multicolumn{3}{c|}{\textbf{CalI2}} & \multicolumn{3}{c|}{\textbf{MSL}} & \multicolumn{3}{c|}{\textbf{SMAP}} & \multicolumn{3}{c|}{\textbf{KDD21}} & \multicolumn{3}{c}{\textbf{Avg}} \\
\cmidrule{3-17}
 & & \textbf{Aff-F} & \textbf{Time} & \textbf{S$\times$} & \textbf{Aff-F} & \textbf{Time} & \textbf{S$\times$} & \textbf{Aff-F} & \textbf{Time} & \textbf{S$\times$} & \textbf{Aff-F} & \textbf{Time} & \textbf{S$\times$} & \textbf{Aff-F} & \textbf{Time} & \textbf{S$\times$}\\
\midrule
\multicolumn{2}{c|}{\textbf{CrossAD}}  & \textbf{0.788} & 0.312 & 1.000 & 0.761 & 6.862 & 1.000 & 0.712 & 38.46 & 1.000 & 0.712 & 7.101 & 1.000 & 0.743 & 13.18 & 1.000 \\
\multicolumn{2}{c|}{\textbf{CARE} (CrossAD as CDM)}  & 0.787 & \textbf{0.091} & \textbf{3.429} & \textbf{0.770} & \textbf{2.587} & \textbf{2.653} & \textbf{0.721} & \textbf{15.71} & \textbf{2.448} & \textbf{0.716} & \textbf{2.354} & \textbf{3.016} & \textbf{0.748} & \textbf{5.185} & \textbf{2.886}\\
\midrule
\multicolumn{2}{c|}{\textbf{MLPMixer}}  & 0.779 & 0.252 & 1.000 & \textbf{0.768} & 7.111 & 1.000 & 0.680 & 40.98 & 1.000 & 0.714 & 4.162 & 1.000 & 0.735 & 13.13 & 1.000\\
\multicolumn{2}{c|}{\textbf{CARE} (MLPMixer as CDM)} & \textbf{0.786} & \textbf{0.101} & \textbf{2.508} & 0.767 & \textbf{2.508} & \textbf{2.835} & \textbf{0.680} & \textbf{16.79} & \textbf{2.439} & \textbf{0.715} & \textbf{1.680} & \textbf{2.477} & \textbf{0.737} & \textbf{5.271} & \textbf{2.565} \\
\midrule
\multicolumn{2}{c|}{\textbf{TranAD}} & 0.780 & 0.265 & 1.000 & \textbf{0.763} & 7.267 & 1.000 & \textbf{0.682} & 42.33 & 1.000 & \textbf{0.714} & 6.108 & 1.000 & \textbf{0.734 }& 13.99 & 1.000\\
\multicolumn{2}{c|}{\textbf{CARE} (TranAD as CDM)} & \textbf{0.788} & \textbf{0.102} & \textbf{2.588} & 0.758 & \textbf{2.314} & \textbf{3.139} & 0.680 & \textbf{13.37} & \textbf{3.165} & 0.707 & \textbf{1.379} & \textbf{4.426} & 0.733 & \textbf{4.293} & \textbf{3.329}\\
\midrule
\multicolumn{2}{c|}{\textbf{ATrans}} & 0.673 & 0.713 & 1.000 & 0.679 & 17.71 & 1.000 & 0.679 & 101.7 & 1.000 & \textbf{0.695} & 10.73 & 1.000 & 0.681 & 32.71 & 1.000\\

\multicolumn{2}{c|}{\textbf{CARE} (Atrans as CDM)} & \textbf{0.744} & \textbf{0.283} & \textbf{2.515} & \textbf{0.707} & \textbf{3.815} & \textbf{4.642} & \textbf{0.679} & \textbf{28.34} & \textbf{3.587} & 0.688 & \textbf{2.798} & \textbf{3.834} & \textbf{0.704} & \textbf{8.810} & \textbf{3.645} \\
\bottomrule
\end{tabular}
}
\label{tab5:care generality}
\end{table}

\section{Stability Analysis}
\label{stability analysis}

We further evaluated the stability of CARE on a representative dataset using different random seeds. We report the mean and standard deviation for both detection quality and inference efficiency metrics, as shown in Table \ref{tab6:stability analysis}. The results demonstrate that CARE maintains stable detection quality and inference efficiency across repeated runs.

\begin{table}[h]
\centering
\tiny
\captionsetup{width=0.8\textwidth}
\caption{Stability analysis of CARE over repeated runs with different random seeds. Each value is reported as mean $\pm$ standard deviation.}
\renewcommand{\arraystretch}{1.3}
\resizebox{0.8\textwidth}{!}{%
\begin{tabular}{c | c c c c c c c c c}

\noalign{\vskip +3ex}
\toprule
\textbf{Metric} & \textbf{NYC} & \textbf{GECCO} & \textbf{ECG} & \textbf{PSM} & \textbf{Avg} \\
\midrule
\textbf{Aff-F} & 0.694 $\pm$ 0.003 & 0.938 $\pm$ 0.005 & 0.813 $\pm$ 0.005 & 0.744 $\pm$ 0.003 & 0.797 $\pm$ 0.004 \\
\textbf{A-P} & 0.090 $\pm$ 0.004 & 0.624 $\pm$ 0.001 & 0.529 $\pm$ 0.004 & 0.569 $\pm$ 0.008 & 0.453 $\pm$ 0.004 \\
\textbf{Time} & 0.184 $\pm$ 0.014 & 3.889 $\pm$ 0.085 & 6.750 $\pm$ 0.143 & 9.057 $\pm$ 0.063 & 4.991 $\pm$ 0.076 \\
\textbf{PSL} & 0.042 $\pm$ 0.003 & 0.056 $\pm$ 0.001 & 0.120 $\pm$ 0.002 & 0.103 $\pm$ 0.001 & 0.081 $\pm$ 0.002 \\
\bottomrule
\end{tabular}
}
\label{tab6:stability analysis}
\end{table}

\section{Broader impacts}\label{broader impacts}

Time series anomaly detection is widely applied across diverse real-world scenarios. The efficient, reliable, and model-agnostic cascaded framework proposed in this paper can accelerate anomaly detection, reduce computational overhead, and facilitate deployment in resource-constrained or edge-computing environments. In addition, its model-agnostic design allows future advanced anomaly detection models to be seamlessly integrated, potentially further improving overall detection performance. Nevertheless, time series anomaly detection systems still cannot completely avoid false positives and false negatives in practical applications. False positives may increase manual inspection costs, while false negatives may cause critical anomalies to be missed. Therefore, detection results should be carefully evaluated in the context of specific deployment scenarios.

\section{Limitations}\label{limitations}

In this paper, we propose CARE, a novel cascaded framework for time-series anomaly detection. CARE introduces a lightweight filtering model to distinguish high-confidence normal samples from hard samples, thereby overcoming the limitation of existing methods that apply complex and uniform inference to all samples. In our current experiments, a unified filtering threshold is used across all datasets. However, the optimal threshold may vary with different data distributions and may require adaptation under evolving time series environments. Developing data-adaptive routing strategies that can dynamically adjust filtering thresholds under distribution shifts or concept drift is an important direction for future work.

\newpage

\section{Full Experimental Results}\label{full results}

Due to space limitations in the main text, this section presents the complete experimental results for CARE. Table~\ref{tab:inference time} reports the full detection-efficiency results in terms of inference time, per-sample latency, and throughput. Tables~\ref{tab:p_r_f1}-\ref{tab:A-R} present the detection-quality results from different perspectives, including Precision, Recall, and F1-score; Affiliated Precision, Affiliated Recall, and Affiliated F1-score; AUC-PR, R-AUC-PR, and VUS-PR; and AUC-ROC, R-AUC-ROC, and VUS-ROC, respectively.  Furthermore, we include the results of several classical anomaly detection methods, including IF (Isolation Forest) \citep{icdm/LiuTZ08}, LOF \citep{BreunigKNS00}, and PCA \citep{shyu2003novel}. These results further demonstrate that CARE significantly improves inference efficiency while maintaining competitive detection quality.  

\begin{table}[h]
\centering
\footnotesize
\caption{Average Time (Inference Time, s), PSL (Per-Sample Latency, ms) and Thr. (Throughput, windows/s) across all datasets. The best results are in bold, while the second-best results are underlined.}
\renewcommand{\arraystretch}{1.5}
\resizebox{\textwidth}{!}{
\begin{tabular}{c | c | c c c c c c c c c c c c c c c c}
\noalign{\vskip +3ex}
\toprule
\textbf{Dataset} & \textbf{Metric} & \textbf{CARE} & \textbf{Cross} & \textbf{CATCH} & \textbf{Mixer} & \textbf{DC} & \textbf{TranAD} & \textbf{ATrans} & \textbf{CALF} & \textbf{GS} & \textbf{Patch} & \textbf{iTrans} & \textbf{TsNet} & \textbf{LSTM} & \textbf{AE} & \textbf{OCSVM}\\
\midrule
\multirow{3}{*}{\textbf{CalIt2}} & \textbf{Time} & \textbf{0.091} & 0.312 & 0.583 & 0.252 & 0.877 & 0.265 & 0.713 & 0.522 & 1.016 & \underline{0.213} & 0.222 & 0.572 & 0.474 & 0.472 & 0.259\\
                       & \textbf{PSL} & \textbf{0.036} & 0.124 & 0.231 & 0.100 & 0.348 & 0.105 & 0.283 & 0.207 & 0.403 & \underline{0.085} & 0.088 & 0.227 & 0.188 & 0.187 & 0.103\\
                        & \textbf{Thr.} & \textbf{27679} & 11412 & 4578 & 10878 & 3103.3 & 10274 & 3744.8 & 5431.1 & 2673.2 & \underline{13984} & 12701 & 4766.8 & 5324.8 & 5343.4 & 9740.4\\
\multirow{3}{*}{\textbf{NYC}} & \textbf{Time} & \textbf{0.184} & 0.501 & 0.914 & 0.762 & 1.938 & 0.441 & 1.074 & 0.829 & 2.058 & \underline{0.323} & 0.751 & 1.663 & 0.834 & 0.557 & 3.622\\
                         & \textbf{PSL} & \textbf{0.042} & 0.113 & 0.207 & 0.173 & 0.439 & 0.100 & 0.243 & 0.188 & 0.466 & \underline{0.073} & 0.170 & 0.377 & 0.189 & 0.126 & 0.103\\
                         & \textbf{Thr.} & \textbf{24014} & 9634.4 & 5034.7 & 6512.7 & 2370.4 & 10574 & 4300.4 & 6162.3 & 2652.6 & 14903 & \underline{15871} & 5270.6 & 5302.6 & 7949.5 & 1219.2\\
\multirow{3}{*}{\textbf{GECCO}} & \textbf{Time} & \textbf{3.889} & 11.98 & 23.64 & 15.20 & 22.79 & 9.520 & 17.03 & 15.22 & 26.01 & 6.121 & \underline{4.654} & 17.44 & 17.77 & 9.712 & 291.5\\
                         & \textbf{PSL} & \textbf{0.056} & 0.173 & 0.341 & 0.219 & 0.329 & 0.137 & 0.246 & 0.220 & 0.375 & 0.088 & \underline{0.067} & 0.252 & 0.257 & 0.140 & 4.210\\
                         & \textbf{Thr.} & \textbf{17809} & 5805.8 & 2935.3 & 4569.4 & 3048.1 & 7306.8 & 4197.9 & 4575.1 & 2671.6 & 11384 & \underline{17311} & 3983.7 & 3898.2 & 7136.5 & 237.54\\
\multirow{3}{*}{\textbf{MSL}}  & \textbf{Time} & \textbf{3.087} & 6.862 & 11.56 & 7.111 & 24.79 & 7.276 & 17.71 & 14.75 & 27.50 & 6.009 & \underline{4.292} & 18.31 & 17.56 & 10.73 & 114.6\\
                         & \textbf{PSL} & \textbf{0.042} & 0.093 & 0.157 & 0.096 & 0.336 & 0.099 & 0.240 & 0.200 & 0.373 & 0.082 & \underline{0.058} & 0.248 & 0.238 & 0.145 & 1.555\\
                         & \textbf{Thr.} & \textbf{23880} & 10791 & 6397.8 & 11350 & 2978.2 & 10166 & 4197.4 & 5018.1 & 2687.6 & 12318 & \underline{17281} & 4049.2 & 4199.2 & 6874.2 & 642.99\\
\multirow{3}{*}{\textbf{ECG}} & \textbf{Time} & \textbf{6.750} & 32.58 & 71.86 & 38.31 & 29.72 & 7.050 & 15.04 & 24.55 & 21.19 & 7.693 & \underline{6.823} & 14.12 & 14.18 & 8.786 & 145.5\\
                         & \textbf{PSL} & \textbf{0.120} & 0.581 & 1.281 & 0.683 & 0.530 & 0.126 & 0.268 & 0.438 & 0.378 & 0.137 & \underline{0.122} & 0.252 & 0.253 & 0.157 & 2.594\\
                         & \textbf{Thr.} & \textbf{8312.1} & 1767.5 & 782.09 & 1485.4 & 1919.8 & 8018.1 & 4015.0 & 2300.6 & 2658.4 & 7425.1 & \underline{8201.8} & 4061.0 & 3969.5 & 6399.8 & 385.53\\
\multirow{3}{*}{\textbf{PSM}} & \textbf{Time} & \textbf{9.057} & 30.562 & 72.60 & 35.70 & 36.81 & 13.50 & 24.06 & 32.78 & 33.33 & 10.86 & \underline{9.365} & 23.45 & 23.81 & 14.86 & 537.0\\
                       & \textbf{PSL} & \textbf{0.103} & 0.348 & 0.827 & 0.406 & 0.419 & 0.154 & 0.274 & 0.373 & 0.379 & 0.124 & \underline{0.107} & 0.267 & 0.271 & 0.169 & 6.114\\
                       & \textbf{Thr.} & \textbf{9699.1} & 2883.6 & 1211.3 & 2470.1 & 2396.3 & 6632.8 & 3932.4 & 2688.4 & 2643.2 & 8236.8 & \underline{9435.0} & 3757.6 & 3693.1 & 5918.0 &163.56\\
\multirow{3}{*}{\textbf{SMAP}} & \textbf{Time} & \textbf{16.71} & 38.46 & 67.94 & 40.98 & 152.4 & 42.33 & 101.7 & 80.23 & 164.9 & 35.88 & \underline{22.67} & 108.6 & 187.6 & 155.7 & 1423\\
                         & \textbf{PSL} & \textbf{0.039} & 0.090 & 0.159 & 0.096 & 0.356 & 0.099 & 0.238 & 0.188 & 0.386 & 0.084 & \underline{0.053} & 0.254 & 0.439 & 0.364 & 3.328\\
                         & \textbf{Thr.} & \textbf{23848} & 11136 & 6353.5 & 10442 & 2808.1 & 10117 & 4216.2 & 5364.2 & 2593.3 & 12116 & \underline{18919} & 3939.6 & 2279.1 & 2746.8 & 300.17\\
\multirow{3}{*}{\textbf{KDD21}} & \textbf{Time}  & \textbf{2.354}  & 7.101  & 13.05 & 4.162  & 11.26 & 6.108  & 10.73 & 8.330  & 23.53 & 3.396  & \underline{2.957}  & 18.519 & 9.861  & 8.409 & 38.39\\
                                & \textbf{PSL}  & \textbf{0.057}  & 0.173  & 0.318  & 0.101  & 0.274  & 0.149  & 0.261  & 0.203  & 0.573  & 0.083  & \underline{0.072}  & 0.451  & 0.240  & 0.205 & 0.935\\
                                & \textbf{Thr.}  & \textbf{19782} & 6248.7 & 3872.1 & 10035 & 3721.4 & 7198.7 & 3842.9 & 5365.5 & 1795.8 & 12557 & \underline{14632} & 2430.3 & 4970.8 & 6189.6 & 7777.7\\

\midrule   
\multirow{3}{*}{\textbf{Avg}} & \textbf{Time}  & \textbf{5.265}  & 16.04 & 32.77 & 17.81 & 35.08 & 10.81 & 23.51 & 22.15 & 37.45 & 8.811  & \underline{6.466}  & 25.34 & 34.01 & 26.15 & 319.3\\
                                & \textbf{PSL}   & \textbf{0.062}  & 0.212  & 0.440  & 0.234  & 0.379  & 0.121  & 0.257  & 0.252  & 0.417  & 0.094  & \underline{0.092}  & 0.291  & 0.259  & 0.187 & 2.457\\
                                & \textbf{Thr.}  & \textbf{19378} & 7460.1 & 3895.6 & 7218.1 & 2793.2 & 8786.3 & 4055.9 & 4613.2 & 2547.0 & 11615 & \underline{14294} & 4032.4 & 4204.7 & 6069.8 & 2558.4\\
\midrule 
\multicolumn{2}{c|}{\textbf{$1^{\text{st}}$ count}} & \textbf{27} & 0 & 0 & 0 & 0 & 0 & 0 & 0 & 0 & 0 & 0 & 0 & 0 & 0 & 0\\
\bottomrule
\end{tabular}
}
\label{tab:inference time}
\end{table}

\begin{table}[t]
\centering
\footnotesize
\caption{Average P (Precision), R (Recall) and F1 (F1-score) across all datasets. The best results are in bold, while the second-best results are underlined.}
\renewcommand{\arraystretch}{1.5}
\resizebox{\textwidth}{!}{
\begin{tabular}{c | c | c c c c c c c c c c c c c c c c c c}
\noalign{\vskip +3ex}
\toprule
\textbf{Dataset} & \textbf{Metric} & \textbf{CARE} & \textbf{Cross} & \textbf{CATCH} & \textbf{Mixer} & \textbf{DC} & \textbf{TranAD} & \textbf{ATrans} & \textbf{CALF} & \textbf{GS} & \textbf{Patch} & \textbf{iTrans} & \textbf{TsNet} & \textbf{LSTM} & \textbf{AE} & \textbf{OCSVM} & \textbf{IF} & \textbf{LOF} & \textbf{PCA}\\
\midrule
\multirow{3}{*}{\textbf{CalIt2}} & \textbf{P} & 0.105 & 0.106 & \textbf{0.120} & 0.114 & 0.029 & 0.089 & 0.029 & 0.096 & 0.103 & 0.108 & 0.114 & 0.097 & 0.104 & 0.102 & \underline{0.114} & 0.029 & 0.056 & 0.096 \\
                         & \textbf{R}   & 0.608 & 0.568 & 0.432 & 0.473 & \underline{1.000} & 0.284 & \textbf{1.000} & 0.149 & 0.324 & 0.365 & 0.324 & 0.432 & 0.527 & 0.604 & 0.324 & 1.000 & 0.149 & 0.568\\
                         & \textbf{F1}  & 0.180 & 0.179 & \textbf{0.188} & \underline{0.183} & 0.057 & 0.136 & 0.057 & 0.116 & 0.156 & 0.166 & 0.168 & 0.158 & 0.174 & 0.174 & 0.169 & 0.057 & 0.081 & 0.164 \\
\multirow{3}{*}{\textbf{NYC}} & \textbf{P} & \underline{0.095} & 0.082 & 0.055 & 0.029 & 0.045 & 0.046 & 0.051 & \textbf{0.117} & 0.080 & 0.045 & 0.061 & 0.035 & 0.078 & 0.052 & 0.022 & 0.022 & 0.024 & 0.070\\
                         & \textbf{R}   & 0.384 & 0.495 & 0.212 & 0.828 & 0.121 & 0.162 & 0.081 & 0.152 & 0.091 & 0.293 & 0.253 & 0.465 & 0.192 & 0.343 & \textbf{1.000} & \underline{0.998} & 0.525 & 0.192\\
                         & \textbf{F1}  & \textbf{0.152} & 0.141 & 0.087 & 0.056 & 0.066 & 0.072 & 0.062 & 0.132 & 0.085 & 0.078 & 0.099 & 0.064 & 0.111 & 0.091 & 0.044 & 0.044 & 0.046 & 0.103 \\
\multirow{3}{*}{\textbf{GECCO}} & \textbf{P} & \textbf{0.782} & 0.473 & 0.322 & 0.672 & 0.011 & 0.556 & 0.011 & 0.369 & 0.296 & 0.350 & 0.229 & 0.472 & 0.763 & 0.777 & 0.047 & 0.011 & 0.131 & \underline{0.779}\\
                         & \textbf{R}   & 0.734 & 0.728 & 0.481 & 0.267 & \textbf{1.000} & 0.321 & 1.000 & 0.426 & 0.388 & 0.497 & 0.290 & 0.534 & 0.268 & 0.267 & 0.430 & \underline{1.000} & 0.299 & 0.262\\
                         & \textbf{F1}   & \textbf{0.758} & \underline{0.573} & 0.386 & 0.382 & 0.021 & 0.407 & 0.021 & 0.396 & 0.336 & 0.411 & 0.256 & 0.501 & 0.397 & 0.398 & 0.085 & 0.021 & 0.182 & 0.392 \\
\multirow{3}{*}{\textbf{MSL}}  & \textbf{P} & 0.246 & 0.211 & 0.200 & 0.228 & 0.105 & 0.305 & 0.105 & 0.158 & 0.212 & 0.176 & 0.186 & 0.187 & \textbf{0.331} & 0.166 & 0.168 & 0.192 & 0.136 & \underline{0.328} \\
                         & \textbf{R}  & 0.426 & 0.740 & 0.821 & 0.432 & \underline{0.999} & 0.261 & \textbf{1.000} & 0.621 & 0.639 & 0.692 & 0.745 & 0.665 & 0.216 & 0.455 & 0.439 & 0.398 & 0.518 & 0.212 \\
                         & \textbf{F1}   & 0.312 & \textbf{0.329} & \underline{0.322} & 0.299 & 0.191 & 0.281 & 0.191 & 0.251 & 0.319 & 0.281 & 0.298 & 0.292 & 0.262 & 0.243 & 0.243 & 0.259 & 0.216 & 0.257 \\
\multirow{3}{*}{\textbf{ECG}} & \textbf{P} & \textbf{0.529} & 0.458 & 0.465 & \underline{0.471} & 0.152 & 0.294 & 0.152 & 0.404 & 0.463 & 0.434 & 0.396 & 0.371 & 0.450 & 0.430 & 0.447 & 0.400 & 0.184 & 0.344 \\
                         & \textbf{R}   & 0.668 & 0.671 & 0.613 & 0.630 & \underline{0.999} & 0.678 & \textbf{1.000} & 0.628 & 0.496 & 0.680 & 0.632 & 0.598 & 0.685 & 0.701 & 0.661 & 0.699 & 0.687 & 0.659 \\
                         & \textbf{F1}   & \textbf{0.590} & \underline{0.545} & 0.529 & 0.539 & 0.263 & 0.410 & 0.263 & 0.491 & 0.479 & 0.530 & 0.487 & 0.458 & 0.543 & 0.533 & 0.533 & 0.509 & 0.291 & 0.452 \\
\multirow{3}{*}{\textbf{PSM}} & \textbf{P} & \textbf{0.425} & \underline{0.422} & 0.291 & 0.362 & 0.278 & 0.325 & 0.278 & 0.278 & 0.278 & 0.279 & 0.286 & 0.296 & 0.302 & 0.315 & 0.295 & 0.278 & 0.404 & 0.319 \\
                         & \textbf{R}   & 0.712 & 0.564 & 0.884 & 0.596 & \textbf{1.000} & 0.852 & 0.998 & 0.998 & 0.999 & 0.982 & 0.904 & 0.843 & 0.870 & 0.959 & 0.952 & \underline{0.999} & 0.843 & 0.957 \\
                         & \textbf{F1}   & \underline{0.532} & 0.483 & 0.438 & 0.451 & 0.435 & 0.471 & 0.435 & 0.435 & 0.435 & 0.435 & 0.435 & 0.438 & 0.448 & 0.474 & 0.451 & 0.435 & \textbf{0.546} & 0.479 \\
\multirow{3}{*}{\textbf{SMAP}} & \textbf{P} & 0.160 & \underline{0.166} & 0.162 & 0.128 & 0.128 & 0.129 & 0.128 & 0.128 & 0.139 & 0.152 & 0.128 & 0.161 & 0.130 & 0.134 & 0.128 & 0.128 & \textbf{0.213} & 0.131 \\
                         & \textbf{R}   & 0.688 & 0.841 & 0.804 & \underline{1.000} & 0.999 & 0.969 & 0.998 & 0.999 & 0.691 & 0.922 & \textbf{1.000} & 0.685 & 0.978 & 0.891 & 0.999 & 0.998 & 0.436 & 0.918 \\
                         & \textbf{F1}   & 0.260 & \underline{0.278} & 0.269 & 0.227 & 0.227 & 0.228 & 0.227 & 0.227 & 0.232 & 0.261 & 0.227 & 0.261 & 0.229 & 0.233 & 0.227 & 0.227 & \textbf{0.286} & 0.230 \\
\multirow{3}{*}{\textbf{KDD21}} & \textbf{P}  & \underline{0.125} & 0.119 & 0.080 & \textbf{0.133} & 0.017 & 0.112 & 0.075 & 0.104 & 0.090 & 0.079 & 0.093 & 0.078 & 0.096 & 0.107 & 0.083 & 0.102 & 0.057 & 0.065 \\
                        & \textbf{R}  & 0.631 & 0.624 & 0.583 & 0.533 & \textbf{0.692} & 0.641 & 0.645 & 0.514 & 0.603 & 0.622 & 0.511 & 0.607 & \underline{0.670} & 0.618 & 0.612 & 0.491 & 0.571 & 0.632 \\
                        & \textbf{F1} & 0.071 & 0.069 & 0.065 & \underline{0.077} & 0.018 & 0.055 & 0.056 & \textbf{0.089} & 0.039 & 0.041 & 0.068 & 0.038 & 0.044 & 0.049 & 0.053 & 0.053 & 0.054 & 0.034 \\

\midrule   
\multirow{3}{*}{\textbf{Avg}} & \textbf{P}  & \textbf{0.309} & 0.255 & 0.212 & 0.267 & 0.096 & 0.232 & 0.103 & 0.207 & 0.208 & 0.203 & 0.187 & 0.212 & \underline{0.282} & 0.260 & 0.163 & 0.145 & 0.151 & 0.267 \\
                            & \textbf{R}  & 0.606 & 0.654 & 0.604 & 0.595 & \textbf{0.851} & 0.521 & \underline{0.840} & 0.561 & 0.529 & 0.632 & 0.582 & 0.604 & 0.551 & 0.605 & 0.677 & 0.823 & 0.503 & 0.550 \\
                            & \textbf{F1} & \textbf{0.357} & \underline{0.325} & 0.285 & 0.277 & 0.160 & 0.257 & 0.164 & 0.267 & 0.260 & 0.275 & 0.255 & 0.276 & 0.276 & 0.274 & 0.226 & 0.201 & 0.213 & 0.264 \\
\midrule 
\multicolumn{2}{c|}{\textbf{$1^{\text{st}}$ count}} & \textbf{8} & 1 & 2 & 1 & 4 & 0 & 3 & 2 & 0 & 0 & 1 & 0 & 1 & 0 & 1 & 0 & 3 & 0\\
\bottomrule
\end{tabular}
}
\label{tab:p_r_f1}
\end{table}

\begin{table}[h]
\centering
\footnotesize
\caption{Average Aff-P (Affiliated-Precision), Aff-R (Affiliated-Recall) and Aff-F (Affiliated-F1-score) across all datasets. The best results are in bold, while the second-best results are underlined.}
\renewcommand{\arraystretch}{1.5}
\resizebox{\textwidth}{!}{
\begin{tabular}{c | c | c c c c c c c c c c c c c c c c c c}
\noalign{\vskip +3ex}
\toprule
\textbf{Dataset} & \textbf{Metric} & \textbf{CARE} & \textbf{Cross} & \textbf{CATCH} & \textbf{Mixer} & \textbf{DC} & \textbf{TranAD} & \textbf{ATrans} & \textbf{CALF} & \textbf{GS} & \textbf{Patch} & \textbf{iTrans} & \textbf{TsNet} & \textbf{LSTM} & \textbf{AE} & \textbf{OCSVM} & \textbf{IF} & \textbf{LOF} & \textbf{PCA}\\
\midrule
\multirow{3}{*}{\textbf{CalIt2}} & \textbf{Aff-P} & 0.651 & 0.658 & 0.649 & \underline{0.666} & 0.507 & 0.650 & 0.507 & 0.641 & 0.657 & 0.648 & 0.622 & 0.610 & 0.644 & 0.651 & 0.662 & 0.507 & 0.598 & \textbf{0.671} \\
                        & \textbf{Aff-R} & 0.994 & 0.984 & 0.978 & 0.938 & 1.000 & 0.975 & \textbf{1.000} & 0.918 & 0.968 & 0.988 & 0.982 & 0.964 & 0.993 & 0.994 & 0.976 & \underline{1.000} & 0.957 & 0.895 \\
                         & \textbf{Aff-F} & \underline{0.787} & \textbf{0.788} & 0.780 & 0.779 & 0.673 & 0.780 & 0.673 & 0.755 & 0.783 & 0.782 & 0.761 & 0.747 & 0.782 & 0.786 & 0.789 & 0.673 & 0.736 & 0.767 \\
\multirow{3}{*}{\textbf{NYC}} & \textbf{Aff-P} & \textbf{0.532} & 0.510 & 0.502 & 0.497 & \underline{0.528} & 0.494 & 0.499 & 0.489 & 0.526 & 0.491 & 0.508 & 0.495 & 0.505 & 0.525 & 0.500 & 0.500 & 0.490 & 0.525 \\
                         & \textbf{Aff-R} & 0.998 & 0.999 & 0.997 & \textbf{1.000} & 0.995 & 0.997 & 0.993 & 0.999 & 0.996 & 0.998 & 0.997 & 0.998 & 0.996 & 0.998 & \underline{1.000} & 0.999 & 0.999 & 0.996 \\
                         & \textbf{Aff-F} & \textbf{0.694} & 0.675 & 0.668 & 0.664 & 0.690 & 0.661 & 0.665 & 0.657 & 0.689 & 0.658 & 0.673 & 0.662 & 0.670 & \underline{0.688} & 0.667 & 0.667 & 0.658 & 0.687 \\
\multirow{3}{*}{\textbf{GECCO}} & \textbf{Aff-P} & \underline{0.886} & 0.809 & 0.767 & 0.560 & 0.502 & \textbf{0.983} & 0.502 & 0.762 & 0.838 & 0.752 & 0.706 & 0.844 & 0.696 & 0.700 & 0.484 & 0.502 & 0.602 & 0.802 \\
                         & \textbf{Aff-R} & 0.997 & 0.974 & 0.988 & 0.286 & \underline{1.000} & 0.091 & \textbf{1.000} & 0.806 & 0.990 & 0.960 & 0.978 & 0.981 & 0.235 & 0.272 & 0.453 & 1.000 & 0.354 & 0.146 \\
                         & \textbf{Aff-F} & \textbf{0.938} & 0.884 & 0.863 & 0.379 & 0.669 & 0.166 & 0.669 & 0.783 & \underline{0.908} & 0.844 & 0.820 & 0.908 & 0.351 & 0.392 & 0.468 & 0.669 & 0.446 & 0.248 \\
\multirow{3}{*}{\textbf{MSL}}  & \textbf{Aff-P} & \textbf{0.648} & 0.629 & 0.626 & 0.636 & 0.514 & \underline{0.644} & 0.514 & 0.540 & 0.626 & 0.612 & 0.617 & 0.613 & 0.536 & 0.527 & 0.527 & 0.624 & 0.539 & 0.602 \\
                         & \textbf{Aff-R} & 0.949 & 0.961 & 0.996 & 0.970 & \underline{0.999} & 0.935 & \textbf{1.000} & 0.992 & 0.994 & 0.998 & 0.997 & 0.997 & 0.419 & 0.994 & 0.993 & 0.993 & 0.868 & 0.355 \\
                         & \textbf{Aff-F} & \textbf{0.770} & 0.761 & \underline{0.769} & 0.768 & 0.679 & 0.763 & 0.679 & 0.699 & 0.768 & 0.759 & 0.762 & 0.760 & 0.470 & 0.689 & 0.688 & 0.766 & 0.665 & 0.446 \\
\multirow{3}{*}{\textbf{ECG}} & \textbf{Aff-P} & \underline{0.834} & 0.791 & 0.773 & 0.794 & 0.572 & 0.616 & 0.572 & 0.703 & 0.758 & 0.764 & 0.689 & 0.660 & \textbf{0.846} & 0.774 & 0.787 & 0.687 & 0.561 & 0.610 \\
                         & \textbf{Aff-R} & 0.793 & 0.809 & 0.773 & 0.778 & \textbf{1.000} & 0.716 & \underline{0.999} & 0.814 & 0.650 & 0.816 & 0.801 & 0.764 & 0.796 & 0.812 & 0.792 & 0.907 & 0.855 & 0.599 \\
                         & \textbf{Aff-F} &\textbf{0.813} & \underline{0.800} & 0.773 & 0.786 & 0.728 & 0.663 & 0.728 & 0.754 & 0.700 & 0.789 & 0.741 & 0.709 & 0.790 & 0.793 & 0.790 & 0.782 & 0.677 & 0.605 \\
\multirow{3}{*}{\textbf{PSM}} & \textbf{Aff-P} & \underline{0.593} & \textbf{0.619} & 0.538 & 0.574 & 0.532 & 0.557 & 0.532 & 0.532 & 0.532 & 0.533 & 0.538 & 0.545 & 0.533 & 0.534 & 0.532 & 0.532 & 0.597 & 0.538 \\
                       & \textbf{Aff-R} & 0.999 & 0.998 & 0.998 & 0.998 & \textbf{1.000} & 0.997 & 0.999 & \underline{1.000} & 1.000 & 1.000 & 1.000 & 0.998 & 0.991 & 0.992 & 0.996 & 0.997 & 0.974 & 0.986 \\
                         & \textbf{Aff-F} & \underline{0.744} & \textbf{0.764} & 0.699 & 0.729 & 0.694 & 0.715 & 0.694 & 0.694 & 0.694 & 0.696 & 0.699 & 0.705 & 0.693 & 0.694 & 0.693 & 0.694 & 0.740 & 0.696 \\
\multirow{3}{*}{\textbf{SMAP}} & \textbf{Aff-P} & \textbf{0.565} & \underline{0.554} & 0.544 & 0.515 & 0.515 & 0.517 & 0.515 & 0.515 & 0.529 & 0.546 & 0.515 & 0.544 & 0.517 & 0.519 & 0.515 & 0.515 & 0.494 & 0.521 \\
                        & \textbf{Aff-R} & 0.995 & 0.996 & 0.996 & \textbf{1.000} & 0.999 & \underline{1.000} & 0.997 & 0.998 & 0.998 & 0.998 & 1.000 & 0.996 & 1.000 & 0.999 & 1.000 & 0.999 & 0.739 & 0.998 \\
                        & \textbf{Aff-F} & \textbf{0.721} & \underline{0.712} & 0.704 & 0.680 & 0.680 & 0.682 & 0.680 & 0.680 & 0.692 & 0.706 & 0.680 & 0.703 & 0.681 & 0.683 & 0.680 & 0.680 & 0.592 & 0.685 \\
\multirow{3}{*}{\textbf{KDD21}}& \textbf{Aff-P} & \textbf{0.565 }& 0.560 & 0.538 & 0.570 & 0.503 & \underline{0.564} & 0.539 & 0.563 & 0.557 & 0.540 & 0.552 & 0.524 & 0.544 & 0.557 & 0.548 & 0.554 & 0.546 & 0.553 \\
                        & \textbf{Aff-R} & \underline{0.999} & 0.997 & 0.997 & 0.996 & 0.997 & 0.996 & 0.997 & 0.999 & 0.995 & 0.993 & 0.995 & 0.994 & 0.998 & 0.998 & 0.997 & 0.996 & \textbf{0.999} & 0.992 \\
                        & \textbf{Aff-F} & \textbf{0.716} & 0.712 & 0.693 & 0.714 & 0.669 & 0.714 & 0.695 & \underline{0.715} & 0.709 & 0.696 & 0.707 & 0.683 & 0.698 & 0.708 & 0.704 & 0.708 & 0.702 & 0.707 \\
\midrule   
\multirow{3}{*}{\textbf{Avg}} & \textbf{Aff-P} & \textbf{0.659} & \underline{0.641} & 0.617 & 0.601 & 0.522 & 0.628 & 0.523 & 0.593 & 0.628 & 0.611 & 0.593 & 0.605 & 0.602 & 0.598 & 0.569 & 0.553 & 0.553 & 0.603 \\
                        & \textbf{Aff-R} & 0.965 & 0.965 & 0.966 & 0.871 & \underline{0.998} & 0.839 & \textbf{0.999} & 0.941 & 0.950 & 0.970 & 0.969 & 0.963 & 0.804 & 0.882 & 0.901 & 0.987 & 0.843 & 0.747 \\
                        & \textbf{Aff-F} & \textbf{0.773} & \underline{0.762} & 0.744 & 0.687 & 0.685 & 0.643 & 0.685 & 0.717 & 0.743 & 0.741 & 0.730 & 0.735 & 0.642 & 0.679 & 0.685 & 0.705 & 0.652 & 0.605 \\
\midrule 
\multicolumn{2}{c|}{\textbf{$1^{\text{st}}$ count}} & \textbf{12} & 3 & 0 & 2 & 2 & 1 & 4 & 0 & 0 & 0 & 0 & 0 & 1 & 0 & 0 & 0 & 1 & 1\\
\bottomrule
\end{tabular}
}
\label{tab:aff-p-r-f1}
\end{table}

\begin{table}[h]
\centering
\footnotesize
\caption{Average A-P (AUC-PR), R-A-P (R-AUC-PR) and V-PR (VUS-PR) across all datasets. The best results are in bold, while the second-best results are underlined.}
\renewcommand{\arraystretch}{1.5}
\resizebox{\textwidth}{!}{
\begin{tabular}{c | c | c c c c c c c c c c c c c c c c c c}
\noalign{\vskip +3ex}
\toprule
\textbf{Dataset} & \textbf{Metric} & \textbf{CARE} & \textbf{Cross} & \textbf{CATCH} & \textbf{Mixer} & \textbf{DC} & \textbf{TranAD} & \textbf{ATrans} & \textbf{CALF} & \textbf{GS} & \textbf{Patch} & \textbf{iTrans} & \textbf{TsNet} & \textbf{LSTM} & \textbf{AE} & \textbf{OCSVM} & \textbf{IF} & \textbf{LOF} & \textbf{PCA}\\
\midrule
\multirow{3}{*}{\textbf{CalIt2}} & \textbf{A-P} & \textbf{0.100} & \underline{0.099} & 0.094 & 0.097 & 0.029 & 0.052 & 0.030 & 0.048 & 0.087 & 0.093 & 0.094 & 0.079 & 0.078 & 0.081 & 0.095 & 0.029 & 0.035 & 0.073 \\
                        & \textbf{R-A-P} & 0.114 & \textbf{0.121} & \underline{0.117} & 0.113 & 0.028 & 0.059 & 0.056 & 0.065 & 0.093 & 0.105 & 0.112 & 0.092 & 0.091 & 0.104 & 0.109 & 0.120 & 0.062 & 0.106 \\
                        & \textbf{V-PR} & 0.111 & \underline{0.119} & 0.114 & 0.112 & 0.028 & 0.057 & 0.059 & 0.063 & 0.091 & 0.101 & 0.109 & 0.093 & 0.090 & 0.101 & 0.109 & \textbf{0.121} & 0.060 & 0.103 \\
\multirow{3}{*}{\textbf{NYC}} & \textbf{A-P} & \textbf{0.090} & \underline{0.086} & 0.048 & 0.024 & 0.041 & 0.027 & 0.026 & 0.051 & 0.034 & 0.032 & 0.040 & 0.033 & 0.047 & 0.043 & 0.020 & 0.022 & 0.020 & 0.046 \\
                         & \textbf{R-A-P} & 0.081 & 0.075 & 0.063 & 0.042 & 0.056 & 0.052 & 0.038 & 0.075 & 0.049 & 0.055 & 0.058 & 0.051 & 0.076 & 0.102 & \textbf{0.333} & \underline{0.121} & 0.053 & 0.076 \\
                         & \textbf{V-PR} & 0.081 & 0.075 & 0.064 & 0.041 & 0.059 & 0.052 & 0.037 & 0.072 & 0.051 & 0.053 & 0.056 & 0.048 & 0.072 & 0.092 & \textbf{0.318} & \underline{0.121} & 0.049 & 0.072 \\
\multirow{3}{*}{\textbf{GECCO}} & \textbf{A-P} & \textbf{0.623} & \underline{0.596} & 0.331 & 0.310 & 0.011 & 0.191 & 0.011 & 0.321 & 0.295 & 0.347 & 0.154 & 0.479 & 0.284 & 0.277 & 0.039 & 0.011 & 0.086 & 0.234 \\
                         & \textbf{R-A-P} & \underline{0.514} & \textbf{0.522} & 0.376 & 0.134 & 0.508 & 0.038 & 0.011 & 0.294 & 0.388 & 0.343 & 0.162 & 0.442 & 0.040 & 0.053 & 0.098 & 0.508 & 0.058 & 0.047 \\
                         & \textbf{V-PR} & \underline{0.509} & \textbf{0.528} & 0.363 & 0.136 & 0.508 & 0.038 & 0.011 & 0.294 & 0.385 & 0.343 & 0.159 & 0.441 & 0.040 & 0.053 & 0.101 & 0.508 & 0.058 & 0.046 \\
\multirow{3}{*}{\textbf{MSL}} & \textbf{A-P} & \underline{0.266} & \textbf{0.272} & 0.228 & 0.247 & 0.108 & 0.188 & 0.098 & 0.178 & 0.225 & 0.201 & 0.213 & 0.207 & 0.193 & 0.144 & 0.153 & 0.173 & 0.128 & 0.183 \\
                         & \textbf{R-A-P} & \underline{0.330} & \textbf{0.335} & 0.300 & 0.305 & 0.186 & 0.237 & 0.111 & 0.241 & 0.295 & 0.281 & 0.287 & 0.274 & 0.179 & 0.194 & 0.185 & 0.226 & 0.188 & 0.172 \\
                         & \textbf{V-PR} & \underline{0.323} & \textbf{0.329} & 0.296 & 0.300 & 0.186 & 0.236 & 0.111 & 0.238 & 0.291 & 0.277 & 0.284 & 0.271 & 0.179 & 0.193 & 0.185 & 0.224 & 0.187 & 0.172 \\
\multirow{3}{*}{\textbf{ECG}} & \textbf{A-P} & \textbf{0.529} & 0.503 & \underline{0.505} & 0.495 & 0.147 & 0.272 & 0.157 & 0.486 & 0.506 & 0.490 & 0.483 & 0.412 & 0.500 & 0.478 & 0.480 & 0.478 & 0.202 & 0.347 \\
                       & \textbf{R-A-P} & \textbf{0.476} & 0.451 & \underline{0.452} & 0.445 & 0.134 & 0.265 & 0.108 & 0.412 & 0.407 & 0.433 & 0.405 & 0.357 & 0.450 & 0.432 & 0.433 & 0.423 & 0.181 & 0.305 \\
                       & \textbf{V-PR} & \textbf{0.505} & 0.485 & \underline{0.489 } & 0.480 & 0.153 & 0.300 & 0.121 & 0.453 & 0.447 & 0.467 & 0.448 & 0.404 & 0.470 & 0.464 & 0.466 & 0.465 & 0.209 & 0.343 \\
\multirow{3}{*}{\textbf{PSM}} & \textbf{A-P} & \textbf{0.569} & \underline{0.465} & 0.381 & 0.438 & 0.278 & 0.380 & 0.290 & 0.344 & 0.379 & 0.370 & 0.377 & 0.394 & 0.433 & 0.476 & 0.418 & 0.337 & 0.462 & 0.477 \\
                         & \textbf{R-A-P} & \textbf{0.524} & \underline{0.465} & 0.382 & 0.434 & 0.440 & 0.378 & 0.282 & 0.344 & 0.381 & 0.370 & 0.379 & 0.395 & 0.402 & 0.429 & 0.369 & 0.337 & 0.422 & 0.412 \\
                        & \textbf{V-PR} & \textbf{0.524} & \underline{0.467} & 0.383 & 0.435 & 0.440 & 0.379 & 0.283 & 0.344 & 0.382 & 0.371 & 0.380 & 0.397 & 0.402 & 0.430 & 0.370 & 0.337 & 0.423 & 0.413 \\
\multirow{3}{*}{\textbf{SMAP}} & \textbf{A-P} & 0.133 & 0.132 & 0.124 & 0.133 & 0.136 & 0.110 & 0.124 & 0.109 & 0.124 & \underline{0.140} & 0.114 & 0.134 & 0.103 & 0.109 & 0.101 & 0.129 & \textbf{0.183} & 0.109 \\
                         & \textbf{R-A-P} & 0.160 & 0.155 & 0.147 & 0.159 & 0.166 & 0.127 & 0.134 & 0.127 & 0.144 & \underline{0.168} & 0.137 & 0.159 & 0.120 & 0.162 & 0.152 & 0.147 & \textbf{0.191} & 0.129 \\
                        & \textbf{V-PR} & 0.160 & 0.155 & 0.147 & 0.158 & 0.166 & 0.127 & 0.135 & 0.127 & 0.144 & \underline{0.167} & 0.137 & 0.158 & 0.121 & 0.162 & 0.152 & 0.147 & \textbf{0.191} & 0.129 \\
\multirow{3}{*}{\textbf{KDD21}} & \textbf{A-P}   & \underline{0.047} & 0.044 & 0.042 & 0.046 & 0.009 & 0.033 & 0.040 & \textbf{0.058} & 0.023 & 0.025 & 0.042 & 0.021 & 0.023 & 0.030 & 0.028 & 0.029 & 0.024 & 0.019 \\
                        & \textbf{R-A-P} & 0.038 & 0.034 & 0.026 & 0.049 & 0.017 & 0.029 & \underline{0.052} & \textbf{0.063} & 0.025 & 0.026 & 0.038 & 0.023 & 0.022 & 0.026 & 0.028 & 0.027 & 0.031 & 0.021 \\
                        & \textbf{V-PR}  & 0.047 & 0.042 & 0.032 & \underline{0.050} & 0.017 & 0.030 & 0.048 & \textbf{0.063} & 0.033 & 0.033 & 0.045 & 0.032 & 0.022 & 0.027 & 0.029 & 0.031 & 0.031 & 0.021 \\
\midrule   
\multirow{3}{*}{\textbf{Avg}} & \textbf{A-P}   & \textbf{0.295} & \underline{0.275} & 0.219 & 0.224 & 0.095 & 0.157 & 0.097 & 0.199 & 0.209 & 0.212 & 0.190 & 0.220 & 0.208 & 0.205 & 0.167 & 0.151 & 0.143 & 0.186 \\
                        & \textbf{R-A-P} & \textbf{0.280} & \underline{0.270} & 0.233 & 0.210 & 0.192 & 0.148 & 0.099 & 0.203 & 0.223 & 0.223 & 0.197 & 0.224 & 0.173 & 0.188 & 0.213 & 0.239 & 0.148 & 0.158 \\
                        & \textbf{V-PR}  & \textbf{0.283} & \underline{0.275} & 0.236 & 0.214 & 0.195 & 0.152 & 0.101 & 0.207 & 0.228 & 0.227 & 0.202 & 0.230 & 0.175 & 0.190 & 0.216 & 0.244 & 0.151 & 0.162 \\
\midrule 
\multicolumn{2}{c|}{\textbf{$1^{\text{st}}$ count}} & \textbf{12} & 6 & 0 & 0 & 0 & 0 & 0 & 3 & 0 & 0 & 0 & 1 & 0 & 0 & 2 & 1 & 3 & 0\\
\bottomrule
\end{tabular}
}
\label{tab:A-P}
\end{table}

\begin{table}[t]
\centering
\footnotesize
\caption{Average A-R (AUC-ROC), R-A-R (R-AUC-ROC) and V-ROC (VUS-ROC) across all datasets. The best results are in bold, while the second-best results are underlined.}
\renewcommand{\arraystretch}{1.5}
\resizebox{\textwidth}{!}{
\begin{tabular}{c | c | c c c c c c c c c c c c c c c c c c}
\noalign{\vskip +3ex}
\toprule
\textbf{Dataset} & \textbf{Metric} & \textbf{CARE} & \textbf{Cross} & \textbf{CATCH} & \textbf{Mixer} & \textbf{DC} & \textbf{TranAD} & \textbf{ATrans} & \textbf{CALF} & \textbf{GS} & \textbf{Patch} & \textbf{iTrans} & \textbf{TsNet} & \textbf{LSTM} & \textbf{AE} & \textbf{OCSVM} & \textbf{IF} & \textbf{LOF} & \textbf{PCA}\\
\midrule
\multirow{3}{*}{\textbf{CalIt2}} & \textbf{A-R} & \underline{0.817} & \textbf{0.830} & 0.775 & 0.787 & 0.474 & 0.500 & 0.505 & 0.567 & 0.735 & 0.765 & 0.783 & 0.741 & 0.674 & 0.727 & 0.804 & 0.500 & 0.494 & 0.790 \\
                        & \textbf{R-A-R} & \underline{0.848} & \textbf{0.856} & 0.814 & 0.821 & 0.465 & 0.527 & 0.499 & 0.652 & 0.774 & 0.812 & 0.813 & 0.770 & 0.722 & 0.788 & 0.840 & 0.503 & 0.546 & 0.796 \\
                        & \textbf{V-ROC} & \underline{0.838} & \textbf{0.845} & 0.799 & 0.808 & 0.465 & 0.519 & 0.500 & 0.627 & 0.762 & 0.797 & 0.806 & 0.762 & 0.710 & 0.772 & 0.828 & 0.503 & 0.519 & 0.786 \\
\multirow{3}{*}{\textbf{NYC}} & \textbf{A-R} & \textbf{0.776} & \underline{0.710} & 0.625 & 0.563 & 0.524 & 0.556 & 0.522 & 0.664 & 0.549 & 0.541 & 0.623 & 0.579 & 0.667 & 0.681 & 0.456 & 0.500 & 0.464 & 0.666 \\
                         & \textbf{R-A-R} & \textbf{0.769} & 0.742 & 0.676 & 0.656 & 0.523 & 0.647 & 0.506 & 0.739 & 0.634 & 0.692 & 0.707 & 0.672 & 0.731 & 0.736 & 0.612 & 0.505 & 0.604 & \underline{0.744} \\
                         & \textbf{V-ROC} & \textbf{0.763} & \underline{0.732} & 0.674 & 0.637 & 0.527 & 0.633 & 0.505 & 0.721 & 0.640 & 0.667 & 0.690 & 0.651 & 0.716 & 0.725 & 0.589 & 0.505 & 0.581 & 0.730 \\
\multirow{3}{*}{\textbf{GECCO}} & \textbf{A-R} & 0.909 & \underline{0.969} & 0.949 & 0.871 & 0.500 & 0.594 & 0.476 & 0.923 & 0.911 & 0.959 & 0.793 & \textbf{0.974} & 0.774 & 0.724 & 0.804 & 0.500 & 0.796 & 0.711 \\
                         & \textbf{R-A-R} & 0.911 & \textbf{0.982} & 0.970 & 0.810 & 0.501 & 0.463 & 0.478 & 0.936 & 0.955 & 0.955 & 0.853 & \underline{0.972} & 0.665 & 0.586 & 0.757 & 0.501 & 0.777 & 0.598 \\
                         & \textbf{V-ROC} & 0.947 & \textbf{0.980} & 0.967 & 0.818 & 0.501 & 0.457 & 0.479 & 0.934 & 0.950 & 0.955 & 0.846 & \underline{0.971} & 0.665 & 0.588 & 0.756 & 0.501 & 0.767 & 0.595 \\
\multirow{3}{*}{\textbf{MSL}} & \textbf{A-R} & 0.732 & \textbf{0.759} & \underline{0.755} & 0.685 & 0.514 & 0.562 & 0.397 & 0.619 & 0.730 & 0.709 & 0.725 & 0.723 & 0.541 & 0.529 & 0.524 & 0.600 & 0.551 & 0.508 \\
                         & \textbf{R-A-R} & 0.788 & \textbf{0.800} & \underline{0.795} & 0.748 & 0.536 & 0.630 & 0.434 & 0.693 & 0.782 & 0.768 & 0.787 & 0.774 & 0.563 & 0.601 & 0.594 & 0.657 & 0.608 & 0.539 \\
                         & \textbf{V-ROC} & 0.779 & \textbf{0.793} & \underline{0.790} & 0.742 & 0.535 & 0.625 & 0.434 & 0.685 & 0.775 & 0.762 & 0.780 & 0.769 & 0.558 & 0.597 & 0.590 & 0.653 & 0.598 & 0.532 \\
\multirow{3}{*}{\textbf{ECG}} & \textbf{A-R} & \textbf{0.866} & \underline{0.865} & 0.861 & 0.859 & 0.479 & 0.750 & 0.451 & 0.838 & 0.826 & 0.858 & 0.837 & 0.811 & 0.858 & 0.860 & 0.858 & 0.842 & 0.589 & 0.784 \\
                         & \textbf{R-A-R} & 0.799 & \textbf{0.810} & 0.803 & 0.805 & 0.434 & 0.652 & 0.420 & 0.773 & 0.757 & 0.796 & 0.771 & 0.738 & 0.792 & 0.800 & \underline{0.801} & 0.800 & 0.419 & 0.647 \\
                         & \textbf{V-ROC} & 0.818 & \textbf{0.829} & 0.825 & 0.825 & 0.444 & 0.676 & 0.427 & 0.799 & 0.783 & 0.816 & 0.798 & 0.767 & 0.805 & 0.818 & 0.820 & \underline{0.823} & 0.470 & 0.672 \\
\multirow{3}{*}{\textbf{PSM}} & \textbf{A-R} & \textbf{0.721} & 0.665 & 0.595 & 0.633 & 0.500 & 0.647 & 0.500 & 0.569 & 0.579 & 0.574 & 0.587 & 0.605 & 0.634 & 0.660 & 0.619 & 0.545 & \underline{0.720} & 0.667 \\
                       & \textbf{R-A-R} & \textbf{0.698} & \underline{0.662} & 0.595 & 0.629 & 0.501 & 0.611 & 0.480 & 0.565 & 0.580 & 0.571 & 0.585 & 0.603 & 0.577 & 0.597 & 0.530 & 0.547 & 0.639 & 0.573 \\
                        & \textbf{V-ROC} & \textbf{0.696} & \underline{0.662} & 0.596 & 0.629 & 0.501 & 0.611 & 0.480 & 0.563 & 0.580 & 0.572 & 0.586 & 0.605 & 0.578 & 0.598 & 0.532 & 0.545 & 0.641 & 0.576 \\
\multirow{3}{*}{\textbf{SMAP}} & \textbf{A-R} & 0.535 & 0.526 & 0.504 & 0.493 & 0.509 & 0.418 & 0.477 & 0.420 & 0.485 & 0.526 & 0.447 & \underline{0.551} & 0.396 & 0.426 & 0.359 & 0.508 & \textbf{0.619} & 0.423 \\
                         & \textbf{R-A-R} & \textbf{0.596} & 0.592 & 0.552 & 0.543 & 0.520 & 0.457 & 0.483 & 0.469 & 0.520 & 0.590 & 0.502 & \underline{0.595} & 0.432 & 0.455 & 0.396 & 0.522 & 0.533 & 0.437 \\
                        & \textbf{V-ROC} & \textbf{0.594} & 0.580 & 0.548 & 0.541 & 0.520 & 0.455 & 0.483 & 0.466 & 0.520 & 0.586 & 0.498 & \underline{0.591} & 0.431 & 0.454 & 0.395 & 0.523 & 0.535 & 0.435 \\
\multirow{3}{*}{\textbf{KDD21}} & \textbf{A-R}   & 0.567 & 0.586 & 0.606 & \underline{0.657} & 0.478 & 0.590 & 0.530 & \textbf{0.667} & 0.598 & 0.605 & 0.631 & 0.581 & 0.582 & 0.622 & 0.609 & 0.498 & 0.592 & 0.575 \\
                        & \textbf{R-A-R} & 0.630 & 0.633 & 0.656 & \underline{0.708} & 0.492 & \textbf{0.634} & 0.541 & 0.713 & 0.665 & 0.664 & 0.690 & 0.652 & 0.649 & 0.672 & 0.663 & 0.582 & 0.637 & 0.627 \\
                        & \textbf{V-ROC} & 0.628 & 0.633 & 0.650 & 0.699 & 0.494 & 0.627 & 0.538 & \textbf{0.719} & 0.658 & 0.661 & \underline{0.685} & 0.648 & 0.647 & 0.669 & 0.654 & 0.579 & 0.637 & 0.625 \\
\midrule   
\multirow{3}{*}{\textbf{Avg}} & \textbf{A-R}   & \textbf{0.740} & \underline{0.739} & 0.709 & 0.694 & 0.497 & 0.577 & 0.482 & 0.658 & 0.677 & 0.692 & 0.678 & 0.696 & 0.641 & 0.654 & 0.629 & 0.562 & 0.603 & 0.641 \\
                    & \textbf{R-A-R} & \underline{0.755} & \textbf{0.760} & 0.733 & 0.715 & 0.497 & 0.578 & 0.480 & 0.693 & 0.708 & 0.731 & 0.713 & 0.722 & 0.641 & 0.655 & 0.649 & 0.577 & 0.595 & 0.620 \\
                    & \textbf{V-ROC} & \textbf{0.758} & \underline{0.757} & 0.731 & 0.712 & 0.498 & 0.575 & 0.481 & 0.689 & 0.708 & 0.727 & 0.711 & 0.720 & 0.639 & 0.653 & 0.646 & 0.579 & 0.594 & 0.619 \\
\midrule 
\multicolumn{2}{c|}{\textbf{$1^{\text{st}}$ count}} & \textbf{11} & 11 & 0 & 0 & 0 & 0 & 0 & 3 & 0 & 0 & 0 & 1 & 0 & 0 & 0 & 0 & 1 & 0\\
\bottomrule
\end{tabular}
}
\label{tab:A-R}
\end{table}

%%%%%%%%%%%%%%%%%%%%%%%%%%%%%%%%%%%%%%%%%%%%%%%%%%%%%%%%%%%%

\end{document}